\documentclass[11pt]{article}
\usepackage{graphicx,psfrag,amsmath,amsfonts,verbatim,mathtools}
\usepackage{times}
\usepackage{fancyhdr}
\usepackage{enumerate}
\usepackage{enumitem}
\usepackage{mathrsfs}
\usepackage{soul}
\usepackage{listings}
\usepackage{setspace}
\usepackage[perpage]{footmisc}
\usepackage[colorlinks,linkcolor=blue,urlcolor=blue]{hyperref}
\usepackage{placeins}
\usepackage{float}
\usepackage{graphicx}
\usepackage{subfigure}
\usepackage{epstopdf}
\usepackage{float}
\usepackage{geometry}
\usepackage{algorithm}
\usepackage{algorithmic}
\usepackage{endnotes}
\usepackage{multirow}
\usepackage{booktabs}
\usepackage{makecell}
\usepackage{footnotebackref}
\usepackage{tablefootnote}
\usepackage{diagbox}
\usepackage{breakcites}
\usepackage{amsmath,amsfonts,bm}
\usepackage{amssymb}
\usepackage{framed}
\usepackage{color}
\usepackage{bm}
\usepackage{amsthm}
\usepackage{xcolor}
\usepackage{framed}
\usepackage{lipsum}
\usepackage{mdframed}
\usepackage{macros}
\definecolor{gray}{rgb}{0.6,0.6,0.6}

\newtheorem{thm}{Theorem}

\newtheorem{lem}{Lemma}
\newtheorem{assum}{Assumption}

\newtheorem{cor}{Corollary}
\newtheorem*{lem*}{Lemma}
\newtheorem*{thm*}{Theorem}
\newtheorem*{cor*}{Corollary}
\newtheorem*{assum*}{Assumption}

\newcommand{\abs}[1]{\left|#1\right|}
\newcommand{\esp}[1]{\mathbb{E}\left[#1\right]}

\newcommand{\reel}{\mathbb{R}}
\newcommand{\nat}{\mathbb{N}}
\newcommand{\ee}{\mathrm{e}}

\newcommand{\dd}{\mathrm{d}}
\newcommand{\xx}{\mathrm{x}}
\newcommand{\mm}{\mathrm{m}}
\newcommand{\up}{\mathrm{v}}
\newcommand{\g}{\mathrm{g}}
\newcommand{\sqep}[1]{\sqrt{\epsilon + #1}}

\newcommand{\deq}{=}%\tr    iangleq}
\newcommand{\bbratio}{\left(\frac{\beta_1}{\beta_2}\right)}

\newcommand{\field}[1]{\mathbb{#1}}
\newcommand{\E}{\field{E}}
\newcommand{\R}{\field{R}}
\colorlet{shadecolor}{blue!20}

\begin{document}

\title{Dynamic Regret Analysis for Online Meta-Learning}
\author{
	Parvin Nazari\thanks{Department of Mathematics \& Computer Science, Amirkabir University of Technology. Emails:
	\href{mailto:siqiz4@illinois.edu}{p$\_$nazari@aut.ac.ir},
	\href{mailto:niaohe@illinois.edu}{eskhor@aut.ac.ir}.
	}
	\and
	Esmaile Khorram\footnotemark[1]
}
\date{}
\maketitle

\begin{abstract}
The online meta-learning framework has arisen as a powerful tool for the continual lifelong learning setting.
The goal for an agent is to quickly learn new tasks by drawing on prior experience, while it faces with tasks one after another.
This formulation involves two levels: outer level which learns meta-learners and inner level which learns task-specific models, with only a small amount of data from the current task.
While existing methods provide static regret analysis for the online meta-learning framework,
we establish performance in terms of dynamic regret which handles changing environments from a global prospective.
We also build off of a generalized version of the adaptive gradient methods that covers both \textsc{Adam} and \textsc{Adagrad} to learn meta-learners in the outer level.
We carry out our analyses in a stochastic setting, and in expectation prove a logarithmic local dynamic regret which depends explicitly on the total number
of iterations $T$ and parameters of the learner.
Apart from, we also indicate high probability bounds on the convergence rates of proposed algorithm with appropriate selection of parameters, which have not been argued before.
\end{abstract}
\section{Introduction}
The ability to design systems that use prior experience to learn new skills quickly is a critical aspect of artificial intelligence problems, from robotics to pattern recognition and image classification. \textit{Meta-learning} casts
this as the problem of \textit{learning to learn}, in which observed tasks is exploited to obtain
common knowledge improving adaptation to unseen tasks.
Meta-learning methods include optimization-based \cite{finn2017model}, model-based \cite{hochreiter2001learning} and metric \cite{vinyals2016matching} approaches. The standard gradient-based meta-learning framework, one of the optimization-based methods, is identified as the Model-Agnostic Meta-Learning (\textsc{Maml})~\cite{finn2017model}, which can be widely used \cite{collins2020distribution} on account of its simplicity and effectiveness.
However, it presumes that the set of tasks are available together as a batch, which thereby doesn't consider the sequential setting in which the new tasks are revealed one after the other.

The online learning algorithm~\cite{cesa2006prediction} has been shown to be a powerful tool in sequential setting in which at each round $t$, a learner selects
a decision $\mathrm{x}_t \in \mathcal{X}\subseteq \Bbb{R}^d$, and an adversary replies with a loss function
$\ell_t:\mathcal{X}\rightarrow \Bbb{R}$. The losses $\ell_t$ are convex functions over $\mathcal{X}$ which is also convex and typically
employed to as the decision set.
Each round suffers a loss of $\ell_t(\xx_t)$ against the learner, whose goal
is to minimize regret described as
\begin{equation}
\sum^T_{t=1}\ell_t(\mathrm{x}_t) - \min_{\mathrm{x}\in \mathcal{X}}\sum^T_{t=1}\ell_t(\mathrm{x})~.
\label{eq:reg1}
\end{equation}
The above regret is usually viewed as \textit{static} regret to highlight that the comparator is static.
On the downside, the static regret may not be a suitable measure in changing environments \cite{besbes2015non}, leading to growing interest in the
\textit{dynamic} regret \cite{besbes2015non,hall2013online,mokhtari2016online}, in which the performance of online learning is compared to a sequence of optimal solutions.
The dynamic regret is recognized as
\begin{equation*}
\sum^T_{t=1}\ell_t(\xx_t) - \sum^T_{t=1}\ell_t(\xx^*_t)~,
\label{eq:reg}
\end{equation*}
where $\xx^*_t \in \argmin_{x\in \mathcal{X}} \ell_t(\xx)$ is an optimal solution of $\ell_t(\xx)$.
All aforementioned works have focused on convex loss functions to upper bound the dynamic regret in terms of certain regularity of the comparator sequence or the function sequence.
Since the standard notion of regret is not a suitable measure of performance for nonconvex loss functions, the local regret are designed through the time window to smoothen the gradient of loss suffered \cite{hazan2017efficient}.
Due to the limitation of this work to the static environments,
the work of \cite{aydore2019dynamic} introduced the notion of dynamic local regret to investigate the non-stationary environment for nonconvex loss functions using rolling  weighted  average  of  past  gradients.

Nevertheless, online learning treats the entire process as a single task without any task-specific adaptation.
As such, neither Meta-Learning nor Online Learning procedure alone is desirable for the continual lifelong learning scheme.
To mitigate this deficiency, several methods have been proposed to merge these two approaches \cite{finn2019online,denevi2019learning},
in which the agent simultaneously utilizes past experiences in a sequential
setting so as to learn good priors, and also adapt quickly to the
current task at hand.
In a follow-up work, \cite{zhuang2020no} introduced a notion of local regret as the performance measure
to provide the first theoretical regret analysis for online meta-learning algorithms in the nonconvex setting.
While its definition of local regret makes sense for learning in the stationary environments, it is not appropriate for dynamic environments in which the properties of the predicted variable change over time in an arbitrary way due to concept drift \cite{lu2017dynamic}.
To alleviate this issue, we extend the regret analysis of \cite{zhuang2020no} to the non-stationary setting.

Choosing an ideal learning rate is pivotal issue in the nice performance of the first and second order optimization procedures.
Amongst variants of stochastic gradient descent (\textsc{Sgd}) methods, adaptive gradient methods that scale the gradient elementwise by some form
of averaging of the past gradients are particularly successful in increasing the convergence speed of the algorithms. \textsc{Adagrad} \cite{duchi2011adaptive} is the
first popular method in this area.
Subsequent adaptive methods, such as \textsc{Adadelta} \cite{zeiler2012adadelta}, \textsc{Rmsprop} \cite{tieleman2012lecture}, \textsc{Adam} \cite{kingma2014adam}, and \textsc{Dadam} \cite{nazari2019dadam}, preclude continual decaying of the learning rates in \textsc{Adagrad} by using the exponential moving averages of squared past gradients as the denominator of the adaptive learning rates in lieu of the arithmetic average.

Recently, a few attempts have been made on high probability results.
\cite{kakade2008generalization} demonstrated high probability bounds for \textsc{Pegasos} algorithm by virtue of Freeman's inequality.
A recent study by \cite{li2020high} indicated a high probability analysis for Delayed \textsc{Adagrad} with momentum applied to smooth nonconvex objective
functions.
Different from recently developed theoretical analysis of \cite{zhuang2020no}, which has been focused on bounds in expectation for online meta-learning framework,
we also propound high probability convergence rates to describe the performance of the proposed algorithm on single runs.

In this paper, we develop an online meta-learning algorithm which copes up with non-stationary environments where the task-environment changes dynamically.
Existing online meta-learning methods have been mostly analyzed \cite{finn2019online,zhuang2020no} driven by a static regret that may not be optimal in changing environments where data are evolving and the optimal decision is drifting over time.
To circumvent this issue, we accomplish dynamic regret analysis for online meta-learning algorithm in the nonconvex setting.
The first main result (Theorem~\ref{thm1}) sheds the light of dynamic regret bound of the proposed online meta-learning framework, in which the first prominent adaptive gradient method, namely, \textsc{Adagrad}, are used to adjust automatically the learning rate.
We establish that the regret bound is of explicit reliance on the total number of iterations $T$ and parameters of the learner.
This bound acquires a logarithmic rate under a mild regularity condition of adaptive gradient methods which matches
with that of existing online meta-learning algorithms in the static environment.
For the sake of generality, we extend these results to the commonly used adaptive gradient method, namely, \textsc{Adam} (Theorem~\ref{adam}).
These results postulate bounds that hold in expectation, arguably a weak guarantee in the sense that it does not rule out
the error of the algorithm from bearing large variance.
Besides this, it is often impossible to perform many runs of algorithm and select the best one.
Thereby, it is essential to be confident that the output of a single run of the algorithm is extremely likely to characterize the guaranteed convergence rate.
To this end, Theorems \ref{thm12} and \ref{adam2} display bounds of dynamic regret that hold with high probability for the online meta-learning framework using \textsc{Adagrad} and \textsc{Adam} optimizer, respectively.
To the best of our knowledge, it is the first high probability convergence guarantees for the online meta-learning framework in the nonconvex setting.

$\mathbf{Outline.}$
The remainder of the paper is organized as follows.
Section~\ref{sec1} expounds the concepts of the online meta-learning framework and adaptive gradient methods.
Section~\ref{sec2} states a detailed description of the proposed online meta-learning algorithm.
Section~\ref{sec3} establishes theoretical guarantees with a notion of dynamic local regret in the non-stationary and nonconvex setting.
Section~\ref{sec6} shows a high probability analysis of online meta-learning framework under assumptions on function and stochastic gradients.
Section~\ref{conclusion} presents some concluding remarks.
In the supplementary material, we provide proofs for our results.
\subsection{Notation}
Throughout, $\Bbb{R}_+$ and $\Bbb{R}^d$ denote the sets of nonnegative real numbers and real coordinate space of $d$ dimensions, respectively.
For any vectors $\mathrm{a},\mathrm{b}\in\Bbb{R}^d$, we use $\sqrt{\mathrm{a}}$ to denote element-wise square root, $\mathrm{a}^2$ to denote element-wise square, $\mathrm{a}/\mathrm{b}$ to denote element-wise division, $\max(\mathrm{a},\mathrm{b})$
to denote element-wise maximum and $ \langle \mathrm{a} , \mathrm{b} \rangle$ to denote the standard Euclidean inner product.
We use $\|\cdot\|$ to denote the $L_2$-norm.
For any positive integer $d$, we set $[d]\triangleq\{1,\ldots,d\}$. Further, for any vector $\mathrm{x}_t\in \Bbb{R}^d$, $x_{t,i}$ denotes its $i^{th}$ coordinate where $i\in[d]$.
Given a function $\ell: \reel^d \rightarrow \reel$, we note $\nabla \ell$ its gradient and $\nabla_i
\ell$ the $i^{th}$ component of the gradient.
We assume everywhere our loss function $\ell$
is bounded from below and denote the infimum by $\ell^*>-\infty$.
We also let $\mathbf{1}$ denote all-ones vector. We denote by $\E[\cdot]$ the expectation with respect to the underlying probability space.
If $a_n$ and $b_n$ are positive sequences, then $a_n = \mathcal{O}(b_n)$ means that $\limsup_n a_n/b_n < \infty$, whereas $a_n=\Omega(b_n)$ means that $\liminf_n a_n/b_n>0$.
We write $a_n=\Theta (b_n)$ if $a_n=\mathcal{O}(b_n)$ and $a_n=\Omega(b_n)$.
\section{Preliminaries }\label{sec1}
In this section, we briefly summarize concepts of online meta-learning setting as well as adaptive gradient methods and discuss some their important
properties. This section consists of two subsections.
\begin{algorithm}[t]
\caption{Online Meta-Learning Setup \cite{finn2019online}}\label{algo:mlol}
\begin{algorithmic}[1]
\STATE{\textbf{Input}: An initial meta-learner $\xx_1$, a loss function $\ell(\cdot)$, a local adapter $U(\cdot)$, an online learning algorithm $\mathcal{A}$}
\FOR{$t = 1,2,\ldots$}
\STATE{Encounter a new task: $\mathcal{T}_t$.}
\STATE{Receive training data for current task: $\mathcal{D}_t^{tr}$.}
\STATE{Adapt $\xx_t$ to current task: $\hat{\xx}_t = U(\xx_t, \mathcal{D}_t^{tr})$.}
\STATE{Receive test data for current task: $\mathcal{D}_t^{ts}$.}
\STATE{Suffer $\ell_t(\xx_t)\triangleq\ell(\hat{\xx}_t, \mathcal{D}^{ts}_t)=\E_{p,q\sim\mathcal{D}^{ts}_t}[\ell(\hat{\xx}_t, p;q)]$.}
\STATE{Update $\xx_{t+1} = \mathcal{A}(\xx_1,\ell_1(\xx_1),\ldots,\ell_t(\xx_t))$.}
\ENDFOR
\end{algorithmic}
\end{algorithm}
\subsection*{2.1 Online Meta-Learning Framework}\label{sec:background}
In Algorithm~\ref{algo:mlol}, each task is presented sequentially to a round, denoted by $t$.
Then, the update of this algorithm is executed as a bi-level learning: (i) inner level and (ii) outer level (meta-learner).

In the inner level, for each new task $\mathcal{T}_t$, we use training data $\mathcal{D}_t^{tr}$ corresponding to this task to adapt meta-learner $\xx_t$ to the current task by following some strategy $U(\cdot)$.
Accordingly, the task-specific parameter $\hat{\xx}_t$ is learned iteratively using $\mathcal{D}^{tr}_t$ based on the rule \cite{finn2019online}
$$\hat{\xx}_t \triangleq U(\xx_t,\mathcal{D}^{tr}_t)=\xx_t - \theta\nabla \E_{p,q\sim\mathcal{D}^{tr}_t}[\ell(\xx_t, p;q)],$$
where $\theta$ is the stepsize.
The object of the online meta-learning algorithm is to sequentially learn meta-learners that generate good task-specific parameters after adaptation.
\\
In the outer level, all task-specific adaptations gather to update the meta-learners by using the loss function.
To this end, the test data $\mathcal{D}_t^{ts}$ will be disclosed for assessing the performance of the adapted learner $\hat{\xx}_t$.
The loss suffered at this round $\ell_t(\xx_t)$ can then be embedded in an online learning algorithm $\mathcal{A}$ to update meta-learner $\xx_t$.
A meta-learner $\xx_t$ is maintained to preserve the prior knowledge learned from past rounds.

The original notion of  regret in \eqref{eq:reg1} enforces the learner to compete with a fixed learner across all tasks which is not meaningful as tasks are very different.
To resolve this issue, \cite{finn2019online} took into account regret of the form:

\begin{equation*}
\sum^T_{t=1}\ell(U(\xx_t, \mathcal{D}_t^{tr}), \mathcal{D}_t^{ts}) - \min_{\xx\in \mathcal{X}}\sum^T_{t=1}\ell(U(\xx, \mathcal{D}_t^{tr}), \mathcal{D}_t^{ts})~,
\label{eq:reg2}
\end{equation*}
which competes with any fixed meta-learner.
They set forth the Follow the Meta Leader algorithm, achieving a logarithmic regret under assumption that $\ell$ is strongly convex function.
The analysis of this work has been studied in the strongly convex case, while many problems of the current interest have a nonconvex nature.
In a subsequent work, \cite{zhuang2020no} developed an efficient algorithm for nonconvex online meta-learning in which the steps of the norm version of the adaptive stochastic gradient method (\textsc{Adagrad}-Norm) \cite{ward2018adagrad} are used to update the meta-learner at the outer level.
After briefly demonstrating why the regret of the form \eqref{eq:reg1} is not a feasible metric of performance to the nonconvex world, they assessed the performance via the notion of the local regret proposed by \cite{hazan2017efficient} as:
\begin{equation*}
 {SLR}_{w}(T)\triangleq
\sum^T_{t=1}\|\nabla F_{t,w}(\xx_t)\|^2~,\quad F_{t,w}(\xx_t)\triangleq \frac{1}{w} \sum^{w-1}_{r=0}  \ell_{t-r}(\xx_{t}),\quad \ell_i(\cdot)=0\,\,\, \text{for}\,\,\, i\le0,
\end{equation*}
which is an appropriate measure of stationarity (as opposed to optimality).
The motivation of using sliding-window in $F$, especially a large window,
follows from Theorem 2.7 in \cite{hazan2017efficient}.

On the other hand, the statistical properties of the predicted variable can be changed over time in an arbitrary way \cite{lu2017dynamic} by a phenomenon called \textit{concept drift}. This notion, too, has recently drawn much attention, mainly owing to the need for online classification of very large, constantly changing data streams \cite{bifet2013pitfalls,kosina2015very} etc.
The online meta-learning framework can be utilized to sequentially learn the initial parameters of a model such that optimizing from this initialization results in fast adaptation and generalization.
In the setting of non-stationary environment of the online meta-learning framework, the task $\mathcal{T}_t$ i) varies over time, and ii) there is no prior information as for the dynamics
of the task $\mathcal{T}_t$ which in turn the optimal initialization varies over time due to a changing environment.
Although non-stationary environment of the online meta-learning framework has been considered in some recent works \cite{al2017continuous,aghapour2020double,raviv2021meta},
its theoretical understanding in nonconvex setting is largely missing.

\subsection*{2.2 ADAGRAD AND ADAM FOR MINIMIZATION PROBLEMS}
We deal with a unified formulation of Adagrad~\cite{duchi2011adaptive} and Adam~\cite{kingma2014adam}, which adjusts the step size for every dimension in accordance with the geometry of the past gradients.
The main objective in these methods is to solve the following minimization problem:
\begin{equation*}
\min_{\xx\in\R^d}F(\xx)=\E_{\zeta\sim\mathcal{P}}f(\xx;\zeta),
\end{equation*}
where $\xx$ is the model parameter, and $\zeta$ is an random variable following distribution $\mathcal{P}$.
We assume we have $0<\beta_2\leq 1, 0\leq \beta_1<\beta_2$ and
a non negative sequence $\eta_{t+1}$. We define three
vectors $\mm_{t+1}, \up_{t+1}, \xx_{t+1}\in \Bbb{R}^d$ iteratively. For start points $\xx_1\in \Bbb{R}^d$, $\mm_1=0$, and $\up_{1}=0$, the update is \cite{defossez2020simple}
\begin{align*}
      \mm_{t+1}&=\beta_1\mm_{t}+\g_{t},\\
\up_{t+1}&=\beta_2\up_{t}+\g^2_{t},
\end{align*}
with $\g_t=\nabla f(\xx_t;\zeta_t)$ and updates
\begin{equation*}
  \xx_{t+1}=\xx_t-\eta_{t+1} \frac{\mm_{t+1}}{\sqrt{\epsilon+\up_{t+1}}},
\end{equation*}
with a small constant $\epsilon>0$ preventing division by zero, $\mm_{t+1}$ and $\up_{t+1}$ are estimates of the mean and variance of $\g$, respectively.
In practice $\epsilon$ is usually chosen sufficiently small.
Taking $\beta_1=0$, $\beta_2=1$ and $\eta_{t+1}=\eta$ gives the first known adaptive gradient method in machine learning, namely, \textsc{Adagrad}, which
is also reduced to plain stochastic gradient descent (\textsc{Sgd}) method when we set $\up_{t+1}=1$.
Comparing with the \textsc{Sgd} method, \textsc{Adagrad} dynamically interpolates knowledge of history gradients to
adaptively change the learning rate, thereby achieving significantly better performance when the gradients are sparse, or in general small.
Another state-of-the-art algorithm for training deep learning models is known as \textsc{Adam}~\cite{kingma2014adam,nazari2019adaptive}, which is a variant of the general class of \textsc{Adagrad}-type algorithms.
When $0<\beta_2<1$, $0\leq \beta_1< \beta_2$, and $ \eta_{t+1} = \eta (1 - \beta_1) \sqrt{(1-\beta_2^{t+1})/(1-\beta_2)}$, the algorithm becomes an algorithm close to \textsc{Adam}.
However, \textsc{Adam} would be exactly recovered by setting $\eta_{t+1} = \eta \frac{(1-\beta_1)}{(1-\beta_1^{t+1})}\sqrt{(1-\beta_2^{t+1})/(1-\beta_2)},$
which complicates the proof.
Recently, there has been a surge of interest to approach the analysis of these methods for nonconvex and weakly-convex optimization \cite{reddi2018adaptive,nazari2020adaptive}.
In particular, \cite{reddi2018adaptive} argued the convergence of Adam in certain nonconvex settings and verified the effect of the mini-batch size
in its convergence.
Later, \cite{nazari2020adaptive} concluded non-asymptotic rates of convergence of first and zeroth-order adaptive methods and their proximal variants for a reasonably broad class
of nonsmooth and nonconvex optimization problems.
More recently, \cite{defossez2020simple} established a simplified and unified proof of convergence for nonconvex \textsc{Adagrad} and \textsc{Adam}, improving the dependence of the iteration complexity on the momentum parameter.

\section{Algorithm}\label{sec2}
\begin{algorithm}[t]
\caption{Dynamic Time-Smoothed Adaptive Gradient (\textsc{Dts-Ag})}
\begin{algorithmic}[1]
\STATE{\textbf{Input}: Initialize $\xx_1\in\R^d$, $\mm_1=0$, $\up_1=0$, window size $w\geq 1$, number of iterations $T$, exponential smoothing parameter $\alpha\rightarrow 1^-$, decay parameters $\beta_2\in[0,1]$ and $0\leq\beta_1<\beta_2$, normalization parameter $W\triangleq \sum_{r=0}^{w-1}\alpha^r$, non negative sequence $\{ \eta_t\}_{t=1}^T$, $\epsilon>0$.}
\FOR{$t = 1,\ldots,T$}
\STATE {Uniformly randomly pick i.i.d samples $\{\xi_{t,t-w+1},\ldots,\xi_{t,t} \}$ according to the distribution $\mathcal{P}$.}
\STATE{Generate $\tilde{\nabla}S_{t,w,\alpha}(\xx_t) = \frac{1}{W}\sum^{w-1}_{r=0}\alpha^r\g_{t-r}(\xx_{t-r}, \xi_{t,t-r})$.}
\STATE{$\mm_{t+1}= \beta_{1} \mm_{t}+\tilde{\nabla}S_{t,w,\alpha}(\xx_t)$.}
\STATE{$\up_{t+1}= \beta_{2}\up_{t}+\big(\tilde{\nabla}S_{t,w,\alpha}(\xx_t)\big)^2$.}
\STATE{$\xx_{t+1} = \xx_{t}-\eta_{t+1}\frac{\mm_{t+1}}{\sqrt{\epsilon+\up_{t+1}}}$.}
\ENDFOR
\end{algorithmic}
\label{algo:adagrad}
\end{algorithm}
Here, we put forth to use the Dynamic Time-Smoothed Adaptive Gradient ($\textsc{Dts-Ag}$) method presented in Algorithm~\ref{algo:adagrad} as the online learning algorithm $\mathcal{A}$ in Algorithm~\ref{algo:mlol} .
We stress that the $\textsc{Dts-Ag}$ algorithm differs from the algorithm proposed in \cite{zhuang2020no} in several aspects:
first, the gradients of the loss functions at each time step are calculated at their corresponding parameters and average the past $w$, rather than to execute $w$ gradient calculations at their most recent parameter $(\xx_t,\xi_{t,t-r})$;
second, we exploit adaptive gradient methods that adapt a vector of per-coefficient stepsizes rather than a single stepsize depending on the norm of the gradient;
third, we make use of factor $\alpha$ which  gives  a higher  weight to the recent gradients  so as to control time weighting and track  a  dynamic  function;
fourth, our approach builds on a generalized version of the adaptive gradient methods \cite{defossez2020simple} that covers both \textsc{Adagrad} and \textsc{Adam} as special cases as opposed to \cite{zhuang2020no} which is largely restricted to using \textsc{Adagrad}-\textsc{Norm} method for achieving state-of-the-art performance.

In Algorithm~\ref{algo:adagrad}, we also need to stipulate the
stepsizes $\{\eta_{t+1}\}$, which will be discussed later.
Moreover, thanks to the randomness of $\mathcal{D}_t^{ts}$ of the whole test-set, we focus on the stochastic setting
where we can only access (unbiased) estimates of the true losses and gradients at each
round. In this setting, we assume that at each round $t$, each call to any stochastic gradient oracle $\g_j$, $j\in\{t-w+1,\ldots,t\}$, brings about an i.i.d.~random vector
$\g_j(\xx_t, \xi_{t,j})$.

We also assume that

\begin{assum}
\label{ass:sg}
For all $t\in [T]$, $j\in\{t-w+1,\ldots,t\}$ and random variable $ \xi_{t,j}\sim \mathcal{P}$, we have
\begin{enumerate}
\item [(i)]
The stochastic gradient $\g_j(\xx_t,\xi_{t,j})$ is unbiased, i.e.
$$\E_{\xi_{t,j}}\left[\g_j(\xx_t,\xi_{t,j})|\xi_{1:t-1}\right]
=\nabla \ell_j(\xx_t);$$
\item [(ii)]
All random samples $ \xi_{t,j}\sim \mathcal{P}$ are selected randomly and independently to each other, i.e. for $j\ne r$,
\begin{align*}
\E_{\xi_{t,j},\xi_{t,r}}[\langle \g_j(\xx_t,\xi_{t,j}),\ \g_r(\xx_t,\xi_{t,r})\rangle|\xi_{1:t-1}]
= \langle\E_{\xi_{t,j}}[\g_j(\xx_t,\xi_{t,j})|\xi_{1:t-1}],\ \E_{\xi_{t,r}}[\g_r(\xx_t,\xi_{t,r})|\xi_{1:t-1}]\rangle,
\end{align*}
\end{enumerate}
where $\xi_{1:t-1}=\{\xi_{1,1},\xi_{2,1},\xi_{2,2},\ldots,\xi_{t-1,t-w},\ldots,\xi_{t-1,t-1}\}$, and $\E_{\xi_{t,j}}[\mathrm{u}|\xi_{1:t-1}]$ stands for the conditional expectation of $\mathrm{u}$ with respect to $\xi_{1:t-1}$. Also note that $\g_j(\cdot)=0$ for $j\le0$.
\end{assum}

We make the following additional assumption.
\begin{assum}\label{ii}
 There is $\sigma>0$ such that for each $t\in [T]$, $j\in\{t-w+1,\ldots,t\}$ and random variable $ \xi_{t,j}\sim \mathcal{P}$ the variance of the stochastic gradient is bounded by
$$\E_{\xi_{t,j}}\left[\left\|\g_j(\xx_t,\xi_{t,j})-\nabla \ell_j(\xx_t)\right\|^2\vert\xi_{1:t-1}\right]
\le\sigma^2.$$
\end{assum}
If $\sigma=0$, the stochastic gradient $\g_j(\xx_t,\xi_{t,j})$ is identified as the exact gradient at point $\xx_t$, i.e., $\g_j(\xx_t,\xi_{t,j})=\nabla \ell_j(\xx_t)$.

\section{Convergence Analysis in Expectation}\label{sec3}
In this section, we assert our theoretical results and their consequences.
The proofs are given later in the supplementary material.
To evaluate the performance of online meta-learning algorithms in non-stationary environments in which the optimal meta-learners of each online loss function $\ell_{t}$  corresponding to task $\mathcal{T}_t$ can be drifted over tasks,
we exploit the notion of dynamic regret introduced in \cite{aydore2019dynamic} as
\begin{equation}\label{dynamic}
\begin{aligned}
& DLR_w(T)\triangleq
\sum^T_{t=1}\|\nabla S_{t,w,\alpha}(\xx_t)\|^2~.
\end{aligned}
\end{equation}
Here
\begin{equation}\label{def}
S_{t,w,\alpha}(\xx_t)\triangleq \frac{1}{W} \sum^{w-1}_{r=0} \alpha^r \ell_{t-r}(\xx_{t-r}),
\end{equation}
where $W\triangleq\sum^{w-1}_{r=0} \alpha^r$, and $\ell_t(\xx_t)=0$ for $t\le0$. This can be viewed as the exponential averaging of the gradients $\nabla \ell_{t-r}$ at their corresponding meta-learners $\xx_{t-r}$ in lieu of the most recent meta-learner $\xx_t$ over a window on each task $t\in[T]$, which thereby are adapted
for non-stationary environments.
Indeed, this regret quantifies the objective of predicting meta-learners with small exponential averaging of the gradients.
Our main results in this section (Theorems \ref{thm1} and \ref{adam}) prove, to the best of our knowledge, the dynamic regret bounds in expectation for online meta-learning Algorithm \ref{algo:mlol} in which the extremely popular algorithms \textsc{Adagrad} and \textsc{Adam} are used as the online learning algorithm $\mathcal{A}$.
Having stated our theorems, we show that our results recover previous rate of the online meta-learning algorithm in stationary setting \cite{zhuang2020no}, which is restricted only to \textsc{Adagrad}-\textsc{Norm} version of adaptive gradient methods.

Our theoretical analysis is based on several assumptions regarding $\ell$.
\begin{assum} \label{ass}
$\ell:\Bbb{R}^d \rightarrow\R$ is twice differentiable. Furthermore, for all $\mathrm{v},\mathrm{u}\in \Bbb{R}^d$, we assume
\begin{enumerate}
\item[(i)]$\ell$ is $L$-Lipschitz, i.e. $\|\ell(\mathrm{u}) - \ell(\mathrm{v})\|\le L\|\mathrm{u}-\mathrm{v}\|$~.
\item [(ii)]$\ell$ is $\gamma$-smooth, that is, $\ell$ is differentiable and its gradient is $\gamma$-Lipschitz, i.e. $\|\nabla \ell(\mathrm{u})-\nabla \ell(\mathrm{v})\| \leq \gamma \|\mathrm{u}-\mathrm{v}\|$~.\label{smooth}\\
\item[(iii)]$\ell$ is $H$-Hessian-Lipschitz, i.e. $\|\nabla^2\ell(\mathrm{u}) - \nabla^2\ell(\mathrm{v})\|\le H\|\mathrm{u}-\mathrm{v}\|.$~
\item[(iv)]$\ell$ is $D$-Bounded, i.e. $|\ell(\mathrm{u})|\le D.$
\end{enumerate}
\label{ass:loss}
\end{assum}
These assumptions are standard in online learning \cite{hazan2017efficient}. In view of Assumption~\ref{ass:loss} of $\ell$, we next give a result on $\ell_t$ from \cite{zhuang2020no}, which is vital to our analysis.

\begin{lem}\cite{zhuang2020no}\label{lm:property}
Let Assumption~\ref{ass:loss} holds. Then, $\ell_t$ is $D$-Bounded, $L'\triangleq((1+\theta \gamma)L)$-Lipschitz, and $\gamma'\triangleq(\theta LH + (1+\theta\gamma)^2\gamma)$-smooth.
\end{lem}
Henceforth, the symbol $\E_t$ represents expectation with respect to $\xi_{t, t-w+1},\ldots,\xi_{t, t}$ conditioned on $\xi_{1:t-1}$.
It is worth noting that each $\tilde{\nabla} S_{t, w, \alpha}(\xx_t)$ is a weighted average of $w$ independently sampled unbiased gradient estimates with a bounded variance $\sigma^2$. Hence, the following Lemma is made for the function $\tilde{\nabla}S_{t,w,\alpha}(\xx_t)$.
We leave proofs of the rest lemmas in the supplementary material.

\begin{lem}\label{p0}
Suppose  Assumptions~\ref{ass:sg} and \ref{ii} hold. Then for $\tilde{\nabla}S_{t,w,\alpha}(\xx_t)$ in the Algorithm $\textsc{Dts-Ag}$, we have
\begin{enumerate}
\item [(a)] $\E_{t}\left[\tilde{\nabla}S_{t,w,\alpha}(\xx_t)\right]
=\nabla S_{t,w,\alpha}(\xx_t),$\label{ass:unbiase}
\item [(b)]$\E_{t}\left[\left\|\tilde{\nabla}S_{t,w,\alpha}(\xx_t)-\nabla S_{t,w,\alpha}(\xx_t)\right\|^2\right]
\le \frac{\sigma^2(1-\alpha^{2w})}{W^2(1-\alpha^2)}\triangleq \mu,$
\end{enumerate}
where $S_{t,w,\alpha}(\xx_t)$ defined in \eqref{def}.
\end{lem}

The succeeding lemma presents a bound for the size of $\frac{\tilde{\nabla}_iS_{t,w,\alpha}(\xx_t)}{\sqrt{\epsilon+\upsilon_{t+1,i}}}$.

\begin{lem}\label{app:lemma:descent_lemmaa}
Suppose Assumptions~\ref{ass:sg} and \ref{ii} hold. Let $\tilde{\nabla}_iS_{t,w,\alpha}(\xx_t)$ and $\upsilon_{t+1,i}$ be the sequences defined in the Algorithm $\textsc{Dts-Ag}$. Then, for any $0<\alpha<1$, $w$, $\epsilon>0$ and $S_{t,w,\alpha}(\xx_t)$ in \eqref{def}, we have
\begin{align*}
\nonumber
 & \esp{  \sum_{i =1}^d \langle \nabla_i S_{t,w,\alpha}(\xx_t) , \frac{\tilde{\nabla}_iS_{t,w,\alpha}(\xx_t)}{\sqrt{\epsilon+\upsilon_{t+1,i}}} \rangle }
 \geq \sum_{i=1}^{d}\frac{(\nabla_i S_{t,w,\alpha}(\xx_t))^2}{2\sqrt{\epsilon+\tilde{\upsilon}_{t+1,i}}}
-2
\sqrt{\mu}\sum_{i=1}^{d}\E_t\left[\frac{(\tilde{\nabla}_iS_{t,w,\alpha}(\xx_t))^2}{\epsilon+\upsilon_{t+1,i}}\right],
\end{align*}
where $\tilde{\upsilon}_{t+1,i}\triangleq \upsilon_{t,i} + (\nabla_i S_{t,w,\alpha}(\xx_t))^2 + \mu$, and $\mu\triangleq\frac{\sigma^2(1-\alpha^{2w})}{W^2(1-\alpha^2)}$ for all $i\in [d]$.
\end{lem}
Equipped with these lemmas, we now state the main result.
\begin{thm}(\textsc{Adagrad})\label{thm1}
Suppose Assumptions~\ref{ass:sg}, \ref{ii} and \ref{ass:loss} hold. Let $\textsc{Dts-Ag}$ be the algorithm $\mathcal{A}$ in Algorithm~\ref{algo:mlol} with parameters $\beta_1=0$, $\beta_2=1$, $\eta_{t+1}=\eta$ with $\eta>0$ and $\alpha\rightarrow 1^-$.
Then, for any $\delta \in (0,1)$ and $S_{t,w,\alpha}(\xx_t)$ in \eqref{def}, with probability at least $1-\delta$,  the iterates $\xx_t$ satisfy the following bound
\begin{equation*}
\sum^T_{t=1}\esp{\|\nabla S_{t,w,\alpha}(\xx_t)\|^2}
\le \frac{4C\sqrt{\epsilon}}{\delta}+\frac{8C\sqrt{\zeta T}}{\delta^{3/2}}+\frac{48C^2}{\delta^2}~.
\end{equation*}
Here, $C \triangleq \varpi_1+\varpi_2d\ln\left(1 + \frac{2(\zeta+L'^2)T}{d\epsilon}\right)$, where
\begin{align*}
\varpi_1 \triangleq  \frac{4DT}{W\eta},\qquad
\varpi_2 \triangleq \frac{\eta\gamma'+4\sqrt{\zeta}}{2},\qquad \zeta \triangleq \frac{\sigma^2}{W}.
\end{align*}
\end{thm}
Note that, theoretical guarantee of Theorem \ref{thm1} is a bound in expectation over the randomness of stochastic gradients, and is therefore only on-average convergence guarantee.
\begin{cor}\label{corr1}
Under the same conditions stated in Theorem \ref{thm1}, using $w\in\Theta(T)$ and $\alpha\rightarrow 1^{-}$ yields a regret bound of order
\begin{equation*}
\sum^T_{t=1}\esp{\|\nabla S_{t,w,\alpha}(\xx_t)\|^2}\leq \mathcal{O}(\ln T).
\end{equation*}
\end{cor}

Corollary \ref{corr1} showed that the algorithm achieved a logarithmic bound on dynamic regret with respect to any parameters $\up_1, \eta>0, \epsilon>0$ with a choice of $w\in\Theta(T)$, which in turn recovers the result of \cite{zhuang2020no} on the online meta-learning in static environments. 
%Compared to this work, our bound is dependent on the variable dimension $d$.

In the sequel, we extend our results to the \textsc{Adam} optimizer, while no such result is considered in previous works.
To this end, we need the following lemma.

\begin{lem}\label{app:lemma:descent_lemma}
Suppose Assumptions~\ref{ass:sg}, \ref{ii} and \ref{ass} hold.
Let $m_{t+1,i}$ and $\upsilon_{t+1,i}$ be the sequences defined in the Algorithm $\textsc{Dts-Ag}$.
Then, for any $0<\alpha<1$, $w$, $\epsilon>0$, $0\leq \beta_1<\beta_2\leq 1$, $1\leq k\leq t$ and $S_{t,w,\alpha}(\xx_t)$ in \eqref{def}, we have
\begin{align*}
\nonumber
 & \esp{  \sum_{i =1}^d \langle \nabla_i S_{t,w,\alpha}(\xx_t) , \frac{m_{t+1,i}}{\sqrt{\epsilon+\upsilon_{t+1,i}}} \rangle }\geq
        \sum_{i=1}^d \sum_{k=0}^{t-1}\beta_1^k\frac{(\nabla_i S_{t+1-k,w,\alpha}(\xx_{t+1-k}) )^2}{2\sqrt{\epsilon+\tilde{\upsilon}_{t+1,i}}}
      \\\nonumber
 &\quad \quad -
      \frac{1}{\sqrt{1-\beta_1}}\esp{\sum_{k=0}^{t-1}
     \bbratio^k (\sqrt{k+1}+2\sqrt{\mu}) \|A_{t+1 -k}\|^2}
     \\\nonumber&\quad\quad-\frac{\eta_{t+1}^2 \gamma'^2}{4}\sqrt{1 - \beta_1} \esp{\sum_{l=1}^{t}\norm{B_{t+1-l}}^2}\sum_{k=l}^{t }\beta_1^k \sqrt{k}  -\sum_{k=0}^{t-1}\beta_1^k\frac{\sqrt{1-\beta_1}\vartheta_t}{2\sqrt{k+1}}.
\end{align*}
Here, $\tilde{\upsilon}_{t+1,i}\triangleq \beta_2\upsilon_{t,i} + (\nabla_i S_{t,w,\alpha}(\xx_t))^2 + \mu$, $\mu\triangleq\frac{\sigma^2(1-\alpha^{2w})}{W^2(1-\alpha^2)}$ for all $i\in [d]$, and
\begin{subequations}
\begin{align*}
 \vartheta_t \triangleq \frac{8L'^2 }{W^2} +\frac{2(1-\alpha^{w-2})\gamma'^2}{W^2(1-\alpha)} \sum_{r=1}^{w-1}  \alpha^{r-1} \|\eta_{t-r+2-k}B_{t-r+2-k} \|^2, \quad B_{t}\triangleq \frac{\mm_{t}}{\sqrt{\epsilon+\up_{t}}}, \quad A_{t}\triangleq \frac{\g_{t}}{\sqrt{\epsilon+\up_{t}}}.
\end{align*}
\end{subequations}
%\end{subequations}
\end{lem}

Having proven Lemma~\ref{app:lemma:descent_lemma}, we can now characterize convergence guarantees of the online meta-learning algorithm \ref{algo:mlol} in
full generality.

\begin{thm}\label{adam}(\textsc{Adam})
Suppose Assumptions \ref{ass:sg}, \ref{ii} and \ref{ass:loss} hold. Let $\textsc{Dts-Ag}$ be the algorithm $\mathcal{A}$ in Algorithm~\ref{algo:mlol} with parameters $\eta_{t+1} = \eta (1 - \beta_1) \sqrt{\frac{1-\beta_2^{t+1}}{1-\beta_2}}$ with $0<\beta_2<1$, $\eta>0$, $0<\beta_1<\beta_2$, and $\alpha \rightarrow 1^-$.
Furthermore, let $\sqrt{\sum_{r=0}^{t} \beta_2^r}\geq \frac{\varsigma}{\sqrt{1-\beta_2}}$ for some $\varsigma>0$ and $t\in [T]$.
Then, for any $\delta \in (0,1)$ and $S_{t,w,\alpha}(\xx_t)$ in \eqref{def}, with probability at least $1-\delta$,  the iterates $\xx_t$ satisfy the following bound
\begin{align*}
\sum^T_{t=1}\esp{\|\nabla S_{t,w,\alpha}(\xx_t)\|^2}
&\le
\frac{\sqrt{1-\beta_2}}{\varsigma\eta(1-\beta_1)}\Big(\frac{4C\sqrt{\epsilon}}{\delta}+\frac{8C\sqrt{\zeta T}}{\delta^{3/2}}\Big)+\frac{48(1-\beta_2)C^2}{\varsigma^2\eta^2(1-\beta_1)^2\delta^2}.
\end{align*}
Here, $C \triangleq \varpi_1 + \varpi_2 \left(d\ln\big(1 + \frac{2(\zeta+L'^2)}{d\epsilon ( 1- \beta_2)}\big) - T \ln(\beta_2)\right)$,
where
\begin{align}
\label{53}\varpi_1&\triangleq\frac{4D T}{W}+\frac{8T\eta (1-\beta_1)L'^2}{\beta_1\sqrt{(1-\beta_2)}W^2},\qquad \qquad \zeta \triangleq\frac{\sigma^2}{W},\\
    \varpi_2 &\triangleq \frac{ d \eta^2(1- \beta_1)\gamma'}{2(1 - \beta_2)(1 - \beta_1 / \beta_2)} +
     \frac{d \eta^3  \gamma'^2 \beta_1}{(1 - \beta_1 / \beta_2) (1 - \beta_2)^{3/2}}
    \nonumber \\&\qquad +
     \frac{ 2d \eta (1+\sqrt{\zeta}) \sqrt{1-\beta_1}}{(1 - \beta_1 / \beta_2)^{3/2}\sqrt{1 - \beta_2}}
     +\frac{2\eta^3(1-\beta_1)^2\gamma'^2}{\beta_1 (1-\beta_2)^{3/2}(1 - \beta_1 / \beta_2)}\label{6}.
\end{align}
\end{thm}

\begin{cor}\label{corr11}
Under the same conditions stated in Theorem \ref{adam}, using $\beta_2=1-1/T$, $\eta=\eta_1/\sqrt{T}$, $\beta_1/\beta_2\approx \beta_1$, $w\in\Theta(T)$, and $\alpha\rightarrow 1^{-}$ yields a regret bound of order
\begin{equation*}
\sum^T_{t=1}\esp{\|\nabla S_{t,w,\alpha}(\xx_t)\|^2}\leq \mathcal{O}(\ln T).
\end{equation*}
\end{cor}
The theorem indicates that when leveraging \textsc{Adam} optimizer,
the result of Theorem \ref{thm1} holds true.
\section{Convergence Analysis with High Probability}\label{sec6}
Theorems \ref{thm1} and \ref{adam} bound the expectation of a weighted average of $w$ gradients over the randomness of
stochastic gradients. Whilst these bounds can guarantee the average performance of a
large number of trials of the algorithm, they cannot preclude extremely bad solutions. Aside from,
towards practical applications, usually we only perform one single
run of the algorithm since that the training process may take long time.
Subsequently, it is crucial to acquire high
probability bounds which guarantee the performance of the algorithm on single runs. To circumvent
this difficulty, in this section, we in addition show high probability bounds of the convergence rate
for the online meta-learning algorithm. The following further assumptions are allowed for.

\begin{assum}\label{212}
	(\textit{Sub-Gaussian Noise}) For all $t\in [T]$ and $j\in\{t-w+1,\ldots,t\}$, the stochastic gradient satisfies
	$$\E_{\xi_{t,j}} \left[\exp\left(\max_{1\leq t\leq T}\|\nabla \ell_j(\xx_t) - \g_j(\xx_t,\xi_{t,j})\|^2/\kappa^2\right)| \xi_{1:t-1}\right]\leq \exp(1),\qquad \forall \xi_{t,j}\sim \mathcal{P}.$$
\end{assum}

\cite{nemirovski2009robust} and \cite{harvey2019tight} and \cite{li2020high} gave the high probability convergence guarantees under the
Assumption \ref{212}.
Intuitively, it results in that the tails of the noise distribution are dominated by tails of a Gaussian distribution. \

\begin{lem}\label{4543}
Suppose  Assumptions~\ref{ass:sg} and \ref{212} hold. Then, for $\tilde{\nabla}S_{t,w,\alpha}(\xx_t)$ in the Algorithm $\textsc{Dts-Ag}$ any $\delta \in (0,1)$ and $S_{t,w,\alpha}(\xx_t)$ in \eqref{def}, with probability at
least $1-\delta$, we have
\begin{enumerate}
\item [(a)] $\E_{t}\left[\tilde{\nabla}S_{t,w,\alpha}(\xx_t)\right]
=\nabla S_{t,w,\alpha}(\xx_t),$\label{ass:unbiase}
\item [(b)]$ \max_{1\leq t\leq T}\left\|\tilde{\nabla}S_{t,w,\alpha}(\xx_t)-\nabla S_{t,w,\alpha}(\xx_t)\right\|^2
\le   \kappa^2\ln \frac{\exp\left(\frac{w\sum^{w-1}_{r=0}\alpha^{2r}}{W^2}\right)}{\delta} \triangleq \bar{\mu}.$
\end{enumerate}
\end{lem}

\begin{lem}\label{app:lemma:descent_lemmaa2}
Suppose Assumptions~\ref{ass:sg} and \ref{212} hold. Let $\tilde{\nabla}_iS_{t,w,\alpha}(\xx_t)$ and $\upsilon_{t+1,i}$ be the sequences defined in the Algorithm $\textsc{Dts-Ag}$. Then, for any $\delta \in (0,1)$, $0<\alpha<1$, $w$, $\epsilon>0$ and $S_{t,w,\alpha}(\xx_t)$ in \eqref{def}, with probability at least $1-\delta$, we have
\begin{align*}
\nonumber
  \sum_{t=1}^{T} \sum_{i =1}^d \langle \nabla_i S_{t,w,\alpha}(\xx_t) , \frac{\tilde{\nabla}_iS_{t,w,\alpha}(\xx_t)}{\sqrt{\epsilon+\upsilon_{t+1,i}}} \rangle
& \geq \sum_{i=1}^{d}\sum_{t=1}^{T}\frac{(\nabla_i S_{t,w,\alpha}(\xx_t))^2}{4\sqrt{\epsilon+\tilde{\upsilon}_{t+1,i}}}
-\frac{2\sqrt{\bar{ \mu}}}{\sqrt{W}}
\sum_{i=1}^{d}\sum_{t=1}^{T}\frac{(\tilde{\nabla}_iS_{t,w,\alpha}(\xx_t))^2}{\epsilon+\upsilon_{t+1,i}}\\
 &\quad-\frac{3(1-\alpha^w)^2\kappa^2}{W^2(1-\alpha)^2\sqrt{\epsilon}}\ln \frac{1}{\delta},
\end{align*}
where $\bar{\mu}=\kappa^2\ln \frac{\exp\left(\frac{w\sum^{w-1}_{r=0}\alpha^{2r}}{W^2}\right)}{\delta}$, and
$\tilde{\upsilon}_{t+1,i}\triangleq \frac{1}{W}\big(\upsilon_{t,i} + (\nabla_i S_{t,w,\alpha}(\xx_t))^2 \big)+ \hat{\mu}_{T,i}$,\\
$\hat{\mu}_{T,i}\triangleq  \frac{1}{W}\left(\tilde{\nabla}_i S_{T,w,\alpha}(\xx_T)-\nabla_i S_{T,w,\alpha}(\xx_T)\right)^2$ for all $i\in [d]$.
\end{lem}

With the results of Lemmas \ref{4543} and \ref{app:lemma:descent_lemmaa2} in hand, we can now prove for the first time convergence of the online meta-learning framework in high probability.
\begin{thm}(\textsc{Adagrad})\label{thm12}
Suppose Assumptions~\ref{ass:sg}, \ref{ass:loss} and \ref{212} hold. Let $\textsc{Dts-Ag}$ be the algorithm $\mathcal{A}$ in Algorithm~\ref{algo:mlol} with parameters $\beta_1=0$, $\beta_2=1$, $\eta_{t+1}=\eta$ with $\eta>0$ and $\alpha\rightarrow 1^-$.
Then, for any $\delta \in (0,1)$ and $S_{t,w,\alpha}(\xx_t)$ in \eqref{def}, with probability at least $1-\delta$, the iterates $\xx_t$ satisfy the following bound
%Then, feeding Algorithm~\ref{algo:adagrad} into Algorithm~\ref{algo:mlol} gives the following upper bound of $DLR_w(T)$, with probability $1-\delta$:
\begin{equation*}
\sum^T_{t=1}\|\nabla S_{t,w,\alpha}(\xx_t)\|^2
\le 4C\sqrt{\epsilon}+4C\sqrt{\frac{2T\zeta}{W} }+\frac{48C^2}{W}~.
\end{equation*}
Here,
\begin{align*}
C &\triangleq \varpi_1
+\varpi_2d\ln\left(1 + \frac{2(\zeta+L'^2)T}{d\epsilon}\right)+\frac{3\kappa^2}{\sqrt{\epsilon+\upsilon_{1,i}}}\ln \frac{1}{\delta},\\
\varpi_1 &\triangleq  \frac{4DT}{W\eta},\qquad
\varpi_2 \triangleq \frac{\eta\gamma'}{2}+\frac{2\sqrt{\zeta}}{\sqrt{W}},\qquad \zeta \triangleq \kappa^2\ln \frac{e}{\delta}.
\end{align*}
\end{thm}

Similar to the discussion in Theorem \ref{thm1}, we can choose $w\in\Theta(T)$, to achieve a logarithmic regret bound for dynamic regret of the online meta-learning algorithm with respect to any parameters $\up_1, \eta>0, \epsilon>0$.

\begin{cor}\label{corrh}
Under the same conditions stated in Theorem \ref{thm12}, using $w\in\Theta(T)$ yields a regret bound of order
\begin{equation*}
\sum^T_{t=1} \|\nabla S_{t,w,\alpha}(\xx_t)\|^2 \leq \mathcal{O}(\ln T).
\end{equation*}
\end{cor}

We next move on to prove high probability bound on the convergence rate of Algorithm~\ref{algo:mlol} in full generality. To this end, we will need to establish the following lemma.
\begin{lem}\label{app:lemma:descent_lemma2}
Suppose Assumptions~\ref{ass:sg}, \ref{ass} and \ref{212} hold.
Let $m_{t+1,i}$ and $\upsilon_{t+1,i}$ be the sequences defined in the Algorithm $\textsc{Dts-Ag}$.
Then, for any $\delta \in (0,1)$, $0<\alpha<1$, $w$, $\epsilon>0$, $0\leq \beta_1<\beta_2\leq 1$, $1\leq k\leq t$, and $S_{t,w,\alpha}(\xx_t)$ in \eqref{def}, with probability at least $1-\delta$, we have
\begin{align*}
\nonumber
 &   \sum_{i =1}^d \sum_{t=1}^{T} \langle \nabla_i S_{t,w,\alpha}(\xx_t) , \frac{m_{t+1,i}}{\sqrt{\epsilon+\upsilon_{t+1,i}}} \rangle \geq
        \sum_{i=1}^d \sum_{t=1}^{T} \sum_{k=0}^{t-1}\beta_1^k\frac{(\nabla_i S_{t+1-k,w,\alpha}(\xx_{t+1-k}) )^2}{4\sqrt{\epsilon+\tilde{\upsilon}_{t+1,i}}}
      \\\nonumber
 &\qquad  -\sum_{t=1}^{T}
     \sum_{k=0}^{t-1} \frac{1}{\sqrt{1-\beta_1}}
     \bbratio^k \left(\sqrt{k+1}+\frac{2\sqrt{\bar \mu} }{\sqrt{W}}\right) \|A_{t+1 -k}\|^2
     \\\nonumber&\qquad-\sum_{t=1}^{T}\frac{\eta_{t+1}^2 \gamma'^2}{4}\sqrt{1 - \beta_1} \sum_{l=1}^{t}\norm{B_{t+1-l}}^2\sum_{k=l}^{t }\beta_1^k \sqrt{k}-\sum_{t=1}^{T}\sum_{k=0}^{t-1}\beta_1^k\frac{\sqrt{1-\beta_1}\vartheta_t}{2\sqrt{k+1}}
     \\&\qquad-\frac{3(1-\alpha^w)^2\kappa^2}{W^2\beta_1^k(1-\alpha)^2\sqrt{\epsilon}}\ln \frac{1}{\delta}.
\end{align*}
Here, $\bar{\mu}=\kappa^2\ln \frac{\exp\left(\frac{w\sum^{w-1}_{r=0}\alpha^{2r}}{W^2}\right)}{\delta}$,
$\tilde{\upsilon}_{t+1,i}\triangleq \frac{1}{W}\big(\beta_2\upsilon_{t,i} + (\nabla_i S_{t,w,\alpha}(\xx_t))^2 \big)+ \hat{\mu}_{T,i}$,\\
 $\hat{\mu}_{T,i}\triangleq \frac{1}{W}\left(\tilde{\nabla}_i S_{T,w,\alpha}(\xx_T)-\nabla_i S_{T,w,\alpha}(\xx_T)\right)^2$ for all $i\in [d]$, and
\begin{subequations}
\begin{align*}
\vartheta_t \triangleq \frac{8L'^2 }{W^2} +\frac{2(1-\alpha^{w-2})\gamma'^2}{W^2(1-\alpha)} \sum_{r=1}^{w-1}  \alpha^{r-1} \|\eta_{t-r+2-k}B_{t-r+2-k} \|^2,\quad
B_{t}\triangleq \frac{\mm_{t}}{\sqrt{\epsilon+\up_{t}}},\quad
A_{t}\triangleq \frac{\g_{t}}{\sqrt{\epsilon+\up_{t}}}.
\end{align*}
\end{subequations}
\end{lem}

\begin{thm}\label{adam2}(\textsc{Adam})
Suppose Assumptions~\ref{ass:sg}, \ref{ass} and \ref{212} hold. Let $\textsc{Dts-Ag}$ be the algorithm $\mathcal{A}$ in Algorithm~\ref{algo:mlol} with parameters $\eta_{t+1} = \eta (1 - \beta_1) \sqrt{\frac{1-\beta_2^{t+1}}{1-\beta_2}}$ with $0<\beta_2<1$, $\eta>0$, $0<\beta_1<\beta_2$, and $\alpha \rightarrow 1^-$.
Furthermore, let $\sqrt{\sum_{r=0}^{t} \beta_2^r}\geq \frac{\varsigma}{\sqrt{1-\beta_2}}$ for some $\varsigma>0$ and $t\in [T]$.
Then, for any $\delta \in (0,1)$ and $S_{t,w,\alpha}(\xx_t)$ in \eqref{def}, with probability at least $1-\delta$,  the iterates $\xx_t$ satisfy the following bound
\begin{align*}
\sum^T_{t=1}\|\nabla S_{t,w,\alpha}(\xx_t)\|^2
&\le
\frac{4\sqrt{1-\beta_2}C}{\varsigma\eta(1-\beta_1)}\Big(\sqrt{\epsilon}+\sqrt{\frac{2T\zeta}{W}}\Big)+\frac{48(1-\beta_2)C^2}{W\varsigma^2\eta^2(1-\beta_1)^2}.
\end{align*}
Here, $$C \triangleq \varpi_1 + \varpi_2 \left(d\ln\big( 1 +\frac{2(\zeta+L'^2)}{d\epsilon(1-\beta_2)} \big) - T \ln(\beta_2)\right)+\varpi_3,$$
where
\begin{align}
\label{w1}\varpi_1&\triangleq\frac{4D T}{W}+\frac{8T\eta (1-\beta_1)L'^2}{\beta_1\sqrt{(1-\beta_2)}W^2},\qquad \qquad
\zeta \triangleq \kappa^2\ln \frac{e}{\delta},\\
  \nonumber  \varpi_2 &\triangleq \frac{ d \eta^2(1- \beta_1)\gamma'}{2(1 - \beta_2)(1 - \beta_1 / \beta_2)} +
     \frac{d \eta^3  \gamma'^2 \beta_1}{(1 - \beta_1 / \beta_2) (1 - \beta_2)^{3/2}}
     \\&\qquad +
     \frac{ 2d \eta (1+\frac{\sqrt{\zeta}}{\sqrt{W}}) \sqrt{1-\beta_1}}{(1 - \beta_1 / \beta_2)^{3/2}\sqrt{1 - \beta_2}}
     +\frac{2\eta^3(1-\beta_1)^2\gamma'^2}{\beta_1 (1-\beta_2)^{3/2}(1 - \beta_1 / \beta_2)},\label{w2}\\
     \varpi_3 &\triangleq \frac{3\eta(1-\beta_1)\kappa^2}{W^2\beta_1^T\sqrt{1-\beta_2}\sqrt{\epsilon}}\ln \frac{1}{\delta}.\label{w3}
\end{align}
\end{thm}

The theorem reports that when using \textsc{Adam} optimizer, the result of Theorem \ref{thm12} maintains
true.

\begin{cor}\label{corr112}
Under the same conditions stated in Theorem \ref{adam2}, using $\beta_2=1-1/T$, $\eta=\eta_1/\sqrt{T}$, $\beta_1/\beta_2\approx \beta_1$, $w\in\Theta(T)$, and $\alpha\rightarrow 1^{-}$ yields a regret bound of order
\begin{equation*}
\sum^T_{t=1} \|\nabla S_{t,w,\alpha}(\xx_t)\|^2\leq \mathcal{O}(\ln T).
\end{equation*}
\end{cor}

\section{Conclusion}\label{conclusion}
In this paper, we extended the static regret analysis of the online meta-learning framework to the non-stationary environments in the nonconvex setting.
We also propose to use  a generalized version of the adaptive gradient methods that covers both \textsc{Adagrad} and \textsc{Adam} to learn meta-learners in the outer level of the online meta-learning framework. Our approach enjoyed a logarithmic dynamic local regret under some mild conditions.
What's more, we proved high probability bounds on the convergence rates of the online meta-learning framework, which have not been concluded before.

\section*{Acknowledgements}\label{conclusion}
We extend our thanks to Davoud Ataee Tarzanagh for useful discussions on this work.

\bibliography{ref}
\bibliographystyle{ieeetr}

\appendix
\newpage
$$\LARGE{\mathbf{\text{Supplementary Material}}}$$
This supplementary material establishes the missing proofs in the paper.

\subsection*{Details of Section \ref{sec3}}

\subsubsection*{PROOF OF LEMMA~\ref{p0}}

\begin{lem*}
Suppose  Assumptions~\ref{ass:sg} and \ref{ii} hold. Then for $\tilde{\nabla}S_{t,w,\alpha}(\xx_t)$ in the Algorithm $\textsc{Dts-Ag}$, we have
\begin{enumerate}
\item [(a)] $\E_{t}\left[\tilde{\nabla}S_{t,w,\alpha}(\xx_t)\right]
=\nabla S_{t,w,\alpha}(\xx_t),$\label{ass:unbiase}
\item [(b)]$\E_{t}\left[\left\|\tilde{\nabla}S_{t,w,\alpha}(\xx_t)-\nabla S_{t,w,\alpha}(\xx_t)\right\|^2\right]
\le \frac{\sigma^2(1-\alpha^{2w})}{W^2(1-\alpha^2)}\triangleq \mu,$
\end{enumerate}
where $S_{t,w,\alpha}(\xx_t)$ is defined as in \eqref{def}.
\end{lem*}
\begin{proof}
The techniques used in this proof are similar to the ones in \cite{aydore2019dynamic}.
We first prove (a).
Recall that $\E_t$ denotes conditioning on $\xi_{1:t-1}$ and take expectation w.r.t.~$\xi_{t, t-w+1},\ldots,\xi_{t, t}$.
In view of Assumption~\ref{ass:sg}(i) we assume $\E_{\xi_{t,r}}\left[\g_r(\xx_t,\xi_{t,r})|\xi_{1:t-1}\right]
=\nabla \ell_r(\xx_t)$ for $r\in\{t-w+1,\ldots,t\}$, the linearity of expectation immediately gives us $\E_{t}\left[\tilde{\nabla}S_{t,w,\alpha}(\xx_t)\right]
=\nabla S_{t,w,\alpha}(\xx_t)$.

For part (b), expanding $\E_t\left[\left\|\tilde{\nabla} S_{t,w,\alpha}(\xx_t)-\nabla S_{t,w,\alpha}(\xx_t)\right\|^2\right]$, we have
\begin{align}\label{nnn}
\nonumber &\frac1{W^2}\E_t\left[\left\|\sum^{w-1}_{r=0}\alpha^r\g_{t-r}(\xx_{t-r},\xi_{t,t-r}) -\alpha^r \nabla\ell_{t-r}(\xx_{t-r})\right\|^2\right]\\
\nonumber &=
\frac1{W^2}\sum^{w-1}_{r=0}\sum^{w-1}_{j=0}\E_t\left[\langle \alpha^r\g_{t-r}(\xx_{t-r},\xi_{t,t-r}) -\alpha^r \nabla\ell_{t-r}(\xx_{t-r}) ,\alpha^j\g_{t-j}(\xx_{t-j},\xi_{t,t-j}) -\alpha^j \nabla\ell_{t-j}(\xx_{t-j})\rangle\right]\\
\nonumber &=\frac1{W^2}\sum^{w-1}_{r=0}\E_t\left[\|\alpha^r\g_{t-r}(\xx_{t-r},\xi_{t,t-r}) -\alpha^r \nabla\ell_{t-r}(\xx_{t-r})\|^2\right]
\\&\quad+
\frac1{W^2}\sum^{w-1}_{r=0}\sum_{j\ne r}\E_t\left[\langle \alpha^r\g_{t-r}(\xx_{t-r},\xi_{t,t-r}) -\alpha^r \nabla\ell_{t-r}(\xx_{t-r}) ,\alpha^j\g_{t-j}(\xx_{t-j},\xi_{t,t-j}) -\alpha^j \nabla\ell_{t-j}(\xx_{t-j})\rangle\right]~.
\end{align}
The first term of RHS of \eqref{nnn} is upper bounded by $\frac{\sigma^2(1-\alpha^{2w})}{W^2(1-\alpha^2)}$ under Assumption~\ref{ii}.

Then for the second term of RHS of \eqref{nnn},
according to the Mutual Independence assumption (namely Assumption~\ref{ass:sg}(ii)), we have
\begin{align*}
&\E_t\left[\langle \alpha^r\g_{t-r}(\xx_{t-r},\xi_{t,t-r}) -\alpha^r \nabla\ell_{t-r}(\xx_{t-r}),\  \alpha^j\g_{t-j}(\xx_{t-j},\xi_{t,t-j}) - \alpha^j \nabla\ell_{t-j}(\xx_{t-j})\rangle\right]\\
&\quad=
\langle\E_t\left[\alpha^r\g_{t-r}(\xx_{t-r},\xi_{t,t-r}) -\alpha^r \nabla\ell_{t-r}(\xx_{t-r})\right],\  \E_t\left[\alpha^j\g_{t-j}(\xx_{t-j},\xi_{t,t-j}) - \alpha^j \nabla\ell_{t-j}(\xx_{t-j})\rangle\right]~,
\end{align*}
which is equal to 0, due to Assumption~\ref{ass:sg}(i). Plugging these two results back into \eqref{nnn}, proves part (b) of this lemma.

\end{proof}
\subsubsection*{ PROOF OF LEMMA~\ref{app:lemma:descent_lemmaa}}
\begin{lem*}
Suppose Assumptions~\ref{ass:sg} and \ref{ii} hold. Let $\tilde{\nabla}_iS_{t,w,\alpha}(\xx_t)$ and $\upsilon_{t+1,i}$ be the sequences defined in the Algorithm $\textsc{Dts-Ag}$. Then, for any $0<\alpha<1$, $w$, $\epsilon>0$ and $S_{t,w,\alpha}(\xx_t)$ in \eqref{def}, we have

\begin{align*}
\nonumber
 & \esp{  \sum_{i =1}^d \langle \nabla_i S_{t,w,\alpha}(\xx_t) , \frac{\tilde{\nabla}_iS_{t,w,\alpha}(\xx_t)}{\sqrt{\epsilon+\upsilon_{t+1,i}}} \rangle }
 \geq \sum_{i=1}^{d}\frac{(\nabla_i S_{t,w,\alpha}(\xx_t))^2}{2\sqrt{\epsilon+\tilde{\upsilon}_{t+1,i}}}
-2
\sqrt{\mu}\sum_{i=1}^{d}\E_t\left[\frac{(\tilde{\nabla}_iS_{t,w,\alpha}(\xx_t))^2}{\epsilon+\upsilon_{t+1,i}}\right],
\end{align*}
where $\tilde{\upsilon}_{t+1,i}\triangleq \upsilon_{t,i} + (\nabla_i S_{t,w,\alpha}(\xx_t))^2 + \mu$, and $\mu\triangleq\frac{\sigma^2(1-\alpha^{2w})}{W^2(1-\alpha^2)}$ for all $i\in [d]$.
\end{lem*}

\begin{proof}
We use LHS of Lemma \ref{p0}(a) and obtain that:
\begin{align}\label{88}
 &-\esp{  \sum_{i =1}^d \langle \nabla_i S_{t,w,\alpha}(\xx_t) , \frac{\tilde{\nabla}_iS_{t,w,\alpha}(\xx_t)}{\sqrt{\epsilon+\upsilon_{t+1,i}}} \rangle }=\sum_{i=1}^{d}\Big(-\frac{(\nabla_i S_{t,w,\alpha}(\xx_t))^2}{\sqrt{\epsilon+\tilde{\upsilon}_{t+1,i}}}+I_1\Big),
\end{align}
where
\begin{align*}
I_1&=\E_t[(\frac1{\sqrt{\epsilon+\tilde{\upsilon}_{t+1,i}}}
-
\frac{1}{\sqrt{\epsilon+\upsilon_{t+1,i}}}) \langle \nabla_i S_{t,w,\alpha}(\xx_t), \tilde{\nabla}_iS_{t,w,\alpha}(\xx_t)\rangle].
\end{align*}

Next, we need to upper bound the term $I_1$. To this end, we observe that from the definition of $\upsilon_{t+1}$ and $\tilde{\upsilon}_{t+1}$, we have
\begin{align*}
\left|\frac{1}{\sqrt{\epsilon+\tilde{\upsilon}_{t+1,i}}}
-
\frac{1}{\sqrt{\epsilon+\upsilon_{t+1,i}}}\right|&=
\frac{\left|(\tilde{\nabla_i}S_{t,w,\alpha}(\xx_t))^2 - (\nabla_i S_{t,w,\alpha}(\xx_t))^2 - \mu\right|}{\sqrt{\epsilon+\upsilon_{t+1,i}}\sqrt{\epsilon+\tilde{\upsilon}_{t+1,i}}\left(\sqrt{\epsilon+\upsilon_{t+1,i}}+\sqrt{\epsilon+\tilde{\upsilon}_{t+1,i}}\right)}\\[0.5em]
&\le\frac{\left|\tilde{\nabla_i}S_{t,w,\alpha}(\xx_t) - \nabla_i S_{t,w,\alpha}(\xx_t)\right|}{\sqrt{\epsilon+\upsilon_{t+1,i}}\sqrt{\epsilon+\tilde{\upsilon}_{t+1,i}}}
+
\frac{ \sqrt{\mu}}{\sqrt{\epsilon+\upsilon_{t+1,i}}\sqrt{\epsilon+\tilde{\upsilon}_{t+1,i}}}~,
\end{align*}
for all $i\in [d]$. Utilizing this result together with the Jensen's inequality on $|\cdot|$ which is a convex function, we conclude that

\begin{align}\label{eq:second1}
\nonumber |I_1|&\leq{{\frac{\E_t\left[\left|\tilde{\nabla}_iS_{t,w,\alpha}(\xx_t) - \nabla_i S_{t,w,\alpha}(\xx_t)\right|(\tilde{\nabla}_iS_{t,w,\alpha}(\xx_t))(\nabla_i S_{t,w,\alpha}(\xx_t))\right]}{\sqrt{\epsilon+\upsilon_{t+1,i}}\sqrt{\epsilon+\tilde{\upsilon}_{t+1,i}}}}}
 \\ &\quad+
\frac{\E_t\left[(\tilde{\nabla}_iS_{t,w,\alpha}(\xx_t))(\nabla_i S_{t,w,\alpha}(\xx_t)) \sqrt{\mu}\right]}{\sqrt{\epsilon+\upsilon_{t+1,i}}\sqrt{\epsilon+\tilde{\upsilon}_{t+1,i}}}~.
\end{align}
By regarding the first term of \eqref{eq:second1} and applying inequality $ab\le\frac{\hslash}{2}a^2 + \frac{1}{2\hslash}b^2$ with $\hslash= \frac{2\mu}{\sqrt{\epsilon+\tilde{\upsilon}_{t+1,i}}}$, $a=\frac{|\tilde{\nabla}_iS_{t,w,\alpha}(\xx_t)|}{\sqrt{\epsilon+\upsilon_{t+1,i}}}$, we obtain an upper bound as
\begin{equation}\label{v}
\sum_{i=1}^{d}\frac{(\nabla_i S_{t,w,\alpha}(\xx_t))^2}{4\sqrt{\epsilon+\tilde{\upsilon}_{t+1,i}}}
+
\sqrt{\mu}\sum_{i=1}^{d}\E_t\left[\frac{(\tilde{\nabla}_iS_{t,w,\alpha}(\xx_t))^2}{\epsilon+\upsilon_{t+1,i}}\right]~,
\end{equation}
where we used that $\left|\|\mathrm{u}\| - \|\mathrm{v}\|\right|\le\|\mathrm{u}- \mathrm{v}\|$ holds for $\forall \mathrm{u},\mathrm{v}\in\R^d$.
Analogously, using $ab\le\frac{\hslash}{2}a^2 + \frac{1}{2\hslash}b^2$ but with $\hslash = \frac{2}{\sqrt{\epsilon+\tilde{\upsilon}_{t+1,i}}}$, $a=\frac{|\tilde{\nabla}_iS_{t,w,\alpha}(\xx_t)|\sqrt{\mu}}{\sqrt{\epsilon+\upsilon_{t+1,i}}}$, yields an upper bound on the second term of \eqref{eq:second1} by
\begin{equation}\label{b}
\sum_{i=1}^{d}\frac{(\nabla_i S_{t,w,\alpha}(\xx_t))^2}{4\sqrt{\epsilon+\tilde{\upsilon}_{t+1,i}}}
+
\sqrt{\mu}\sum_{i=1}^{d}\E_t\left[\frac{(\tilde{\nabla}_iS_{t,w,\alpha}(\xx_t))^2}{\epsilon+\upsilon_{t+1,i}}\right]~.
\end{equation}
Then, by plugging the above bounds in \eqref{v} and \eqref{b} into equation \eqref{eq:second1} we get
\begin{equation*}
  |I_1|\leq \sum_{i=1}^{d}\frac{(\nabla_i S_{t,w,\alpha}(\xx_t))^2}{2\sqrt{\epsilon+\tilde{\upsilon}_{t+1,i}}}
+2
\sqrt{\mu}\sum_{i=1}^{d}\E_t\left[\frac{(\tilde{\nabla}_iS_{t,w,\alpha}(\xx_t))^2}{\epsilon+\upsilon_{t+1,i}}\right],
\end{equation*}
which implies
\begin{equation}\label{dcc}
  I_1\leq \sum_{i=1}^{d}\frac{(\nabla_i S_{t,w,\alpha}(\xx_t))^2}{2\sqrt{\epsilon+\tilde{\upsilon}_{t+1,i}}}
+2
\sqrt{\mu}\sum_{i=1}^{d}\E_t\left[\frac{(\tilde{\nabla}_iS_{t,w,\alpha}(\xx_t))^2}{\epsilon+\upsilon_{t+1,i}}\right].
\end{equation}
After plugging \eqref{dcc} into \eqref{88}, we get
\begin{align*}
\nonumber-\esp{  \sum_{i =1}^d \langle \nabla_i S_{t,w,\alpha}(\xx_t) , \frac{\tilde{\nabla}_iS_{t,w,\alpha}(\xx_t)}{\sqrt{\epsilon+\upsilon_{t+1,i}}} \rangle }&\leq -\sum_{i=1}^{d}\frac{(\nabla_i S_{t,w,\alpha}(\xx_t))^2}{\sqrt{\epsilon+\tilde{\upsilon}_{t+1,i}}}
+\sum_{i=1}^{d}\frac{(\nabla_i S_{t,w,\alpha}(\xx_t))^2}{2\sqrt{\epsilon+\tilde{\upsilon}_{t+1,i}}}
\\&\quad+2
\sqrt{\mu}\sum_{i=1}^{d}\E_t\left[\frac{(\tilde{\nabla}_iS_{t,w,\alpha}(\xx_t))^2}{\epsilon+\upsilon_{t+1,i}}\right],
\end{align*}
which finishes the proof of the claim.
\end{proof}

\subsubsection*{ PROOF OF THEOREM~\ref{thm1}}
\begin{thm*}(\textsc{Adagrad})
Suppose Assumptions~\ref{ass:sg}, \ref{ii} and \ref{ass:loss} hold. Let $\textsc{Dts-Ag}$ be the algorithm $\mathcal{A}$ in Algorithm~\ref{algo:mlol} with parameters $\beta_1=0$, $\beta_2=1$, $\eta_{t+1}=\eta$ with $\eta>0$ and $\alpha\rightarrow 1^-$.
Then, for any $\delta \in (0,1)$ and $S_{t,w,\alpha}(\xx_t)$ in \eqref{def}, with probability at least $1-\delta$,  the iterates $\xx_t$ satisfy the following bound
%Then, feeding Algorithm~\ref{algo:adagrad} into Algorithm~\ref{algo:mlol} gives the following upper bound of $DLR_w(T)$, with probability $1-\delta$:
\begin{equation*}
\sum^T_{t=1}\esp{\|\nabla S_{t,w,\alpha}(\xx_t)\|^2}
\le \frac{4C\sqrt{\epsilon}}{\delta}+\frac{8C\sqrt{\zeta T}}{\delta^{3/2}}+\frac{48C^2}{\delta^2}~.
\end{equation*}
Here, $C \triangleq \varpi_1+\varpi_2d\ln\left(1 + \frac{2(\zeta+L'^2)T}{d\epsilon}\right)$, where
\begin{align*}
\varpi_1 \triangleq  \frac{4DT}{W\eta},\qquad
\varpi_2 \triangleq \frac{\eta\gamma'+4\sqrt{\zeta}}{2},\qquad \zeta \triangleq \frac{\sigma^2}{W}.
\end{align*}
\end{thm*}

\begin{proof}

In light of Lemma~\ref{lm:property}, $\ell_t$ functions are $\gamma'$-smooth, which lead to $\gamma'$-smoothness of $S_{t,w,\alpha}$.

By the definition of $\xx_{t+1}$, we have:

\begin{align*}
\frac{S_{t,w,\alpha}(\xx_{t+1})-S_{t,w,\alpha}(\xx_t)}{\eta}&\le \langle \nabla S_{t,w,\alpha}(\xx_t),\xx_{t+1}-\xx_t\rangle+\frac{\gamma'}{2}\|\xx_{t+1}-\xx_t \|^2
\\&=-\sum_{i=1}^{d}\langle \nabla_i S_{t,w,\alpha}(\xx_t),
\frac{\tilde{\nabla}_iS_{t,w,\alpha}(\xx_t)}{\sqrt{\epsilon+\upsilon_{t+1,i}}} \rangle
+
\frac{\eta\gamma'}{2}\sum_{i=1}^{d}\frac{(\tilde{\nabla}_iS_{t,w,\alpha}(\xx_t))^2}{\epsilon+\upsilon_{t+1,i}}~.
\end{align*}
Then, taking expectation w.r.t.~$\xi_{t, t-w+1},\ldots,\xi_{t, t}$ conditioned on $\xi_{1:t-1}$ (namely $\E_t[\cdot]$) gives
\begin{align}\label{eq:inner}
 \nonumber \frac{\E_t[{S_{t,w,\alpha}(\xx_{t+1})-S_{t,w,\alpha}(\xx_t)}]}{\eta}
&\le
-\sum_{i=1}^{d}\E_t\left[\langle \nabla_i S_{t,w,\alpha}(\xx_t),
\frac{\tilde{\nabla}_iS_{t,w,\alpha}(\xx_t)}{\sqrt{\epsilon+\upsilon_{t+1,i}}} \rangle\right]
\\&\quad+
\frac{\eta\gamma'}{2}\sum_{i=1}^{d}\E_t\left[\frac{(\tilde{\nabla}_iS_{t,w,\alpha}(\xx_t))^2}{\epsilon+\upsilon_{t+1,i}}\right]~.
\end{align}
From Lemma~\ref{app:lemma:descent_lemmaa} and \eqref{eq:inner}, we have

\begin{align*}
&\frac{\E_t[{S_{t,w,\alpha}(\xx_{t+1})]-S_{t,w,\alpha}(\xx_t)}}{\eta}
\le
-\sum_{i=1}^{d}\frac{(\nabla_i S_{t,w,\alpha}(\xx_t))^2}{2\sqrt{\epsilon+\tilde{\upsilon}_{t+1,i}}}
+
\left(\frac{\eta\gamma'}{2}+2\sqrt{\mu}\right)\sum_{i=1}^{d}\E_t\left[\frac{(\tilde{\nabla}_iS_{t,w,\alpha}(\xx_t))^2}{\epsilon+\upsilon_{t+1,i}}\right]~,
\end{align*}
where $\mu\triangleq\frac{\sigma^2(1-\alpha^{2w})}{W^2(1-\alpha^2)}$ and $\tilde{\upsilon}_{t+1,i}\triangleq \upsilon_{t,i} + (\nabla_i S_{t,w,\alpha}(\xx_t))^2 + \mu$.

On rearranging the terms, taking expectation w.r.t.~$\xi_{1:t-1}$  on both sides and summing over $t$ from $1$ to $T$, we obtain
\begin{align}\label{eq:term2}
\nonumber \sum^{T}_{t=1}\E\left[\sum_{i=1}^{d}\frac{(\nabla_i S_{t,w,\alpha}(\xx_t))^2}{2\sqrt{\epsilon+\tilde{\upsilon}_{t+1,i}}}\right]
&\le
\frac{\sum^{T}_{t=1}\big(\E[S_{t,w,\alpha}(\xx_t)]-\E[S_{t,w,\alpha}(\xx_{t+1})]\big)}{\eta}
\\&\quad+
\frac{\eta\gamma'+4\sqrt{\mu}}{2}\sum_{i=1}^{d}\esp{\sum^{T}_{t=1}\frac{(\tilde{\nabla}_iS_{t,w,\alpha}(\xx_t))^2}{\epsilon+\upsilon_{t+1,i}}}~.
\end{align}
We then upper bound the last term in \eqref{eq:term2}. Recall that $\upsilon_{t+1,i}=\sum_{j=1}^{t}(\tilde{\nabla}_iS_{j,w,\alpha}(\xx_j))^2$, $\upsilon_{1,i}=0$.  Therefore, we have
\begin{align}\label{5}
\nonumber &\E\left[\sum^{T}_{t=1}\|\frac{\tilde{\nabla}S_{t,w,\alpha}(\xx_t)}{\sqrt{\epsilon+\up_{t+1}}}\|^2\right]
 =\E\left[\sum_{i=1}^{d}\sum^{T}_{t=1}\frac{(\tilde{\nabla}_iS_{t,w,\alpha}(\xx_t))^2}{\epsilon+\sum_{j=1}^{t}(\tilde{\nabla}_iS_{j,w,\alpha}(\xx_j))^2}\right]\\\nonumber
&\stackrel{(i)} \leq \E\left[\sum_{i=1}^{d}\ln\big(1 +\frac{\sum^{T}_{t=1}(\tilde{\nabla}_iS_{t,w,\alpha}(\xx_t) )^2}{\epsilon} \big)\right]\stackrel{(ii)} \leq \E\left[d\ln\Big(\frac{1}{d}\sum_{i=1}^{d}\big(1 +\frac{\sum^{T}_{t=1}(\tilde{\nabla}_iS_{t,w,\alpha}(\xx_t) )^2}{\epsilon} \big)\Big)\right]\\\nonumber
 &= \E\left[d\ln \big(1 +\frac{\sum^{T}_{t=1}\sum_{i=1}^{d}(\tilde{\nabla}_iS_{t,w,\alpha}(\xx_t) )^2}{ d\epsilon} \big)\right]\stackrel{(iii)} \le
d\ln\big(1 +\frac{\sum^{T}_{t=1}\E\left[\|\tilde{\nabla}S_{t,w,\alpha}(\xx_t) \|^2\right]}{ d\epsilon} \big)
\\\nonumber &\stackrel{(iv)} \le
d\ln\big(1 +\frac{2\sum^{T}_{t=1}\E\left[\|\tilde{\nabla}S_{t,w,\alpha}(\xx_t)-\nabla S_{t,w,\alpha}(\xx_t) \|^2\right]+2\sum^{T}_{t=1}\E\left[\|\nabla S_{t,w,\alpha}(\xx_t) \|^2\right]}{ d\epsilon} \big)
\\&\stackrel{(v)}\le d\ln\big(1 +\frac{2T(\mu+L'^2)}{d\epsilon} \big)~,
\end{align}
where (i) holds because Lemma~\ref{lemma:sum_ratio} with $\beta_2=1$, (ii) is due to the convex inequality
$\frac{1}{d}\sum_{i=1}^{d}\ln (a_i)\leq \ln (\frac{1}{d}\sum_{i=1}^{d} a_i)$, (iii) follows from
$\E [ \ln(A)]\leq \ln \E[A]$, for any positive random variable $A$, (iv) follows from $\| a+b\|^2\leq 2\|a \|^2+2\|b\|^2$, (v) follows from Lemmas Lemma~\ref{lm:property} and
\ref{p0}(b).

Therefore, plugging \eqref{5} into \eqref{eq:term2}, we get
\begin{align*}
 I_1&=\sum^{T}_{t=1}\E\left[\sum_{i=1}^{d}\frac{(\nabla_i S_{t,w,\alpha}(\xx_t))^2}{2\sqrt{\epsilon+\tilde{\upsilon}_{t+1,i}}}\right]
\\\nonumber &\le
\frac{\sum^{T}_{t=1}\big(\E[S_{t,w,\alpha}(\xx_t)]-\E[S_{t,w,\alpha}(\xx_{t+1})]\big)}{\eta}+
\left(\frac{\eta\gamma'+4\sqrt{\mu}}{2}\right)d\ln\big(1 + \frac{2T(\mu+L'^2)}{d\epsilon}\big)
\\\nonumber&=
\frac{\sum^{T}_{t=1}\big(\E[S_{t,w,\alpha}(\xx_t)]-\E[S_{t+1,w,\alpha}(\xx_{t+1})]\big)}{\eta}
+\frac{\sum^{T}_{t=1}\big(\E[S_{t+1,w,\alpha}(\xx_{t+1})]-\E[S_{t,w,\alpha}(\xx_{t+1})]\big)}{\eta}
\\\nonumber&\quad+\left(\frac{\eta\gamma'+4\sqrt{\mu}}{2}\right)d\ln\big(1 + \frac{2T(\mu+L'^2)}{d\epsilon}\big).
\end{align*}
Using Lemmas~\ref{lm:sumobj22} and \ref{lm:sumobj2}, we get
\begin{align*}
\nonumber I_1&\leq \frac{2D(1-\alpha^w)T}{W(1-\alpha)\eta}+\frac{D(1+\alpha^{w-1})T}{W\eta}
+\frac{D(1-\alpha^{w-1})(1+\alpha)T}{W(1-\alpha)\eta}
\\ &\qquad+\left(\frac{\eta\gamma'+4\sqrt{\mu}}{2}\right)d\ln\big(1 + \frac{2T(\mu+L'^2)}{d\epsilon}\big)\triangleq C.
\end{align*}
Now, by Markov's inequality, we have with probability at least $1-\delta_1$,

\begin{equation}\label{dd1}
\sum^{T}_{t=1}\E\left[\sum_{i=1}^{d}\frac{(\nabla_i S_{t,w,\alpha}(\xx_t))^2}{2\sqrt{\epsilon+\tilde{\upsilon}_{t+1,i}}}\right]
\le
\frac{C}{\delta_1}~.
\end{equation}

Additionally, it follows from Markov's inequality and Lemma~\ref{p0}(b) that with probability at least $1-\delta_2$,
\begin{equation*}
\sum^{T}_{t=1}\|\nabla S_{t,w,\alpha}(\xx_t) -\tilde{\nabla}S_{t,w,\alpha}(\xx_t)\|^2\le\frac{T\mu}{\delta_2}
\end{equation*}
holds.

Using the above inequality along with the bound $\|a+b\|^2\leq 2\|a \|^2+2\| b\|^2$, we also have
\begin{align*}
\sum^{T}_{t=1}\|\tilde{\nabla}S_{t,w,\alpha}(\xx_t)\|^2
\nonumber&\le
2\sum^{T}_{t=1}\|\tilde{\nabla}S_{t,w,\alpha}(\xx_t) - \nabla S_{t,w,\alpha}(\xx_t)\|^2
+ 2\sum^{T}_{t=1}\|\nabla S_{t,w,\alpha}(\xx_t)\|^2
\\&\le
\frac{2T\mu}{\delta_2}
+ 2\sum^{T}_{t=1}\|\nabla S_{t,w,\alpha}(\xx_t)\|^2.
\end{align*}
Now, recalling that $\up_{t+1}= \up_{t}+(\tilde{\nabla}S_{t,w,\alpha}(\xx_t))^2=\sum_{l=1}^{t}(\tilde{\nabla}S_{l,w,\alpha}(\xx_l))^2$, $\up_{1}=0$ and using the above inequality yields that with probability at least $1-\delta_2$,

\begin{align}\label{332}
\nonumber  \sum_{i=1}^{d}&\upsilon_{T,i} +\sum_{i=1}^{d} (\nabla_i S_{T,w,\alpha}(\xx_{T}))^2 + \mu
=\sum_{t=1}^{T-1}\sum_{i=1}^{d}(\tilde{\nabla}_i S_{t,w,\alpha}(\xx_t))^2
+\sum_{i=1}^{d}(\nabla_i S_{T,w,\alpha}(\xx_{T}))^2+ \mu
\\\nonumber&\leq \frac{2\mu(T-1)}{\delta_2}+2\sum_{t=1}^{T-1}\sum_{i=1}^{d}(\nabla_i S_{t,w,\alpha}(\xx_t))^2+\sum_{i=1}^{d}(\nabla_i S_{T,w,\alpha}(\xx_{T}))^2+\mu
\\ &\leq \frac{2\mu(T-1)}{\delta_2}+2\sum_{t=1}^{T-1}\sum_{i=1}^{d}(\nabla_i S_{t,w,\alpha}(\xx_t))^2+2\sum_{i=1}^{d}(\nabla_i S_{T,w,\alpha}(\xx_{T}))^2
+ \frac{2\mu}{\delta_2}
\le\
2Z + \frac{2T\mu}{\delta_2},
\end{align}

where $Z\triangleq\sum^{T}_{t=1}\|\nabla S_{t,w,\alpha}(\xx_t)\|^2$. Based on this notation, with probability at least $1-\delta_2$, we have
\begin{align}\label{dd}
\nonumber & \sum^{T}_{t=1}\sum_{i=1}^{d}\frac{(\nabla_i S_{t,w,\alpha}(\xx_t))^2}{2\sqrt{\epsilon+\tilde{\upsilon}_{t+1,i}}}
=\sum^{T}_{t=1}\sum_{i=1}^{d}\frac{(\nabla_i S_{t,w,\alpha}(\xx_t))^2}{2\sqrt{\epsilon+\upsilon_{t,i} + (\nabla_i S_{t,w,\alpha}(\xx_t))^2 + \mu}}
\\\nonumber&\ge
\sum^{T}_{t=1}\sum_{i=1}^{d}\frac{(\nabla_i S_{t,w,\alpha}(\xx_t))^2}{2\sqrt{\epsilon+\upsilon_{T,i}
+ (\nabla_i S_{T,w,\alpha}(\xx_T))^2 + \sum_{t=1}^{T}(\nabla_i S_{t,w,\alpha}(\xx_t))^2 + \mu}}
\\\nonumber&\ge
\sum^{T}_{t=1}\sum_{i=1}^{d}\frac{(\nabla_i S_{t,w,\alpha}(\xx_t))^2}{2\sqrt{\epsilon+\sum_{i=1}^{d}\upsilon_{T,i} + \sum_{i=1}^{d}(\nabla_i S_{T,w,\alpha}(\xx_T))^2 + \sum_{t=1}^{T}\sum_{i=1}^{d}(\nabla_i S_{t,w,\alpha}(\xx_t))^2 +\mu}}
\\&\ge
\frac{\sum^{T}_{t=1}\|\nabla S_{t,w,\alpha}(\xx_t)\|^2}{2\sqrt{\epsilon + 3Z + \frac{2T\mu}{\delta_2}}}~,
\end{align}
where in the last inequality we used \eqref{332}.
From \eqref{dd} and \eqref{dd1}, with probability at least $1-\delta_1-\delta_2$, we obtain
\begin{equation*}
\frac{Z}{2\sqrt{\epsilon+ 3Z + \frac{2T\mu}{\delta_2}}}
\le
\frac{C}{\delta_1}~.
\end{equation*}
From Lemma \ref{lem:quadratic} and

setting $\delta_1=\delta_2=\frac{\delta}{2}$, we have

\begin{equation}\label{mmm}
\sum^T_{t=1}\esp{\|\nabla S_{t,w,\alpha}(\xx_t)\|^2}
\le \frac{4C\sqrt{\epsilon}}{\delta}+\frac{8C\sqrt{\mu T}}{\delta^{3/2}}+\frac{48C^2}{\delta^2}~.
\end{equation}
Here, $C \triangleq \nu+ud\ln\left(1 + \frac{2(\mu+L'^2)T}{d\epsilon}\right)$, where
\begin{align*}
\nu &\triangleq  \frac{2D(1-\alpha^w)T}{W(1-\alpha)\eta}+\frac{D(1+\alpha^{w-1})T}{W\eta}+\frac{D(1-\alpha^{w-1})(1+\alpha)T}{W(1-\alpha)\eta},\\
u &\triangleq \frac{\eta\gamma'+4\sqrt{\mu}}{2},\qquad \mu \triangleq \frac{\sigma^2(1-\alpha^{2w})}{W^2(1-\alpha^2)}.
\end{align*}
Then, we can bound $\nu$ as
\begin{align}\label{frd1}
\nonumber \nu &= \frac{D T}{W\eta}\Big(\frac{2(1-\alpha^w)}{(1-\alpha)}+(1+\alpha^{w-1})+\frac{(1-\alpha^{w-1})(1+\alpha)}{(1-\alpha)}\Big)
\\\nonumber&\leq \frac{D T}{W\eta}\Big(\frac{2(1-\alpha^w)}{(1-\alpha)}+(1+\alpha^{w-1})+\frac{(1-\alpha^{w})(1+\alpha)}{(1-\alpha)}\Big)\\
\nonumber&\leq \frac{D T}{W\eta}\Big(\frac{(1-\alpha^w)}{(1-\alpha)}(3+\alpha)+(1+\alpha^{w-1})\Big)
\leq \frac{D T}{W\eta}\Big(4\frac{(1-\alpha^w)}{(1-\alpha)}+(1+\alpha^{w-1})\Big)
\\\nonumber &\leq \frac{D T}{W\eta}\Big(4\frac{(1-\alpha^w)}{(1-\alpha)}+\frac{(1+\alpha^{w-1})}{1-\alpha}\Big)
\\&\leq \frac{4D T}{W\eta}\Big(\frac{(1-\alpha^w)}{(1-\alpha)}+\frac{(1+\alpha^{w-1})}{1-\alpha}\Big)
= \frac{4D T}{W\eta}\Big(\frac{2-\alpha^{w}+\alpha^{w-1}}{1-\alpha}\Big).
\end{align}
As $\alpha \rightarrow 1^-$, we have
\begin{equation*}\label{fff}
  \nu \leq \frac{4D T}{W\eta}\triangleq \varpi_1,\qquad
  \mu=\frac{\sigma^2(1-\alpha^{2w})}{W^2(1-\alpha^2)} \stackrel{\alpha \rightarrow 1^-}=\frac{\sigma^2}{W}\triangleq\zeta.
\end{equation*}
Plugging these bounds back to \eqref{mmm}, we get the stated bound.
\end{proof}
\subsubsection*{ PROOF OF COROLLARY~\ref{corr1}}
\begin{cor*}
Under the same conditions stated in Theorem \ref{thm1}, using $w\in\Theta(T)$ and $\alpha\rightarrow 1^{-}$ yields a regret bound of order
\begin{equation*}
\sum^T_{t=1}\esp{\|\nabla S_{t,w,\alpha}(\xx_t)\|^2}\leq \mathcal{O}(\ln T).
\end{equation*}
\end{cor*}
\begin{proof}
By Theorem \ref{thm1}, we have
\begin{equation*}
\sum^T_{t=1}\esp{\|\nabla S_{t,w,\alpha}(\xx_t)\|^2}
\le \frac{4C\sqrt{\epsilon}}{\delta}+\frac{8C\sqrt{\zeta T}}{\delta^{3/2}}+\frac{48C^2}{\delta^2}\triangleq I_1+I_2+I_3~,
\end{equation*}
where $C = \varpi_1+\varpi_2d\ln\left(1 + \frac{2(\zeta+L'^2)T}{d\epsilon}\right)$, $\varpi_1 =  \frac{4DT}{W\eta}$, $\varpi_2 = \frac{\eta\gamma'+4\sqrt{\zeta}}{2}$, and $\zeta =\frac{\sigma^2}{W}$.\\
Recall that $W = \sum_{r=0}^{w-1}\alpha^r$. As $\alpha\rightarrow 1^{-}$ and $w\in\Theta(T)$, we get the following equalities:
\begin{align*}
C&=\frac{4DT}{W\eta}+(\frac{\eta\gamma'+4\sqrt{\frac{\sigma^2}{W}}}{2})d\ln\left(1 + \frac{2(\frac{\sigma^2}{W}+L'^2)T}{d\epsilon}\right)=\mathcal{O}(\ln T),\\
  I_1&=\frac{4C\sqrt{\epsilon}}{\delta} =\mathcal{O}(\ln T),\qquad
  I_2=\frac{8C\sqrt{\zeta T}}{\delta^{3/2}} =\frac{8C}{\delta^{3/2}}\sqrt{\frac{\sigma^2}{W} T}=\mathcal{O}(\ln T), \qquad
  I_3=\frac{48C^2}{\delta^2}=\mathcal{O}(\ln T).
\end{align*}
Combine the above results we can easily have the desired result.
\end{proof}

\subsubsection*{ PROOF OF LEMMA~\ref{app:lemma:descent_lemma}}
The proof relies on the following lemma. For the sake of concision, we use the notations
\begin{equation}\label{cvbb}
  B_{t}\triangleq \frac{\mm_{t}}{\sqrt{\epsilon+\up_{t}}},\quad
A_{t}\triangleq \frac{\g_{t}}{\sqrt{\epsilon+\up_{t}}}
\end{equation}
 below.
\begin{lem}\label{881}
Suppose Assumption~\ref{smooth} holds. Then, for any $0<\alpha<1$, $w$, $1\leq k\leq t$ and $S_{t,w,\alpha}(\xx_t)$ in \eqref{def}, we have
\begin{align}\label{cvbbb}
\nonumber &\norm{\nabla S_{t+1-k,w,\alpha}(\xx_{t+1-k})-\nabla S_{t,w,\alpha}(\xx_{t+1-k})}^2
\\&\qquad \leq \frac{8L'^2 }{W^2} +\frac{2(1-\alpha^{w-2})\gamma'^2}{W^2(1-\alpha)} \sum_{r=1}^{w-1}  \alpha^{r-1} \|\eta_{t-r+2-k}B_{t-r+2-k} \|^2\triangleq \vartheta_t.
\end{align}
\end{lem}

\begin{proof}
By the definition of $S_{t,w,\alpha}(\xx_t)$ in \eqref{def}, we have
\begin{align*}
\nonumber &\norm{\nabla S_{t+1-k,w,\alpha}(\xx_{t+1-k})-\nabla S_{t,w,\alpha}(\xx_{t+1-k})}^2
\\&\quad= \frac{1}{W^2}\norm{ \sum_{r=0}^{w-1} \alpha^r \left( \nabla  \ell_{t+1-k-r}(\xx_{t+1-k-r})-\nabla \ell_{t-r}(\xx_{t+1-k-r})  \right)}^2 \notag \\
&\quad= \frac{1}{W^2}\left\lVert \nabla \ell_{t+1-k}(\xx_{t+1-k}) -\nabla \ell_{t}(\xx_{t+1-k}) + \alpha \nabla\ell_{t-k}(\xx_{t-k})
 \right. \notag  \\
&  \left.   \quad\quad- \alpha \nabla \ell_{t-1}(\xx_{t-k})
+ \cdots+ \alpha^{w-1} \nabla\ell_{t-w+2-k}(\xx_{t-k-w+2})
- \alpha^{w-1} \nabla\ell_{t-w+1}(\xx_{t-k-w+2}) \right\lVert^2 \notag \\\nonumber
&\quad\leq \frac{2}{W^2} \left\lVert \nabla\ell_{t+1-k}(\xx_{t+1-k})-\alpha^{w-1} \nabla\ell_{t-w+1}(\xx_{t-k-w+2}) \right\lVert^2
\\\nonumber &\quad \quad+ \frac{2}{W^2}\left\lVert \sum_{r=1}^{w-1}\big(\alpha^r \nabla\ell_{t-r+1-k}(\xx_{t-r+1-k})-  \alpha^{r-1} \nabla\ell_{t-r+1-k}(\xx_{t-r+2-k}) \big) \right\lVert^2
\\
&\quad \leq \frac{8L'^2 }{W^2} +\frac{2(1-\alpha^{w-2})}{W^2(1-\alpha)} \sum_{r=1}^{w-1}  \alpha^{r-1} \gamma'^2\|\eta_{t-r+2-k}B_{t-r+2-k} \|^2,
\end{align*}
where the last inequality follows from the following argument:
\begin{align*}
\nonumber & \left\lVert\nabla\ell_{t+1-k}(\xx_{t+1-k}) - \alpha^{w-1} \nabla\ell_{t-w+1}(\xx_{t-k-w+2})\right\lVert^2
\\\nonumber&\stackrel{(i)}\leq  2 \|\nabla\ell_{t+1-k}(\xx_{t+1-k}) \|^2
+2\alpha^{2w-2} \| \nabla\ell_{t-w+1}(\xx_{t-k-w+2}) \|^2
\stackrel{(ii)}\leq 2L'^2(1+\alpha^{2w-2})< 4L'^2,
\end{align*}
where (i) is valid by considering $\| a+b\|^2\leq 2\|a \|^2+2\| b\|^2$, and (ii) is due to Lemma~\ref{lm:property}. Note that
\begin{align*}
&\left\lVert \sum_{r=1}^{w-1} \big(\alpha^r \nabla\ell_{t-r+1-k}(\xx_{t-r+1-k}) - \alpha^{r-1} \nabla\ell_{t-r+1-k}(\xx_{t-r+2-k})\big) \right\lVert^2
\\&=
\sum_{r=1}^{w-1} \sum_{j=1}^{w-1}\langle \alpha^r \nabla\ell_{t-r+1-k}(\xx_{t-r+1-k}) - \alpha^{r-1} \nabla\ell_{t-r+1-k}(\xx_{t-r+2-k})
\\&\qquad \qquad \qquad \qquad,\alpha^j \nabla\ell_{t-j+1-k}(\xx_{t-j+1-k}) - \alpha^{j-1} \nabla\ell_{t-j+1-k}(\xx_{t-j+2-k}) \rangle.
\end{align*}
Furthermore,
\begin{align*}
&\left\lVert \sum_{r=1}^{w-1} \big(\alpha^r \nabla\ell_{t-r+1-k}(\xx_{t-r+1-k}) - \alpha^{r-1} \nabla\ell_{t-r+1-k}(\xx_{t-r+2-k})\big) \right\lVert^2
\\& \stackrel{(i)}\leq\sum_{r=1}^{w-1} \sum_{j=1}^{w-1} \alpha^{r-1} \alpha^{j-1} \frac{1}{2}
(\|\alpha \nabla\ell_{t-r+1-k}(\xx_{t-r+1-k}) - \nabla\ell_{t-r+1-k}(\xx_{t-r+2-k}) \|^2
\\&\quad+\|\alpha \nabla\ell_{t-j+1-k}(\xx_{t-j+1-k}) -  \nabla\ell_{t-j+1-k}(\xx_{t-j+2-k}) \|^2)
\\& \stackrel{(ii)}\leq \sum_{j=1}^{w-1} \alpha^{j-1} \sum_{r=1}^{w-1}\alpha^{r-1}
\|\alpha \nabla\ell_{t-r+1-k}(\xx_{t-r+1-k}) - \nabla\ell_{t-r+1-k}(\xx_{t-r+2-k}) \|^2
\\& \stackrel{(iii)}\leq \frac{(1-\alpha^{w-2})}{(1-\alpha)}\sum_{r=1}^{w-1}  \alpha^{r-1}
\gamma'^2\| \xx_{t-r+1-k} - \xx_{t-r+2-k} \|^2
 =\frac{(1-\alpha^{w-2})}{(1-\alpha)}\sum_{r=1}^{w-1}  \alpha^{r-1}
\gamma'^2\|\eta_{t-r+2-k}B_{t-r+2-k} \|^2,
\end{align*}
where (i) is due to $\langle a,b \rangle\leq \frac{1}{2}(\|a \|^2+\| b\|^2)$, (ii) is due to symmetry of $r$ and $j$ in summation, and (iii) is due to Lemma~\ref{lm:property}.
\end{proof}

\begin{lem*}
Suppose Assumptions~\ref{ass:sg}, \ref{ii} and \ref{ass} hold.
Let $m_{t+1,i}$ and $\upsilon_{t+1,i}$ be the sequences defined in the Algorithm $\textsc{Dts-Ag}$.
Then, for any $0<\alpha<1$, $w$, $\epsilon>0$, $0\leq \beta_1<\beta_2\leq 1$, $1\leq k\leq t$ and $S_{t,w,\alpha}(\xx_t)$ in \eqref{def}, we have
\begin{align*}
\nonumber
 & \esp{  \sum_{i =1}^d \langle \nabla_i S_{t,w,\alpha}(\xx_t) , \frac{m_{t+1,i}}{\sqrt{\epsilon+\upsilon_{t+1,i}}} \rangle }\geq
        \sum_{i=1}^d \sum_{k=0}^{t-1}\beta_1^k\frac{(\nabla_i S_{t+1-k,w,\alpha}(\xx_{t+1-k}) )^2}{2\sqrt{\epsilon+\tilde{\upsilon}_{t+1,i}}}
      \\\nonumber
 &\quad \quad -
      \frac{1}{\sqrt{1-\beta_1}}\esp{\sum_{k=0}^{t-1}
     \bbratio^k (\sqrt{k+1}+2\sqrt{\mu}) \|A_{t+1 -k}\|^2}
     \\\nonumber&\quad\quad-\frac{\eta_{t+1}^2 \gamma'^2}{4}\sqrt{1 - \beta_1} \esp{\sum_{l=1}^{t}\norm{B_{t+1-l}}^2}\sum_{k=l}^{t }\beta_1^k \sqrt{k}  -\sum_{k=0}^{t-1}\beta_1^k\frac{\sqrt{1-\beta_1}\vartheta_t}{2\sqrt{k+1}}.
\end{align*}
Here, $\tilde{\upsilon}_{t+1,i}\triangleq \beta_2\upsilon_{t,i} + (\nabla_i S_{t,w,\alpha}(\xx_t))^2 + \mu$, $\mu\triangleq\frac{\sigma^2(1-\alpha^{2w})}{W^2(1-\alpha^2)}$ for all $i\in [d]$, and $A_t$, $B_t$ and $\vartheta_t$ are defined as in \eqref{cvbb} and \eqref{cvbbb}.
\end{lem*}
\begin{proof}
First, we have
\begin{align} \label{app:eq:descent_big}
  & -\sum_{i=1}^d \langle \nabla_i S_{t,w,\alpha}(\xx_t) , \frac{m_{t+1,i}}{\sqrt{\epsilon+\upsilon_{t+1,i}}} \rangle
=
       - \sum_{i=1}^d \sum_{k=0}^{t-1} \beta_1^k \langle \nabla_i S_{t,w,\alpha}(\xx_t) , \frac{\tilde{\nabla}_i S_{t+1-k,w,\alpha}(\xx_{t+1-k})}{\sqrt{\epsilon+\upsilon_{t+1,i}}} \rangle
      =I_1+ I_2,
     \end{align}
     where
     \begin{align*}
     I_1=&-\sum_{i=1}^d \sum_{k=0}^{t-1}\beta_1^k \langle \nabla_i S_{t+1-k,w,\alpha}(\xx_{t+1-k}) , \frac{\tilde{\nabla}_i S_{t+1-k,w,\alpha}(\xx_{t+1-k})}{\sqrt{\epsilon+\upsilon_{t+1,i}}} \rangle,\\
       I_2=& \sum_{i=1}^d \sum_{k=0}^{t-1}  \beta_1^k \langle (\nabla_i S_{t+1-k,w,\alpha}(\xx_{t+1-k})-\nabla_i S_{t,w,\alpha}(\xx_t)) , \frac{\tilde{\nabla}_i S_{t+1-k,w,\alpha}(\xx_{t+1-k})}{\sqrt{\epsilon+\upsilon_{t+1,i}}} \rangle.
     \end{align*}
Next, we proceed to upper bound $I_1$ and $I_2$ terms.
For convenience, we denote
\begin{align*}
g_{t+1-k,i}&\triangleq \tilde{\nabla}_i S_{t+1-k,w,\alpha}(\xx_{t+1-k}),\,\, \text{for all}\,\, i\in [d].
\end{align*}
\begin{itemize}

\item Bound for $I_2$.

Applying $ab\le\frac{\hslash}{2}a^2 + \frac{1}{2\hslash}b^2$
with $\hslash = \frac{\sqrt{1 - \beta_1}}{2 \sqrt{k + 1}}$,
 \begin{equation*}
    a = \abs{\nabla_i S_{t+1-k,w,\alpha}(\xx_{t+1-k})-\nabla_i S_{t,w,\alpha}(\xx_t)}, \;
    b = \frac{\abs{g_{t+1-k,i}}}{\sqep{\upsilon_{t+1,i}}},
 \end{equation*}
we can upper bound $I_2$ term as
\begin{align}
  &  \abs{I_2} \leq  \sum_{k=0}^{t-1} \beta_1^k \left(
        \frac{\sqrt{1 - \beta_1}}{4  \sqrt{k+1}}\sum_{i=1}^{d}\big(\nabla_i S_{t+1-k,w,\alpha}(\xx_{t+1-k})-\nabla_i S_{t,w,\alpha}(\xx_t)\big)^2
         + \frac{\sqrt{k+1}}{\sqrt{1-\beta_1}} \frac{g_{t+1-k,i}^2}{\eps + \upsilon_{t+1, i}}
    \right)
    \label{app:eq:B_bound_first}.
\end{align}
Following the notations of \eqref{cvbb} and \eqref{cvbbb}, the first term in \eqref{app:eq:B_bound_first} can be bounded as
\begin{align}
\nonumber
   & \|\nabla S_{t+1-k,w,\alpha}(\xx_{t+1-k})-\nabla S_{t,w,\alpha}(\xx_t)\|^2
   \\\nonumber &\leq 2 \|\nabla S_{t+1-k,w,\alpha}(\xx_{t+1-k})-\nabla S_{t,w,\alpha}(\xx_{t+1-k})\|^2
+2 \|\nabla S_{t,w,\alpha}(\xx_{t+1-k})-\nabla S_{t,w,\alpha}(\xx_t)\|^2
   \\ &\leq 2\vartheta_t+2\gamma'^2 \|\xx_{t } - \xx_{t +1- k }\|^2
   = 2\vartheta_t+2\gamma'^2 \| \sum_{l=1}^{k} \eta_{t+1 -l} B_{t+1 - l}\|^2\leq 2\vartheta_t+2\eta_{t+1}^2 \gamma'^2 k \sum_{l=1}^{k} \|B_{t+1-l}\|^2
        \label{app:eq:delta_first},
\end{align}
where the first inequality uses $\|\mathrm{a}+\mathrm{b}\|^2\leq 2\|\mathrm{a}\|^2+2\|\mathrm{b}\|^2$, and the second inequality is from  $\gamma'$-smoothness of $S_{t,w,\alpha}$ given by Lemma~\ref{lm:property} and Lemma~\ref{881}.\\
For the second term in \eqref{app:eq:B_bound_first}, notice that for any dimension $i \in [d]$, we have
$$ \eps + \upsilon_{t+1, i} \geq \eps + \beta_2^{k} \upsilon_{t+1 -k, i} \geq \beta_2^{k} (\eps + \upsilon_{t+1-k, i}),$$
so that
\begin{equation}
\label{app:eq:vbeta}
\frac{g_{t+1-k,i}^2}{\eps + \upsilon_{t+1, i}} \leq \frac{g_{t+1-k,i}^2}{\beta_2^k(\eps + \upsilon_{t+1-k, i})}.
 \end{equation}

Injecting \eqref{app:eq:vbeta} and \eqref{app:eq:delta_first} into \eqref{app:eq:B_bound_first}, we obtain
\begin{align}
\nonumber
    \abs{I_2} &\leq
         \sum_{k=0}^{t-1} \frac{\eta_{t+1}^2 {\gamma'}^2}{4 }\sqrt{1 - \beta_1}\beta_1^k \sqrt{k} \sum_{l=1}^{k} \|B_{t+1-l}\|^2
        + \sum_{k=0}^{t-1} \beta_1^k
        \frac{\sqrt{1 - \beta_1}\vartheta_t}{2 \sqrt{k+1}}
        \\\nonumber&\quad+
         \sum_{k=0}^{t-1} \frac{1}{\sqrt{1-\beta_1}}  \bbratio^k\sqrt{k + 1}\norm{A_{t+1-k}}^2\\\nonumber
    &=
                \frac{\eta_{t+1}^2 \gamma'^2}{4 }\sqrt{1 - \beta_1} \sum_{l=1}^{t}\norm{B_{t+1-l}}^2\sum_{k=l}^{t}\beta_1^k \sqrt{k}
                + \sum_{k=0}^{t-1} \beta_1^k
        \frac{\sqrt{1 - \beta_1}\vartheta_t}{2  \sqrt{k+1}}
       \\& \quad+\frac{1}{\sqrt{1 - \beta_1}} \sum_{k=0}^{t-1} \bbratio^k \sqrt{k+1}\norm{A_{t+1-k}}^2
    \label{app:eq:B_bound_second},
\end{align}
where $A_{t+1-k}$ is defined as in \eqref{cvbb}.
 \item Bound for $I_1$.\\
Let us denote $\tilde{\upsilon}_{t+1,i}\triangleq \beta_2\upsilon_{t,i} + (\nabla_i S_{t,w,\alpha}(\xx_t))^2 + \mu$ for all $i\in [d]$. With our new notation, we have
\begin{align}
    \esp{I_1} = I_{11} + I_{12},
    \label{app:eq:cool_plus_bad}
\end{align}
where
\begin{align*}
I_{11}&=-\esp{\sum_{i=1}^{d}\frac{1}{\sqrt{\epsilon+\tilde{\upsilon}_{t+1,i}}}\langle \nabla_i S_{t+1-k,w,\alpha}(\xx_{t+1-k}) , g_{t+1-k,i}\rangle},\\
I_{12}&=\esp{\sum_{i=1}^{d}(\frac{1}{\sqrt{\epsilon+\tilde{\upsilon}_{t+1,i}}}-\frac{1}{\sqrt{\epsilon+\upsilon_{t+1,i}}})\langle \nabla_i S_{t+1-k,w,\alpha}(\xx_{t+1-k}) , g_{t+1-k,i} \rangle}.
\end{align*}
First, we give an upper bound on $I_{11}$ term in \eqref{app:eq:cool_plus_bad}.

By using Lemma~\ref{p0}(a), we have
\begin{align}\label{ggf}
\nonumber I_{11}&= -\E_{t+1-k}\left[\frac{1}{\sqrt{\epsilon+\tilde{\upsilon}_{t+1,i}}}\langle \nabla_i S_{t+1-k,w,\alpha}(\xx_{t+1-k}) \right.\\\nonumber&\qquad\qquad \qquad,\left. \tilde{\nabla}_i S_{t+1-k,w,\alpha}(\xx_{t+1-k})\rangle - \nabla_i S_{t+1-k,w,\alpha}(\xx_{t+1-k})\rangle\right]\\\nonumber
&\qquad-\E_{t+1-k}\left[\frac{1}{\sqrt{\epsilon+\tilde{\upsilon}_{t+1,i}}}\langle \nabla_i S_{t+1-k,w,\alpha}(\xx_{t+1-k}) , \nabla_i S_{t+1-k,w,\alpha}(\xx_{t+1-k})\rangle \right]
\\&= \frac{-(\nabla_i S_{t+1-k,w,\alpha}(\xx_{t+1-k}) )^2}{\sqrt{\epsilon+\tilde{\upsilon}_{t+1,i}}}.
\end{align}

We next bound the $I_{12}$ in \eqref{app:eq:cool_plus_bad}.
To this end, we first observe that by the definition of $\upsilon_{t+1,i}$ and $\tilde{\upsilon}_{t+1,i}$, we have

\begin{align}\label{vgh}
\nonumber &\left|\frac{1}{\sqrt{\epsilon+\tilde{\upsilon}_{t+1,i}}}
-
\frac{1}{\sqrt{\epsilon+\upsilon_{t+1,i}}}\right|=
\frac{\left|(\tilde{\nabla}_iS_{t,w,\alpha}(\xx_t))^2 - (\nabla_i S_{t,w,\alpha}(\xx_t))^2 - \mu\right|}{\sqrt{\epsilon+\upsilon_{t+1,i}}\sqrt{\epsilon+\tilde{\upsilon}_{t+1,i}}\left(\sqrt{\epsilon+\upsilon_{t+1,i}}
+\sqrt{\epsilon+\tilde{\upsilon}_{t+1,i}}\right)}\\
& \le\frac{\left||\tilde{\nabla}_iS_{t,w,\alpha}(\xx_t)| - |\nabla_i S_{t,w,\alpha}(\xx_t)|\right|}{\sqrt{\epsilon+\upsilon_{t+1,i}}\sqrt{\epsilon+\tilde{\upsilon}_{t+1,i}}}
+
\frac{ \sqrt{\mu}}{\sqrt{\epsilon+\upsilon_{t+1,i}}\sqrt{\epsilon+\tilde{\upsilon}_{t+1,i}}}~,
\end{align}

for all $i\in [d]$. Since as Jensen's inequality on $|\cdot|$ and \eqref{vgh}, we can find that
\begin{align}\label{pl}
\nonumber &|I_{12}|\leq
\frac{1}{\sqrt{\epsilon+\upsilon_{t+1,i}}\sqrt{\epsilon+\tilde{\upsilon}_{t+1,i}}}
\\\nonumber &\E_{t+1-k}\left[\left||\tilde{\nabla}_iS_{t,w,\alpha}(\xx_t)| - |\nabla_i S_{t,w,\alpha}(\xx_t)|\right||\tilde{\nabla}_i S_{t+1-k,w,\alpha}(\xx_{t+1-k})||\nabla_i S_{t+1-k,w,\alpha}(\xx_{t+1-k})|\right]
\\&+\frac{1}{\sqrt{\epsilon+\upsilon_{t+1,i}}\sqrt{\epsilon+\tilde{\upsilon}_{t+1,i}}}
\E_{t+1-k}\left[|\tilde{\nabla}_i S_{t+1-k,w,\alpha}(\xx_{t+1-k})||\nabla_i S_{t+1-k,w,\alpha}(\xx_{t+1-k})| \sqrt{\mu}\right]~.
\end{align}
Now, in order to obtain a bound on the first term in the RHS of \eqref{pl}, we
invoke inequality $ab\le\frac{\hslash}{2}a^2 + \frac{1}{2\hslash}b^2$ with $\hslash = \frac{2\mu}{\sqrt{1-\beta_1}\sqrt{\epsilon+\tilde{\upsilon}_{t+1,i}}}$, $a=\frac{|\tilde{\nabla}_i S_{t+1-k,w,\alpha}(\xx_{t+1-k})|}{\sqrt{\epsilon+\upsilon_{t+1,i}}},$ which
leads to the bound
\begin{align*}
&\frac{(\nabla_i S_{t+1-k,w,\alpha}(\xx_{t+1-k}))^2}{4\sqrt{\epsilon+\tilde{\upsilon}_{t+1,i}}}
+
\frac{\sqrt{\mu}}{\sqrt{1-\beta_1}}\E_{t+1-k}\left[\frac{(\tilde{\nabla}_i S_{t+1-k,w,\alpha}(\xx_{t+1-k}))^2}{\epsilon+\upsilon_{t+1,i}}\right]~,
\end{align*}
where we used that $\left|\|\mathrm{u}\| - \|\mathrm{v}\|\right|\le\|\mathrm{u}- \mathrm{v}\|$ holds for $\forall \mathrm{u},\mathrm{v} \in\R^d$.

To upper bound the second term on the RHS of \eqref{pl}, using inequality $ab\le\frac{\hslash}{2}a^2 + \frac{1}{2\hslash}b^2$ with $\hslash = \frac{2}{\sqrt{1-\beta_1}\sqrt{\epsilon+\tilde{\upsilon}_{t+1,i}}}$, $a=\frac{|\tilde{\nabla}_i S_{t+1-k,w,\alpha}(\xx_{t-k})|\sqrt{\mu}}{\sqrt{\epsilon+\upsilon_{t+1,i}}}$ gives us the bound

\begin{align*}
&\frac{(\nabla_i S_{t+1-k,w,\alpha}(\xx_{t+1-k}))^2}{4\sqrt{\epsilon+\tilde{\upsilon}_{t+1,i}}}
+
\frac{\sqrt{\mu}}{\sqrt{1-\beta_1}}\E_{t+1-k}\left[\frac{(\tilde{\nabla}_i S_{t+1-k,w,\alpha}(\xx_{t+1-k}))^2}{\epsilon+\upsilon_{t+1,i}}\right]~.
\end{align*}
Together from the above two inequalities, it follows that
\begin{align}\label{dss}
\nonumber | I_{12}|&\leq \frac{(\nabla_i S_{t+1-k,w,\alpha}(\xx_{t+1-k}))^2}{2\sqrt{\epsilon+\tilde{\upsilon}_{t+1,i}}}
+
\frac{2\sqrt{\mu}}{\sqrt{1-\beta_1}}\E_{t+1-k}\left[\frac{(\tilde{\nabla}_i S_{t+1-k,w,\alpha}(\xx_{t+1-k}))^2}{\epsilon+\upsilon_{t+1,i}}\right]\\\nonumber
& \leq \frac{(\nabla_i S_{t+1-k,w,\alpha}(\xx_{t+1-k}))^2}{2\sqrt{\epsilon+\tilde{\upsilon}_{t+1,i}}}
+
\frac{2\sqrt{\mu}}{\sqrt{1-\beta_1}\beta_2^k}\E_{t+1-k}\left[\frac{(\tilde{\nabla}_i S_{t+1-k,w,\alpha}(\xx_{t+1-k}))^2}{\epsilon+\upsilon_{t+1-k,i}}\right]\\
& \leq \frac{(\nabla_i S_{t+1-k,w,\alpha}(\xx_{t+1-k}))^2}{2\sqrt{\epsilon+\tilde{\upsilon}_{t+1,i}}}
+
\frac{2\sqrt{\mu}}{\sqrt{1-\beta_1}\beta_2^k}\E_{t+1-k}\left[\|A_{t+1-k}\|^2\right],
\end{align}
where the second inequality is by
$$\epsilon+\upsilon_{t+1,i}\geq \epsilon+\beta_2^k\upsilon_{t+1-k,i} \geq  \beta_2^k(\epsilon+\upsilon_{t+1-k,i}).$$
Therefore, substituting \eqref{ggf} and \eqref{dss} into \eqref{app:eq:cool_plus_bad}, we then obtain

\begin{align}
\nonumber   \esp{I_1} &\leq \sum_{k=0}^{t-1}\beta_1^k \left(-\sum_{i=1}^d\frac{(\nabla_i S_{t+1-k,w,\alpha}(\xx_{t+1-k}) )^2}{\sqrt{\epsilon+\tilde{\upsilon}_{t+1,i}}}
\right.\\\nonumber&\qquad \qquad\quad\left.+  \sum_{i=1}^d\frac{(\nabla_i S_{t+1-k,w,\alpha}(\xx_{t+1-k}))^2}{2\sqrt{\epsilon+\tilde{\upsilon}_{t+1,i}}}
     +\frac{2\sqrt{\mu}}{\sqrt{1-\beta_1}\beta_2^k}\E\left[\|A_{t+1-k}\|^2\right]
    \right)\\
    &=- \sum_{i=1}^d \sum_{k=0}^{t-1}\beta_1^k\frac{(\nabla_i S_{t+1-k,w,\alpha}(\xx_{t+1-k}) )^2}{2\sqrt{\epsilon+\tilde{\upsilon}_{t+1,i}}}
 +\sum_{k=0}^{t-1}
    \frac{2 \sqrt{\mu}}{\sqrt{1-\beta_1}}
     \bbratio^k  \esp{\|A_{t+1 -k}\|^2}
    \label{app:eq:descent_proof:almost_there}.
\end{align}

\end{itemize}
Finally, injecting \eqref{app:eq:descent_proof:almost_there} and \eqref{app:eq:B_bound_second} into \eqref{app:eq:descent_big} gives the desired result.
\end{proof}

\subsubsection*{ PROOF OF THEOREM~\ref{adam}}

\begin{thm*}(\textsc{Adam})
Suppose Assumptions \ref{ass:sg}, \ref{ii} and \ref{ass:loss} hold. Let $\textsc{Dts-Ag}$ be the algorithm $\mathcal{A}$ in Algorithm~\ref{algo:mlol} with parameters $\eta_{t+1} = \eta (1 - \beta_1) \sqrt{\sum_{j=0}^{t} \beta_2^j}$ with $0<\beta_2<1$, $\eta>0$, $0<\beta_1<\beta_2$, and $\alpha \rightarrow 1^-$.
Furthermore, let $\sqrt{\sum_{r=0}^{t} \beta_2^r}\geq \frac{\varsigma}{\sqrt{1-\beta_2}}$ for some $\varsigma>0$ and $t\in [T]$.
Then, for any $\delta \in (0,1)$ and $S_{t,w,\alpha}(\xx_t)$ in \eqref{def}, with probability at least $1-\delta$,  the iterates $\xx_t$ satisfy the following bound
\begin{align*}
\sum^T_{t=1}\esp{\|\nabla S_{t,w,\alpha}(\xx_t)\|^2}
&\le
\frac{\sqrt{1-\beta_2}}{\varsigma\eta(1-\beta_1)}\big(\frac{4C\sqrt{\epsilon}}{\delta}+\frac{8C\sqrt{\zeta T}}{\delta^{3/2}}\big)+\frac{48(1-\beta_2)C^2}{\varsigma^2\eta^2(1-\beta_1)^2\delta^2}.
\end{align*}
Here, $C \triangleq \varpi_1 + \varpi_2 \left(d\ln\big(1 + \frac{2(\zeta+L'^2)}{d \epsilon ( 1- \beta_2)}\big) - T \ln(\beta_2)\right)$,
where
\begin{align*}
\varpi_1&\triangleq\frac{4D T}{W}+\frac{8T\eta (1-\beta_1)L'^2}{\beta_1\sqrt{(1-\beta_2)}W^2},\qquad \qquad \zeta \triangleq\frac{\sigma^2}{W},\\
    \varpi_2 &\triangleq \frac{ d \eta^2(1- \beta_1)\gamma'}{2(1 - \beta_2)(1 - \beta_1 / \beta_2)} +
     \frac{d \eta^3  \gamma'^2 \beta_1}{(1 - \beta_1 / \beta_2) (1 - \beta_2)^{3/2}}
     \\&\qquad +
     \frac{ 2d \eta (1+\sqrt{\zeta}) \sqrt{1-\beta_1}}{(1 - \beta_1 / \beta_2)^{3/2}\sqrt{1 - \beta_2}}
     +\frac{2\eta^3(1-\beta_1)^2\gamma'^2}{\beta_1 (1-\beta_2)^{3/2}(1 - \beta_1 / \beta_2)}.
\end{align*}
\end{thm*}

\begin{proof}
By the $\gamma'$-smoothness of $\ell_t$ functions, $S_t$ is $\gamma'$-smooth as well. Hence, we have
\begin{align*}
    S_{t,w,\alpha}(\xx_{t+1})-S_{t,w,\alpha}(\xx_{t})  &\leq  - \eta_{t+1} \langle \nabla S_{t,w,\alpha}(\xx_t) , \frac{\mm_{t+1}}{\sqrt{\epsilon+\up_{t+1}}} \rangle
 + \frac{\eta_{t+1}^2 \gamma'}{2}\norm{\frac{\mm_{t+1}}{\sqrt{\epsilon+\up_{t+1}}}}^2.
\end{align*}
Taking expectation on both sides of the above inequality, and using Lemma~\ref{app:lemma:descent_lemma}, we can get
\begin{align}
\nonumber
    \esp{S_{t,w,\alpha}(\xx_{t+1})}  \leq &\esp{S_{t,w,\alpha}(\xx_{t})}
                  - \eta_{t+1}
                  \sum_{i=1}^d \sum_{k=0}^{t-1}\beta_1^k\frac{(\nabla_i S_{t+1-k,w,\alpha}(\xx_{t+1-k}) )^2}{2\sqrt{\epsilon+\tilde{\upsilon}_{t+1,i}}}
 \\\nonumber
   & \quad  +
     \frac{ \eta_{t+1}}{\sqrt{1-\beta_1}} \esp{\sum_{k=0}^{t-1}
     \bbratio^k (\sqrt{k+1}+2\sqrt{\mu}) \norm{A_{t+1 -k}}^2}
     \\\nonumber&\quad+\frac{\eta_{t+1}^3 \gamma'^2}{4}\sqrt{1 - \beta_1} \esp{\sum_{l=1}^{t}\norm{B_{t+1-l}}^2}\sum_{k=l}^{t }\beta_1^k \sqrt{k}
     \\&\quad
     + \eta_{t+1}\sum_{k=0}^{t-1}\beta_1^k\frac{\sqrt{1-\beta_1}\vartheta_t}{2\sqrt{k+1}}
   + \frac{\eta_{t+1}^2 \gamma'}{2}\esp{\norm{B_{t+1}}^2},
 \label{app:eq:common_first}
\end{align}
where $A_t$, $B_t$, and $\vartheta_t$ are defined as in \eqref{cvbb} and \eqref{cvbbb} and $\tilde{\upsilon}_{t+1,i}\triangleq \beta_2\upsilon_{t,i} + (\nabla_i S_{t,w,\alpha}(\xx_t))^2 + \mu$.
Hence, rearranging the above inequality, and using the fact that $\eta_{t+1}$ is non-decreasing, we obtain:

\begin{align}\label{kk}
\nonumber &\sum_{t=1}^{T} \eta_{t+1}
                  \sum_{i=1}^d \sum_{k=0}^{t-1}\beta_1^k\frac{(\nabla_i S_{t+1-k,w,\alpha}(\xx_{t+1-k}) )^2}{2\sqrt{\epsilon+\tilde{\upsilon}_{t+1,i}}}
                  \leq  \underbrace{\sum_{t=1}^{T} \E[S_{t,w,\alpha}(\xx_{t})-S_{t,w,\alpha}(\xx_{t+1})]}_{I_1}\\\nonumber
                  &+ \underbrace{\frac{ \eta_{T+1}}{\sqrt{1-\beta_1}}\esp{\sum_{t=1}^T \sum_{k=0}^{t-1}
     \bbratio^k (\sqrt{k+1}+2\sqrt{\mu}) \norm{A_{t+1 -k}}^2}}_{I_2}
     + \underbrace{\frac{\eta_{T+1}^2 \gamma'}{2}\esp{\sum_{t=1}^{T}\norm{B_{t+1}}^2}}_{I_3}\\&+ \underbrace{\eta_{T+1}\sum_{t=1}^{T}\sum_{k=0}^{t-1}\beta_1^k\frac{\sqrt{1-\beta_1}\vartheta_t}{2\sqrt{k+1}}}_{I_4}
                 +\underbrace{\frac{\eta_{T+1}^3 \gamma'^2}{4}\sqrt{1 - \beta_1} \esp{\sum_{t=1}^{T} \sum_{l=1}^{t}\norm{B_{t+1-l}}^2}\sum_{k=l}^{t }\beta_1^k \sqrt{k} }_{I_5}.
\end{align}
We bound the term $I_1$ in the following manner:
%We first give an upper bound of the term $I_1$ as below
\begin{subequations}\label{bbbv}
\begin{align}
\nonumber I_1&=
\sum^{T}_{t=1}\E[S_{t,w,\alpha}(\xx_t)-S_{t+1,w,\alpha}(\xx_{t+1})]
+\sum^{T}_{t=1}\E[S_{t+1,w,\alpha}(\xx_{t+1})-S_{t,w,\alpha}(\xx_{t+1})]\\
&\leq \frac{2D(1-\alpha^w)T}{W(1-\alpha)}+\frac{D(1+\alpha^{w-1})T}{W}
+\frac{D(1-\alpha^{w-1})(1+\alpha)T}{W(1-\alpha)}
\label{eq:cons}~,
\end{align}
where the inequality follows from Lemmas~\ref{lm:sumobj22} and \ref{lm:sumobj2}.
\\
We next bound the term $I_2$, by the change of index $j = t+1-k$ as follows:

\begin{align}\label{app:eq:common_d_bound}
\nonumber
    I_2 &=\tilde{\eta} \esp{ \sum_{t=1}^T \sum_{j=2}^{t+1} \bbratio^{t+1 -j } \Big(\sqrt{2 + t - j}+2\sqrt{\mu}\Big) \norm{A_j}^2}\\
\nonumber
        &= \tilde{\eta}\esp{ \sum_{j=2}^T \norm{A_j}^2
       \sum_{t=j-1}^T \bbratio^{t+1 -j } \Big( \sqrt{2 + t - j}+2\sqrt{\mu}\Big)}\\\nonumber
        &\leq \tilde{\eta}\esp{\sum_{j=2}^T \norm{A_j}^2} \Big(\frac{2}{(1 - \beta_1 / \beta_2)^{3/2}}+\frac{2\sqrt{\mu}}{(1 - \beta_1 / \beta_2)}\Big)\\
        & \leq  \frac{2 \tilde{\eta}  (1+\sqrt{\mu}) }{(1 - \beta_1 / \beta_2)^{3/2}}\esp{\sum_{i=1}^d \Big( \ln\big(1 + \frac{\sum^{T}_{t=1}\beta_2^{T-t}(\tilde{\nabla}_i S_{t,w,\alpha}(\xx_t) )^2}{\eps}\big) - T \ln(\beta_2)\Big)},
\end{align}
where $\tilde{\eta}\triangleq\frac{ \eta_{T+1} }{\sqrt{1-\beta_1}}$, the first inequality is by Lemma~\ref{app:lemma:sum_geom_sqrt} and the second inequality follows from Lemma~\ref{app:lemma:sum_ratio} with $\upsilon_{t+1,i}=\sum_{j=1}^{t}\beta_2^{t-j}(\tilde{\nabla}_iS_{j,w,\alpha}(\xx_j))^2$, $\upsilon_{1,i}=0$ and $\beta_1 = 0$.\\
Using Lemma~\ref{app:lemma:sum_ratio}, we have the following bound on $I_3$ in \eqref{kk}:
\begin{align}
   I_3 &\leq \frac{\eta_{T+1}^2 \gamma'}{2 (1 - \beta_1)(1 - \beta_1 / \beta_2)}\esp{\sum_{i=1}^d \Big( \ln \big(1 + \frac{\sum^{T}_{t=1}\beta_2^{T-t}(\tilde{\nabla}_iS_{t,w,\alpha}(\xx_t) )^2}{\eps}\big)- T \ln(\beta_2)\Big)}.
    \label{app:eq:common_b_bound}
\end{align}
Based on Lemma~\ref{bn}, for the term $I_4$ in \eqref{kk} we have
\begin{align}\label{I5}
\nonumber I_4&=\frac{\eta_{T+1}\sqrt{1-\beta_1}}{2}\sum_{t=1}^{T}\sum_{k=0}^{t-1}\frac{\beta_1^k \vartheta_t}{\sqrt{k+1}}=
\frac{\eta_{T+1}\sqrt{1-\beta_1}}{2}\sum_{t=1}^{T}\sum_{k=0}^{t-1}\frac{\beta_1^k }{\sqrt{k+1}}
\left(\frac{8L'^2 }{W^2}\right)
\\\nonumber &\qquad+\frac{\eta_{T+1}\sqrt{1-\beta_1}(1-\alpha^{w-2})\gamma'^2}{W^2(1-\alpha)}\underbrace{\sum_{t=1}^{T}\sum_{k=0}^{t-1}\frac{\beta_1^k }{\sqrt{k+1}}
\left( \sum_{r=1}^{w-1}  \alpha^{r-1} \|\eta_{t-r+2-k}B_{t-r+2-k} \|^2\right)}_{I_{41}}\\\nonumber
&\leq\frac{T\eta_{T+1} 8L'^2 }{\beta_1W^2}
+\frac{2\eta_{T+1}^3(1-\alpha^{w-2})^2\gamma'^2}{\beta_1 W^2(1-\alpha)^2(1-\beta_1)(1 - \beta_1 / \beta_2)}
\\&\qquad \qquad \qquad \esp{\sum_{i=1}^d \Big( \ln \big(1 + \frac{\sum^{T}_{t=1}\beta_2^{T-t}(\tilde{\nabla}_iS_{t,w,\alpha}(\xx_t) )^2}{\eps}\big)- T \ln(\beta_2)\Big)},
\end{align}
where the last inequality holds since by changing index $j=t-k$, we have
\begin{align*}
\nonumber & I_{41} \leq  \sum_{r=1}^{w-1}  \alpha^{r-1} \sum_{t=1}^{T} \sum_{j=1}^{t}\frac{\beta_1^{t-j} }{\sqrt{t-j+1}}\|\eta_{j-r+2}B_{j-r+2} \|^2\\
 \nonumber &=  \sum_{r=1}^{w-1}  \alpha^{r-1} \sum_{j=1}^{T} \|\eta_{j-r+2}B_{j-r+2} \|^2\sum_{t=j}^{T}\frac{\beta_1^{t-j} }{\sqrt{t-j+1}}\stackrel{(i)}\leq  \frac{2\eta_{T+1}^2}{\beta_1\sqrt{1-\beta_1}}\sum_{r=1}^{w-1}  \alpha^{r-1} \sum_{j=1}^{T} \|B_{j-r+2} \|^2 \\\nonumber
  &\leq  \frac{2\eta_{T+1}^2}{\beta_1\sqrt{1-\beta_1}}\sum_{r=1}^{w-1}  \alpha^{r-1} \sum_{j=1}^{T} \|B_{j+1} \|^2\leq \frac{2\eta_{T+1}^2(1-\alpha^{w-2})}{\beta_1\sqrt{1-\beta_1}(1-\alpha)} \sum_{j=1}^{T} \|B_{j+1} \|^2\\
  &\stackrel{(ii)}\leq\frac{ 2\eta_{T+1}^2(1-\alpha^{w-2})}{\beta_1(1-\alpha) (1 - \beta_1)^{3/2}(1 - \beta_1 / \beta_2)}\esp{\sum_{i=1}^d \Big( \ln \big(1 + \frac{\sum^{T}_{t=1}\beta_2^{T-t}(\tilde{\nabla}_iS_{t,w,\alpha}(\xx_t) )^2}{\eps}\big)- T \ln(\beta_2)\Big)},
\end{align*}
where (i) follows from Lemma~\ref{bn} and (ii) follows from Lemma~\ref{app:lemma:sum_ratio}.

According to the change of index $j  =t+1 -k$, we bound the term $I_5$ as follows:
\begin{align}\label{app:eq:common_c_bound}
\nonumber
   I_5 &= \frac{\eta_{T+1}^3 \gamma'^2}{4 }\sqrt{1-\beta_1}  \esp{\sum_{t=1}^T \sum_{j=2}^{t+1}
    \norm{B_{j}}^2\sum_{k=t+1 - j}^{t} \beta_1^k \sqrt{k}} \\
\nonumber
    &= \frac{\eta_{T+1}^3 \gamma'^2}{4 }\sqrt{1-\beta_1} \esp{ \sum_{j=1}^T \norm{B_{j}}^2\sum_{t=j-1}^{T}
    \sum_{k=t+1 - j}^{t} \beta_1^k \sqrt{k}}\\
\nonumber
    &= \frac{\eta_{T+1}^3 \gamma'^2}{4 }\sqrt{1-\beta_1}  \esp{ \sum_{j=1}^T\norm{B_{j}}^2
    \sum_{k=0}^{T-1} \beta_1^k \sqrt{k}\sum_{t=j-1}^{j + k} 1}\\
\nonumber
    &= \frac{\eta_{T+1}^3 \gamma'^2}{4 }\sqrt{1-\beta_1}\esp{ \sum_{j=1}^T \norm{B_{j}}^2
    \sum_{k=0}^{T-1} \beta_1^k \sqrt{k} (k + 1)}\leq \eta_{T+1}^3 \gamma'^2 \esp{ \sum_{j=1}^T \norm{B_{j}}^2 \frac{\beta_1}{(1 - \beta_1)^{2}}}\\
    & \leq  \frac{\eta_{T+1}^3  \gamma'^2 \beta_1}{ (1 - \beta_1)^{3}(1 - \beta_1/\beta_2)}
    \E\left[ \sum_{i=1}^d \Big( \ln \big(1
    + \frac{\sum^{T}_{t=1}\beta_2^{T-t}(\tilde{\nabla}_iS_{t,w,\alpha}(\xx_t) )^2}{\eps}\big)
     - T \ln(\beta_2)\Big)\right],
\end{align}
where the first inequality is due to the Lemma~\ref{app:lemma:sum_geom_32} and the second inequality holds due to Lemma~\ref{app:lemma:sum_ratio}.\\
\end{subequations}

What's more, we have
 \begin{align}\label{hhj}
\nonumber &\E\left[\sum_{i=1}^{d}\ln\big(1 +\frac{\sum^{T}_{t=1}\beta_2^{T-t}(\tilde{\nabla}_iS_{t,w,\alpha}(\xx_t) )^2}{\epsilon} \big)\right]\stackrel{(i)} \leq \E\left[d\ln\Big(\frac{1}{d}\sum_{i=1}^{d}\big(1 +\frac{\sum^{T}_{t=1}\beta_2^{T-t}(\tilde{\nabla}_iS_{t,w,\alpha}(\xx_t) )^2}{\epsilon} \big)\Big)\right]\\\nonumber
 &= \E\left[d\ln \big(1 +\frac{\sum^{T}_{t=1}\beta_2^{T-t}\sum_{i=1}^{d}(\tilde{\nabla}_iS_{t,w,\alpha}(\xx_t) )^2}{\epsilon d} \Big)\right]\stackrel{(ii)} \le
d\ln\Big(1 +\frac{\sum^{T}_{t=1}\beta_2^{T-t}\E\left[\|\tilde{\nabla}S_{t,w,\alpha}(\xx_t) \|^2\right]}{\epsilon d} \Big)
\\\nonumber &\stackrel{(iii)} \le
d\ln\Big(1 +\frac{2\sum^{T}_{t=1}\beta_2^{T-t}\E\left[\|\tilde{\nabla}S_{t,w,\alpha}(\xx_t)-\nabla S_{t,w,\alpha}(\xx_t) \|^2\right]
+2\sum^{T}_{t=1}\beta_2^{T-t}\E\left[\|\nabla S_{t,w,\alpha}(\xx_t) \|^2\right]}{\epsilon d} \Big)
\\&\stackrel{(iv)}\le d\ln\Big(1 +\frac{2(\mu+L'^2)}{d\epsilon(1-\beta_2)} \Big)~,
\end{align}
where (i) is due to the convex inequality
$\frac{1}{d}\sum_{i=1}^{d}\ln (a_i)\leq \ln (\frac{1}{d}\sum_{i=1}^{d} a_i)$, (ii) follows from
$\E [ \ln(A)]\leq \ln \E[A]$, for any positive random variable $A$, (iii) follows from $\| a+b\|^2\leq 2\|a \|^2+2\|b\|^2$, (iv) follows from Lemmas~\ref{lm:property} and
\ref{p0}(b).

Substituting \eqref{eq:cons}-\eqref{app:eq:common_c_bound} into \eqref{kk} and using \eqref{hhj} as well as the fact that
$\eta_{T+1}\leq \eta \frac{1-\beta_1}{\sqrt{1-\beta_2}}, $
we obtain

\begin{align}\label{kkb}
 &\sum_{t=1}^{T}\eta_{t+1}\sum_{i=1}^d \sum_{k=0}^{t-1}\beta_1^k\frac{(\nabla_i S_{t+1-k,w,\alpha}(\xx_{t+1-k}))^2}{2\sqrt{\epsilon+\tilde{\upsilon}_{t+1,i}}}
                  \leq
    \nu + u \left(d\ln\Big(1 + \frac{2(\mu+L'^2)}{d\epsilon ( 1- \beta_2)}\Big) - T \ln(\beta_2)\right)\triangleq C,
\end{align}
where
\begin{align}
\label{24}  \nu&\triangleq\frac{2D(1-\alpha^w)T}{W(1-\alpha)}+\frac{D(1+\alpha^{w-1})T}{W}
    +\frac{D(1-\alpha^{w-1})(1+\alpha)T}{W(1-\alpha)}+\frac{T\eta (1-\beta_1) 8L'^2 }{\sqrt{1-\beta_2}\beta_1W^2},\\\nonumber
    u &\triangleq \frac{ d \eta^2(1- \beta_1)\gamma'}{2(1 - \beta_2)(1 - \beta_1 / \beta_2)} +
     \frac{d \eta^3  \gamma'^2 \beta_1}{(1 - \beta_1 / \beta_2) (1 - \beta_2)^{3/2}}\\&\qquad+
     \frac{2 d \eta (1+\sqrt{\mu}) \sqrt{1-\beta_1}}{(1 - \beta_1 / \beta_2)^{3/2}\sqrt{1 - \beta_2}}
     +\frac{2\eta^3(1-\beta_1)^2(1-\alpha^{w-2})^2\gamma'^2}{\beta_1 W^2(1-\alpha)^2(1-\beta_2)^{3/2}(1 - \beta_1 / \beta_2)}\label{25}.
\end{align}
By \eqref{kkb} and Markov's inequality, we have, with probability at least $1-\delta_1$,
\begin{equation}\label{41}
\sum_{t=1}^{T}\eta_{t+1}\sum_{i=1}^d \sum_{k=0}^{t-1}\beta_1^k\frac{(\nabla_i S_{t+1-k,w,\alpha}(\xx_{t+1-k}))^2}{2\sqrt{\epsilon+\tilde{\upsilon}_{t+1,i}}}
\le
\frac{C}{\delta_1}~.
\end{equation}
From Lemma~\ref{p0}(b), observe that, with probability at least $1-\delta_2$, Markov's inequality implies that
\begin{equation*}
\sum_{t=1}^{T}\|\nabla S_{t,w,\alpha}(\xx_t) -\tilde{\nabla}S_{t,w,\alpha}(\xx_t)\|^2\le\frac{T\mu}{\delta_2}~,
\end{equation*}
where $\mu\triangleq\frac{\sigma^2(1-\alpha^{2w})}{W^2(1-\alpha^2)}$. In addition, let us denote $Z\triangleq \sum_{t=1}^{T}\sum_{i=1}^{d}(\nabla_i S_{t,w,\alpha}(\xx_t))^2$.
Using the above inequality along with the bound $\|a+b\|^2\leq 2\|a \|^2+2\| b\|^2$, with probability $1-\delta_2$, we also have
\begin{align}\label{vg}
\sum^{T}_{t=1}\|\tilde{\nabla}S_{t,w,\alpha}(\xx_t)\|^2
\nonumber&\le
2\sum^{T}_{t=1}\|\tilde{\nabla}S_{t,w,\alpha}(\xx_t) - \nabla S_{t,w,\alpha}(\xx_t)\|^2
+ 2\sum^{T}_{t=1}\|\nabla S_{t,w,\alpha}(\xx_t)\|^2
\\&\le
\frac{2  T\mu }{\delta_2}
+ 2\sum^{T}_{t=1}\|\nabla S_{t,w,\alpha}(\xx_t)\|^2.
\end{align}
Now, recalling that $\up_{t+1}=\beta_2 \up_{t}+(\tilde{\nabla}S_{t,w,\alpha}(\xx_t))^2=\sum_{l=1}^{t}\beta_2^{t-l}(\tilde{\nabla}S_{l,w,\alpha}(\xx_l))^2$, $\up_1=0$.
Then, by the bound \eqref{vg}, it follows that with probability at least $1-\delta_2$,
\begin{align}\label{mk}
\nonumber &\qquad \beta_2\sum_{i=1}^{d}\upsilon_{T,i} +\sum_{i=1}^{d} (\nabla_i S_{T,w,\alpha}(\xx_{T}))^2 + \mu \\
\nonumber &=\sum_{i=1}^{d}\sum_{t=1}^{T-1}\beta_2^{T-t}(\tilde{\nabla}_i S_{t,w,\alpha}(\xx_t))^2
+\sum_{i=1}^{d}(\nabla_i S_{T,w,\alpha}(\xx_{T}))^2+ \mu\\
\nonumber &\leq\sum_{i=1}^{d}\sum_{t=1}^{T-1}(\tilde{\nabla}_i S_{t,w,\alpha}(\xx_t))^2
+\sum_{i=1}^{d}(\nabla_i S_{T,w,\alpha}(\xx_{T}))^2+ \mu
\\ \nonumber&\leq \frac{2(T-1)\mu}{\delta_2}
+ 2\sum^{T-1}_{t=1}\|\nabla S_{t,w,\alpha}(\xx_t)\|^2
+\sum_{i=1}^{d}(\nabla_i S_{T,w,\alpha}(\xx_{T}))^2+\mu
\\\nonumber &\leq \frac{2(T-1)\mu}{\delta_2}
+ 2\sum^{T-1}_{t=1}\|\nabla S_{t,w,\alpha}(\xx_t)\|^2
+2\sum_{i=1}^{d}(\nabla_i S_{T,w,\alpha}(\xx_{T}))^2+\frac{2\mu}{\delta_2}
\\&\le\
 2Z + \frac{2T\mu}{\delta_2},
\end{align}
where $Z\triangleq\sum^{T}_{t=1}\|\nabla S_{t,w,\alpha}(\xx_t)\|^2$.
Using the definition of $\tilde{\upsilon}_{t+1,i}$ and $\eta_{t+1}=\eta (1 - \beta_1) \sqrt{\sum_{r=0}^{t} \beta_2^r} $, with probability at least $1-\delta_2$, we obtain
\begin{align}\label{xf}
\nonumber &\qquad \sum_{t=1}^{T}\eta_{t+1}\sum_{i=1}^d \frac{\sum_{k=0}^{t-1}\beta_1^k(\nabla_i S_{t+1-k,w,\alpha}(\xx_{t+1-k}))^2}{2\sqrt{\epsilon+\tilde{\upsilon}_{t+1,i}}}
\\\nonumber&= \eta \sum_{t=1}^{T}  \sqrt{\sum_{r=0}^{t} \beta_2^r}\sum_{i=1}^d \frac{\sum_{k=0}^{t-1}(1 - \beta_1)\beta_1^k(\nabla_i S_{t+1-k,w,\alpha}(\xx_{t+1-k}))^2}{2\sqrt{\epsilon+\beta_2\upsilon_{t,i} + (\nabla_i S_{t,w,\alpha}(\xx_t))^2 + \mu}}\\\nonumber
&\ge
\eta \sum_{t=1}^{T} \sqrt{\sum_{r=0}^{t} \beta_2^r}\sum_{i=1}^d\frac{\sum_{k=0}^{t-1}(1 - \beta_1)\beta_1^k(\nabla_i S_{t+1-k,w,\alpha}(\xx_{t+1-k}))^2}
{2\sqrt{\epsilon+\beta_2\upsilon_{T,i}
+ (\nabla_i S_{T,w,\alpha}(\xx_T))^2 + \sum_{t=1}^{T}  (\nabla_i S_{t,w,\alpha}(\xx_t))^2+\mu}}
\\
&\ge
\frac{\varsigma\eta }{\sqrt{1-\beta_2}}\sum_{t=1}^{T} \sum_{i=1}^d\frac{\sum_{k=0}^{t-1}(1 - \beta_1)\beta_1^k(\nabla_i S_{t+1-k,w,\alpha}(\xx_{t+1-k}))^2}
{2\sqrt{\epsilon+\beta_2\sum_{i=1}^{d}\upsilon_{T,i} +\sum_{i=1}^{d} (\nabla_i S_{T,w,\alpha}(\xx_T))^2 + \sum_{t=1}^{T} \sum_{i=1}^{d} (\nabla_i S_{t,w,\alpha}(\xx_t))^2 +\mu}},
\end{align}
where the last inequality is due to our assumption that $\sqrt{\sum_{r=0}^{t} \beta_2^r}\geq \frac{\varsigma}{\sqrt{1-\beta_2}}$.
Now by changing of index $j = t +1-k$, for all $i\in [d]$ we have
\begin{align}\label{cv}
\nonumber &\sum_{t=1}^{T}\sum_{k=0}^{t-1}(1 - \beta_1)\beta_1^k (\nabla_i S_{t+1-k,w,\alpha}(\xx_{t+1-k}))^2
                  = \sum^{T}_{t=1} \sum_{j=2}^{t+1}(\nabla_i  S_{j,w,\alpha}(\xx_{j}))^2(1 - \beta_1)\beta_1^{t+1-j}\\\nonumber
                  &= \sum^{T}_{t=1} \sum_{j=1}^{t}(\nabla_i  S_{j+1,w,\alpha}(\xx_{j+1}))^2(1 - \beta_1)\beta_1^{t-j}
                  = \sum^{T}_{t=1} (\nabla_i  S_{t+1,w,\alpha}(\xx_{t+1}))^2\sum_{j=t}^{T}(1 - \beta_1)\beta_1^{j-t}
                  \\\nonumber&=\sum_{t=1}^{T} (1-\beta_1^{T+1-t})(\nabla_i S_{t+1,w,\alpha}(\xx_{t+1}))^2
                  \geq  (1-\beta_1)\sum_{t=2}^{T+1} (\nabla_i S_{t,w,\alpha}(\xx_{t}))^2
                  \\&\geq  (1-\beta_1)\sum_{t=2}^{T} (\nabla_i S_{t,w,\alpha}(\xx_{t}))^2
                  =  (1-\beta_1)\sum_{t=1}^{T} (\nabla_i S_{t,w,\alpha}(\xx_{t}))^2,
\end{align}
where the last equality is by assumption $\nabla_i S_{1,w,\alpha}(\xx_{1})=0$.
Combining \eqref{cv} with \eqref{xf} yields
\begin{align}\label{43}
\nonumber &\qquad \sum_{t=1}^{T}\eta_{t+1}\sum_{i=1}^d \frac{\sum_{k=0}^{t-1}\beta_1^k(\nabla_i S_{t+1-k,w,\alpha}(\xx_{t+1-k}))^2}{2\sqrt{\epsilon+\tilde{\upsilon}_{t+1,i}}}
\\\nonumber
&\ge
\frac{\varsigma\eta(1-\beta_1)\sum_{t=1}^{T}\sum_{i=1}^d(\nabla_i S_{t,w,\alpha}(\xx_{t}))^2}{2\sqrt{1-\beta_2}\sqrt{\epsilon+\beta_2\sum_{i=1}^{d}\upsilon_{T,i} +\sum_{i=1}^{d} (\nabla_i S_{T,w,\alpha}(\xx_T))^2 + Z +\mu}}\\
&\ge
\frac{\varsigma\eta(1-\beta_1)Z}{2\sqrt{1-\beta_2}\sqrt{\epsilon+ 3Z + \frac{2T\mu}{\delta_2}}}~,
\end{align}
where the last inequality follows from \eqref{mk}.\\

Considering equations \eqref{41} and \eqref{43}, we then observe that with probability $1-\delta_2-\delta_1$,
\begin{equation*}
\frac{\varsigma\eta(1-\beta_1)Z}{2\sqrt{1-\beta_2}\sqrt{\epsilon + 3Z + \frac{2T\mu}{\delta_2}}}
\le
\frac{C}{\delta_1}~.
\end{equation*}
In view of Lemma~\ref{lem:quadratic} and letting $\delta_1=\delta_2=\frac{\delta}{2}$, we get
\begin{align}\label{99}
 & \sum^T_{t=1}\esp{\|\nabla S_{t,w,\alpha}(\xx_t)\|^2}\le
\frac{\sqrt{1-\beta_2}}{\varsigma\eta(1-\beta_1)}\Big(\frac{4C\sqrt{\epsilon}}{\delta}+\frac{8C\sqrt{\mu T}}{\delta^{3/2}}\Big)
+\frac{48(1-\beta_2)C^2}{\varsigma^2\eta^2(1-\beta_1)^2\delta^2}.
\end{align}
Here, $C \triangleq \nu + u \left(d\ln\big(1 + \frac{2(\mu+L'^2)}{d\epsilon ( 1- \beta_2)}\big) - T \ln(\beta_2)\right)$, and $\mu\triangleq\frac{\sigma^2(1-\alpha^{2w})}{W^2(1-\alpha^2)},$
where $\nu$ and $u$ are defined in \eqref{24} and \eqref{25}, respectively.
Further, we can decompose
\begin{align}\label{vdd}
 \nu&=\nu_1+\frac{T\eta (1-\beta_1) 8L'^2 }{\sqrt{1-\beta_2}\beta_1W^2},
\end{align}
where
\begin{align*}
\nu_1&=\frac{2D(1-\alpha^w)T}{W(1-\alpha)}+\frac{D(1+\alpha^{w-1})T}{W}+\frac{D(1-\alpha^{w-1})(1+\alpha)T}{W(1-\alpha)}.
\end{align*}
Now use a similar argument as in equation \eqref{frd1} in the term $\nu_1$ to have
\begin{align}\label{frd}
 \nu_1&\leq
\frac{4D T}{W}\Big(\frac{2-\alpha^{w}+\alpha^{w-1}}{1-\alpha}\Big).
\end{align}
Then plugging \eqref{frd} into \eqref{vdd} yields
\begin{align*}
\nu &\leq \frac{4D T}{W}\Big(\frac{2-\alpha^{w}+\alpha^{w-1}}{1-\alpha}\Big)+\frac{T\eta (1-\beta_1) 8L'^2 }{\sqrt{1-\beta_2}\beta_1W^2}.
\end{align*}
As $\alpha \rightarrow 1^{-}$, we get
\begin{equation}
\begin{aligned}\label{nnb}
  &\nu \leq \frac{4D T}{W}+\frac{T\eta (1-\beta_1) 8L'^2 }{\sqrt{1-\beta_2}\beta_1W^2}\triangleq\varpi_1,\qquad
  \mu=\frac{\sigma^2(1-\alpha^{2w})}{W^2(1-\alpha^2)} \stackrel{\alpha \rightarrow 1^{-}}=\frac{\sigma^2}{W}\triangleq\zeta,\\
&  u \stackrel{\alpha \rightarrow 1^{-}}= \frac{ d \eta^2(1- \beta_1)\gamma'}{2(1 - \beta_2)(1 - \beta_1 / \beta_2)} +
     \frac{d \eta^3  \gamma'^2 \beta_1}{(1 - \beta_1 / \beta_2) (1 - \beta_2)^{3/2}}\\&\qquad+
     \frac{2 d \eta (1+\sqrt{\mu}) \sqrt{1-\beta_1}}{(1 - \beta_1 / \beta_2)^{3/2}\sqrt{1 - \beta_2}}
     +\frac{2\eta^3(1-\beta_1)^2\gamma'^2}{\beta_1 (1-\beta_2)^{3/2}(1 - \beta_1 / \beta_2)}\triangleq \varpi_2.
\end{aligned}
\end{equation}
The desired result then follows by inserting \eqref{nnb} to \eqref{99}.
\end{proof}

\subsubsection*{ PROOF OF COROLLARY~\ref{corr11}}
\begin{cor*}
Under the same conditions stated in Theorem \ref{adam}, using $\beta_2=1-1/T$, $\eta=\eta_1/\sqrt{T}$, $\beta_1/\beta_2\approx \beta_1$, $w\in\Theta(T)$, and $\alpha\rightarrow 1^{-}$ yields a regret bound of order
\begin{equation*}
\sum^T_{t=1}\esp{\|\nabla S_{t,w,\alpha}(\xx_t)\|^2}\leq \mathcal{O}(\ln T).
\end{equation*}
\end{cor*}
\begin{proof}
By Theorem \ref{adam}, we have
\begin{equation*}
\sum^T_{t=1}\esp{\|\nabla S_{t,w,\alpha}(\xx_t)\|^2}
\le \frac{\sqrt{1-\beta_2}}{\varsigma\eta(1-\beta_1)}\Big(\frac{4C\sqrt{\epsilon}}{\delta}+\frac{8C\sqrt{\zeta T}}{\delta^{3/2}}\Big)+\frac{48(1-\beta_2)C^2}{\varsigma^2\eta^2(1-\beta_1)^2\delta^2}\triangleq I_1+I_2+I_3~,
\end{equation*}
where $C = \varpi_1 + \varpi_2 \left(d\ln\big(1 + \frac{2(\zeta+L'^2)}{d\epsilon ( 1- \beta_2)}\big) - T \ln(\beta_2)\right)$, $\zeta = \frac{\sigma^2}{W}$,  $\varpi_1$ and $\varpi_2$ are defined as in \eqref{53} and \eqref{6}, respectively.
Recall that $W= \sum_{r=0}^{w-1}\alpha^r$. As $\alpha\rightarrow 1^{-}$ and $w\in\Theta(T)$, by setting $\beta_2=1-1/T$, $\eta=\eta_1/\sqrt{T}$, $\beta_1/\beta_2\approx \beta_1$ we get the following equalities:
\begin{align*}
\varpi_1&=\frac{4D T}{W}+\frac{8T\eta_1 (1-\beta_1)L'^2}{\beta_1W^2}=\mathcal{O}(1+\frac{1}{T}),\\
    \varpi_2 &\approx \frac{ d \eta_1^2\gamma'}{2} +
     \frac{d \eta_1^3  \gamma'^2 \beta_1 }{(1 - \beta_1 )} +
     \frac{ 2d \eta_1 (1+\sqrt{\frac{\sigma^2}{W}}) }{(1 - \beta_1 )}
     +\frac{2\eta_1^3(1-\beta_1)\gamma'^2}{\beta_1 }=\mathcal{O}(1+\frac{1}{\sqrt{T}}),\\
C&=\varpi_1 + \varpi_2 \big(d\ln\big(1 + \frac{2(\frac{\sigma^2}{W}+L'^2)T}{d\epsilon }\big) - T \ln(1-1/T)\big)\\
&=\varpi_1 + \varpi_2 \left(d\ln\big(1 + \frac{2(\frac{\sigma^2}{W}+L'^2)T}{d\epsilon }\big) +1\right)=\mathcal{O}(\ln T).
\end{align*}
This implies that
\begin{align*}
  I_1&=\frac{\sqrt{1-\beta_2}4C\sqrt{\epsilon}}{\varsigma\eta(1-\beta_1)\delta}=\frac{4C\sqrt{\epsilon}}{\varsigma\eta_1(1-\beta_1)\delta} =\mathcal{O}(\ln T),\\
  I_2&=\frac{\sqrt{1-\beta_2}8C\sqrt{\zeta T}}{\varsigma \eta (1-\beta_1)\delta^{3/2}} =\frac{8C\sqrt{\frac{\sigma^2}{W} T}}{\varsigma \eta_1 (1-\beta_1)\delta^{3/2}}=\mathcal{O}(\ln T), \qquad
  I_3=\frac{48(1-\beta_2)C^2}{\varsigma^2 \eta^2 (1-\beta_1)^2\delta^2}=\mathcal{O}(\ln T).
\end{align*}
Combine the above results we can easily have the desired result.
\end{proof}

\subsection*{Details of Section \ref{sec6}}
\subsubsection*{ PROOF OF LEMMA~\ref{4543}}

\begin{lem*}
Suppose  Assumptions~\ref{ass:sg} and \ref{212} hold. Then, for $\tilde{\nabla}S_{t,w,\alpha}(\xx_t)$ in the Algorithm $\textsc{Dts-Ag}$, any $\delta\in(0,1)$, and $S_{t,w,\alpha}(\xx_t)$ in \eqref{def}, with probability at least $1-\delta$, we have
\begin{enumerate}
\item [(a)] $\E_{t}\left[\tilde{\nabla}S_{t,w,\alpha}(\xx_t)\right]
=\nabla S_{t,w,\alpha}(\xx_t),$\label{ass:unbiase}
\item [(b)]$\max_{1\leq t \leq T}\left\|\tilde{\nabla}S_{t,w,\alpha}(\xx_t)-\nabla S_{t,w,\alpha}(\xx_t)\right\|^2
\le
\kappa^2\ln \frac{\exp\left(\frac{w\sum^{w-1}_{r=0}\alpha^{2r}}{W^2}\right)}{\delta}\triangleq \bar{\mu}.$
\end{enumerate}

\end{lem*}

\begin{proof}
Similar to the proof for Lemma \ref{p0}(a), we can show that part (a) of the Lemma. For part (b),
for any $\Gamma>0$, we observe that
\begin{align*}
\nonumber	\mathbb{P} &\left( \max_{1\leq t \leq T}\left\|\tilde{\nabla}S_{t,w,\alpha}(\xx_t)-\nabla S_{t,w,\alpha}(\xx_t)\right\|^2 > \Gamma  \right)
	\\\nonumber& = \mathbb{P} \left(\exp \left( \frac{\max_{1\leq t \leq T}\left\|\tilde{\nabla}S_{t,w,\alpha}(\xx_t)-\nabla S_{t,w,\alpha}(\xx_t)\right\|^2 }{\kappa^2} \right) > \exp \left( \frac{\Gamma}{\kappa^2}\right)  \right)\\\nonumber
	& \leq  \exp \left(-\frac{ \Gamma}{\kappa^2} \right) \E \left[ \exp \left(\frac{\max_{1\leq t \leq T}\left\|\tilde{\nabla}S_{t,w,\alpha}(\xx_t)-\nabla S_{t,w,\alpha}(\xx_t)\right\|^2}{\kappa^2}\right)\right]
\\\nonumber
	& \leq  \exp \left(-\frac{ \Gamma}{\kappa^2} \right) \E \left[ \exp \left(\frac{\max_{1\leq t \leq T}\left\|\sum^{w-1}_{r=0}\alpha^r\g_{t-r}(\xx_{t-r},\xi_{t,t-r}) -\alpha^r \nabla\ell_{t-r}(\xx_{t-r})\right\|^2}{W^2\kappa^2}\right) \right]
\\\nonumber
	&\leq \exp \left(-\frac{ \Gamma}{\kappa^2} \right) \E \left[ \exp \left(\frac{\max_{1\leq t \leq T} w\sum^{w-1}_{r=0}\alpha^{2r}\left\|\g_{t-r}(\xx_{t-r},\xi_{t,t-r}) - \nabla\ell_{t-r}(\xx_{t-r})\right\|^2}{W^2\kappa^2}\right) \right],
\end{align*}
where the first inequality is obtained by Markov's inequality, the second inequality holds because \eqref{def}, and the third inequality follows from $\|\sum_{i=1}^{n} a_i\|^2\leq n \sum_{i=1}^{n}\| a_i\|^2$. Then, we can write
	\begin{align*}
\nonumber	\mathbb{P} &\left( \max_{1\leq t \leq T}\left\|\tilde{\nabla}S_{t,w,\alpha}(\xx_t)-\nabla S_{t,w,\alpha}(\xx_t)\right\|^2 > \Gamma  \right)
\\\nonumber&\leq \exp \left(-\frac{ \Gamma}{\kappa^2} \right)
 \E \left[\exp \left(\frac{\max_{1\leq t \leq T} w\sum^{w-2}_{r=0}\alpha^{2r}\left\|\g_{t-r}(\xx_{t-r},\xi_{t,t-r}) - \nabla\ell_{t-r}(\xx_{t-r})\right\|^2}{W^2\kappa^2}\right)
\right.\\ &\left.\qquad\E \left[
\exp \left(\frac{\max_{1\leq t \leq T}w\alpha^{2(w-1)}\left\|\g_{t-w+1}(\xx_{t-w+1},\xi_{t,t-w+1}) - \nabla\ell_{t-w+1}(\xx_{t-w+1})\right\|^2}{W^2\kappa^2}\right)\right]
\right]
\\&\leq \exp \left(-\frac{ \Gamma}{\kappa^2} \right) \E \left[
\exp \left(\frac{\max_{1\leq t \leq T}w\sum^{w-2}_{r=0}\alpha^{2r}\left\|\g_{t-r}(\xx_{t-r},\xi_{t,t-r}) - \nabla\ell_{t-r}(\xx_{t-r})\right\|^2}{W^2\kappa^2}\right)
\right.\\ &\left.\qquad\left(\E \left[\exp \left( \frac{\max_{1\leq t \leq T}\left\|\g_{t-w+1}(\xx_{t-w+1},\xi_{t,t-w+1}) - \nabla\ell_{t-w+1}(\xx_{t-w+1})\right\|^2}{\kappa^2}\right)\right]\right)^{\frac{w\alpha^{2(w-1)}}{W^2}}
\right]\\&\leq \exp \left(-\frac{ \Gamma}{\kappa^2} \right) \E \left[ \exp \left(\frac{\max_{1\leq t \leq T}w\sum^{w-2}_{r=0}\alpha^{2r}\left\|\g_{t-r}(\xx_{t-r},\xi_{t,t-r}) - \nabla\ell_{t-r}(\xx_{t-r})\right\|^2}{W^2\kappa^2}\right) \left( \exp(1)\right)^{\frac{w\alpha^{2(w-1)}}{W^2}} \right],
	\end{align*}
where the last inequality is due to Assumption \ref{212}. We obtain from the above inequality that:
	\begin{align*}
\nonumber	&\mathbb{P} \left( \max_{1\leq t \leq T}\left\|\tilde{\nabla}S_{t,w,\alpha}(\xx_t)-\nabla S_{t,w,\alpha}(\xx_t)\right\|^2 > \Gamma  \right)
 \leq \exp \left(-\frac{ \Gamma}{\kappa^2} \right) \left( \exp(1)\right)^{\frac{w \sum_{r=0}^{w-1}\alpha^{2r}}{W^2}}\\
	&= \exp \left(-\frac{ \Gamma}{\kappa^2} \right)\exp \left(\frac{w \sum_{r=0}^{w-1}\alpha^{2r}}{W^2}\right)
 =  \exp \left(-\frac{\Gamma}{\kappa^2}+\frac{w\sum^{w-1}_{r=0}\alpha^{2r}}{W^2} \right) ~.
	\end{align*}
That completes the proof.
\end{proof}

\subsubsection*{ PROOF OF LEMMA~\ref{app:lemma:descent_lemmaa2}}
The proof relies on the following lemma.
\begin{lem}\label{lemma:sub_gaussian}\cite[Lemma~1]{li2020high}
	Assume that $Z_1, Z_2, ..., Z_T$ is a martingale difference sequence with respect to $\xi_1, \xi_2, ..., \xi_T$ and $\E_t \left[\exp(Z_t^2/\kappa_t^2)\right] \leq \exp(1)$ for all $1\leq t \leq T$, where $\kappa_t$ is a sequence of random variables with respect to $\xi_1, \xi_2, \dots, \xi_{t-1}$.
	Then, for any fixed $\lambda > 0$ and $\delta \in (0,1)$, with probability at least $1-\delta$, we have
	\[
	\sum_{t=1}^T Z_t \leq \frac{3}{4} \lambda \sum_{t=1}^T \kappa_t^2 + \frac{1}{\lambda} \ln \frac{1}{\delta}~.
	\]
\end{lem}
\begin{lem*}
Suppose Assumptions~\ref{ass:sg} and \ref{212} hold. Let $\tilde{\nabla}_iS_{t,w,\alpha}(\xx_t)$ and $\upsilon_{t+1,i}$ be the sequences defined in the Algorithm $\textsc{Dts-Ag}$. Then, for any $\delta \in (0,1)$, $0<\alpha<1$, $w$, $\epsilon>0$ and $S_{t,w,\alpha}(\xx_t)$ in \eqref{def}, with probability at least $1-\delta$, we have

\begin{align*}
\nonumber
  \sum_{t=1}^{T} \sum_{i =1}^d \langle \nabla_i S_{t,w,\alpha}(\xx_t) , \frac{\tilde{\nabla}_iS_{t,w,\alpha}(\xx_t)}{\sqrt{\epsilon+\upsilon_{t+1,i}}} \rangle
& \geq \sum_{i=1}^{d}\sum_{t=1}^{T}\frac{(\nabla_i S_{t,w,\alpha}(\xx_t))^2}{4\sqrt{\epsilon+\tilde{\upsilon}_{t+1,i}}}
-2\sqrt{\frac{\bar{ \mu}}{W}}
\sum_{i=1}^{d}\sum_{t=1}^{T}\frac{(\tilde{\nabla}_iS_{t,w,\alpha}(\xx_t))^2}{\epsilon+\upsilon_{t+1,i}}\\
 &\quad-\frac{3(1-\alpha^w)^2\kappa^2}{W^2(1-\alpha)^2\sqrt{\epsilon}}\ln \frac{1}{\delta},
\end{align*}
where $\bar{\mu}=\kappa^2\ln \frac{\exp\left(\frac{w\sum^{w-1}_{r=0}\alpha^{2r}}{W^2}\right)}{\delta}$,  $\tilde{\upsilon}_{t+1,i}\triangleq \frac{1}{W}\big(\upsilon_{t,i} + (\nabla_i S_{t,w,\alpha}(\xx_t))^2 \big)+ \hat{\mu}_{T,i}$, and\\
$\hat{\mu}_{T,i}\triangleq  \frac{1}{W}\left(\tilde{\nabla}_i S_{T,w,\alpha}(\xx_T)-\nabla_i S_{T,w,\alpha}(\xx_T)\right)^2$ for all $i\in [d]$.
\end{lem*}

\begin{proof}
By definition of $\tilde{\upsilon}_{t+1,i}$, we decompose the LHS as
\begin{align}\label{882}
\nonumber &-\sum_{t=1}^{T}  \sum_{i =1}^d \langle \nabla_i S_{t,w,\alpha}(\xx_t) , \frac{\tilde{\nabla}_iS_{t,w,\alpha}(\xx_t)}{\sqrt{\epsilon+\upsilon_{t+1,i}}} \rangle
 \\& =\sum_{i=1}^{d}\Big(-\sum_{t=1}^{T}\frac{(\nabla_i S_{t,w,\alpha}(\xx_t))^2}{\sqrt{\epsilon+\tilde{\upsilon}_{t+1,i}}}+\sum_{t=1}^{T}I_1
 -\sum_{t=1}^{T}\langle \nabla_i S_{t,w,\alpha}(\xx_t),\frac{\tilde{\nabla}_i S_{t,w,\alpha}(\xx_t)-\nabla_i S_{t,w,\alpha}(\xx_t)}{\sqrt{\epsilon+\tilde{\upsilon}_{t+1,i}}} \rangle \Big),
\end{align}
where
\begin{align*}
I_1&=(\frac1{\sqrt{\epsilon+\tilde{\upsilon}_{t+1,i}}}
-
\frac{1}{\sqrt{\epsilon+\upsilon_{t+1,i}}}) \langle \nabla_i S_{t,w,\alpha}(\xx_t), \tilde{\nabla}_iS_{t,w,\alpha}(\xx_t)\rangle.
\end{align*}
Following the lines in the proof of Lemma \ref{app:lemma:descent_lemmaa}, we have
\begin{align}\label{dcc2}
\nonumber  I_1&\leq\sum_{i=1}^{d}\frac{(\nabla_i S_{t,w,\alpha}(\xx_t))^2}{2\sqrt{\epsilon+\tilde{\upsilon}_{t+1,i}}}
+2\sum_{i=1}^{d}\frac{\sqrt{\hat{\mu}_{T,i}}(\tilde{\nabla}_iS_{t,w,\alpha}(\xx_t))^2}{\epsilon+\upsilon_{t+1,i}}\\
&\leq\sum_{i=1}^{d}\frac{(\nabla_i S_{t,w,\alpha}(\xx_t))^2}{2\sqrt{\epsilon+\tilde{\upsilon}_{t+1,i}}}
+ 2\sqrt{\frac{\bar{ \mu}}{W}}
\sum_{i=1}^{d}\frac{(\tilde{\nabla}_iS_{t,w,\alpha}(\xx_t))^2}{\epsilon+\upsilon_{t+1,i}},
\end{align}
where the last inequality follows from observing that
\begin{align}\label{ww}
 \sqrt{\hat{\mu}_{T,i}}\leq \sqrt{\max_{1\leq t \leq T}\max_{1\leq i \leq d}\hat{\mu}_{t,i}}
 &= \sqrt{\frac{\max_{1\leq t \leq T}\max_{1\leq i \leq d}\big(\tilde{\nabla}_i S_{t,w,\alpha}(\xx_t)-\nabla_i S_{t,w,\alpha}(\xx_t)\big)^2}{W}}  \leq
  \sqrt{\frac{\bar{ \mu}}{W}},
\end{align}
due to inequality $\|\cdot \|_{\infty} \leq \| \cdot \|_2$ and Lemma \ref{4543}(b).
Now, we derive an upper bound for the last term in \eqref{882}.
By definition $S_{t,w,\alpha}(\xx_t)$ in \eqref{def}, we obtain
\begin{align}\label{90}
 \nonumber &-\sum_{t=1}^{T}\sum_{i=1}^{d}\langle \nabla_i S_{t,w,\alpha}(\xx_t),\frac{\tilde{\nabla}_i S_{t,w,\alpha}(\xx_t)-\nabla_i S_{t,w,\alpha}(\xx_t)}{\sqrt{\epsilon+\tilde{\upsilon}_{t+1,i}}}\rangle\\
 &=-\frac{1}{W}\sum^{w-1}_{r=0}\alpha^r\sum_{t=1}^{T}\sum_{i=1}^{d}\langle \nabla_i S_{t,w,\alpha}(\xx_t),\frac{\g_{t-r,i}(\xx_{t-r},\xi_{t,t-r}) - \nabla_i\ell_{t-r}(\xx_{t-r})}{\sqrt{\epsilon+\tilde{\upsilon}_{t+1,i}}}\rangle.
\end{align}
Next, let us denote $L_t\triangleq -\langle \nabla S_{t,w,\alpha}(\xx_t),\frac{\g_{t-r}(\xx_{t-r},\xi_{t,t-r}) - \nabla\ell_{t-r}(\xx_{t-r})}{\sqrt{\epsilon+\tilde{\upsilon}_{t+1}}} \rangle$, and $N_t\triangleq \| \frac{\nabla S_{t,w,\alpha}(\xx_t)}{\sqrt{\epsilon+\tilde{\upsilon}_{t+1}}}\|^2\kappa^2$.
For any $1\leq t\leq T$, we have
\begin{align*}
\nonumber \E\left[\exp \left( \frac{L_t^2}{N_t}\right)\right]
&= \E\left[\exp \left( \frac{\| \frac{\nabla S_{t,w,\alpha}(\xx_t)}{\sqrt{\epsilon+\tilde{\upsilon}_{t+1}}}\|^2   \|\g_{t-r}(\xx_{t-r},\xi_{t,t-r}) - \nabla\ell_{t-r}(\xx_{t-r})\|^2 }{N_t}\right)\right]
\\ &=\E\left[\exp \left(\frac{\left\| \g_{t-r}(\xx_{t-r},\xi_{t,t-r}) - \nabla\ell_{t-r}(\xx_{t-r}) \right\|^2}{\kappa^2} \right)\right]\\
&\leq\E\left[\exp \left(\frac{\max_{1\leq t\leq T}\left\| \g_{t-r}(\xx_{t-r},\xi_{t,t-r}) - \nabla\ell_{t-r}(\xx_{t-r}) \right\|^2}{\kappa^2} \right)\right]
\\&\leq  \exp \left( 1 \right),
\end{align*}
where the last line is true due to Assumption \ref{212}.
By invoking Assumption~\ref{ass:sg}(i), it can be obtained that
\begin{align*}
 \E_{t}[L_t] & =-\sum_{i=1}^{d}\langle \nabla_i S_{t,w,\alpha}(\xx_t),\frac{\E_t[\g_{t-r,i}(\xx_{t-r},\xi_{t,t-r}) - \nabla_i\ell_{t-r}(\xx_{t-r})]}{\sqrt{\epsilon+\tilde{\upsilon}_{t+1,i}}} \rangle =0.
\end{align*}
According to Lemma \ref{lemma:sub_gaussian}, with probability at least $1-\delta$, any $\lambda>0$, we have
\begin{align}\label{ohu}
\nonumber  \sum_{t=1}^{T} L_t \leq \frac{3}{4}\lambda \sum_{t=1}^{T} N_t+\frac{1}{\lambda}\ln \frac{1}{\delta}
   &\leq \frac{3}{4}\lambda \sum_{t=1}^{T} \| \frac{\nabla S_{t,w,\alpha}(\xx_t)}{\sqrt{\epsilon+\tilde{\upsilon}_{t+1}}}\|^2 \kappa^2+\frac{1}{\lambda}\ln \frac{1}{\delta}
  \\\nonumber &= \frac{3}{4}\lambda \sum_{t=1}^{T}\sum_{i=1}^{d}  \frac{(\nabla_i S_{t,w,\alpha}(\xx_t))^2}{\sqrt{\epsilon+\tilde{\upsilon}_{t+1,i}}\sqrt{\epsilon+\tilde{\upsilon}_{t+1,i}}} \kappa^2+\frac{1}{\lambda}\ln \frac{1}{\delta}
  \\\nonumber&\leq \frac{3}{4}\lambda \sum_{t=1}^{T}\sum_{i=1}^{d}  \frac{(\nabla_i S_{t,w,\alpha}(\xx_t))^2}{\sqrt{\epsilon+\tilde{\upsilon}_{t+1,i}}\sqrt{\epsilon+\upsilon_{t,i}}} \kappa^2+\frac{1}{\lambda}\ln \frac{1}{\delta}
  \\ &\leq \frac{3}{4}\lambda \sum_{t=1}^{T}\sum_{i=1}^{d}  \frac{(\nabla_i S_{t,w,\alpha}(\xx_t))^2}{\sqrt{\epsilon+\tilde{\upsilon}_{t+1,i}}\sqrt{\epsilon}} \kappa^2+\frac{1}{\lambda}\ln \frac{1}{\delta}.
\end{align}
Replacing \eqref{ohu} in \eqref{90}, we obtain
\begin{align}\label{er}
\nonumber  -\sum_{t=1}^{T}&\langle \nabla_i S_{t,w,\alpha}(\xx_t),\frac{\tilde{\nabla}_i S_{t,w,\alpha}(\xx_t)-\nabla_i S_{t,w,\alpha}(\xx_t)}{\sqrt{\epsilon+\tilde{\upsilon}_{t+1,i}}}\rangle \\
\nonumber &\leq \frac{1}{W}\sum^{w-1}_{r=0}\alpha^r (\frac{3}{4}\lambda \sum_{t=1}^{T}\sum_{i=1}^{d}  \frac{(\nabla_i S_{t,w,\alpha}(\xx_t))^2}{\sqrt{\epsilon+\tilde{\upsilon}_{t+1,i}}\sqrt{\epsilon}} \kappa^2+\frac{1}{\lambda}\ln \frac{1}{\delta})\\
  &\leq \frac{(1-\alpha^w)}{W(1-\alpha)}(\frac{3}{4}\lambda \sum_{t=1}^{T}\sum_{i=1}^{d}  \frac{(\nabla_i S_{t,w,\alpha}(\xx_t))^2}{\sqrt{\epsilon+\tilde{\upsilon}_{t+1,i}}\sqrt{\epsilon}} \kappa^2+\frac{1}{\lambda}\ln \frac{1}{\delta}).
\end{align}
Now, substituting \eqref{dcc2} and \eqref{er} into \eqref{882} and setting $\lambda=\frac{W(1-\alpha)\sqrt{\epsilon}}{3(1-\alpha^w)\kappa^2}$, it can be concluded that

\begin{align*}
\nonumber -&\sum_{t=1}^{T}\sum_{i =1}^d \langle \nabla_i S_{t,w,\alpha}(\xx_t) , \frac{\tilde{\nabla}_iS_{t,w,\alpha}(\xx_t)}{\sqrt{\epsilon+\upsilon_{t+1,i}}} \rangle \leq -\sum_{i=1}^{d}\sum_{t=1}^{T}\frac{(\nabla_i S_{t,w,\alpha}(\xx_t))^2}{\sqrt{\epsilon+\tilde{\upsilon}_{t+1,i}}}
+\sum_{i=1}^{d}\sum_{t=1}^{T}\frac{(\nabla_i S_{t,w,\alpha}(\xx_t))^2}{2\sqrt{\epsilon+\tilde{\upsilon}_{t+1,i}}}
\\&+ 2\sqrt{\frac{\bar{ \mu}}{W}}
\sum_{i=1}^{d}\sum_{t=1}^{T}\frac{(\tilde{\nabla}_iS_{t,w,\alpha}(\xx_t))^2}{\epsilon+\upsilon_{t+1,i}}
+\frac{1}{4} \sum_{t=1}^{T}\sum_{i=1}^{d}  \frac{(\nabla_i S_{t,w,\alpha}(\xx_t))^2}{\sqrt{\epsilon+\tilde{\upsilon}_{t+1,i}}}
+\frac{3(1-\alpha^w)^2\kappa^2}{W^2(1-\alpha)^2\sqrt{\epsilon}}\ln \frac{1}{\delta},
\end{align*}
which completes the proof.
\end{proof}

\subsubsection*{ PROOF OF THEOREM~\ref{thm12}}
\begin{thm*}(\textsc{Adagrad})
Suppose Assumptions~\ref{ass:sg}, \ref{ass:loss} and \ref{212} hold. Let $\textsc{Dts-Ag}$ be the algorithm $\mathcal{A}$ in Algorithm~\ref{algo:mlol} with parameters $\beta_1=0$, $\beta_2=1$, $\eta_{t+1}=\eta$ with $\eta>0$ and $\alpha\rightarrow 1^-$.
Then, for any $\delta \in (0,1)$ and $S_{t,w,\alpha}(\xx_t)$ in \eqref{def}, with probability at least $1-\delta$, the iterates $\xx_t$ satisfy the following bound
%Then, feeding Algorithm~\ref{algo:adagrad} into Algorithm~\ref{algo:mlol} gives the following upper bound of $DLR_w(T)$, with probability $1-\delta$:
\begin{equation*}
\sum^T_{t=1}\|\nabla S_{t,w,\alpha}(\xx_t)\|^2
\le 4C\sqrt{\epsilon}+4C\sqrt{\frac{2T\zeta}{W} }+\frac{48C^2}{W}~.
\end{equation*}
Here,
\begin{align*}
C &\triangleq \varpi_1
+\varpi_2d\ln\left(1 + \frac{2(\zeta+L'^2)T}{d\epsilon}\right)+\frac{3\kappa^2}{\sqrt{\epsilon}}\ln \frac{1}{\delta},\\
\varpi_1 &\triangleq  \frac{4DT}{W\eta},\qquad
\varpi_2 \triangleq \frac{\eta\gamma'}{2}+\frac{2\sqrt{\zeta}}{\sqrt{W}},
\qquad \zeta \triangleq \kappa^2\ln \frac{e}{\delta}.
\end{align*}
\end{thm*}
\begin{proof}
The proof follows along similar lines as Theorem \ref{thm1} with some important differences. We start with the following observation:
\begin{align*}
\frac{S_{t,w,\alpha}(\xx_{t+1})-S_{t,w,\alpha}(\xx_t)}{\eta}&\le \langle \nabla S_{t,w,\alpha}(\xx_t),\xx_{t+1}-\xx_t\rangle+\frac{\gamma'}{2}\|\xx_{t+1}-\xx_t \|^2
\\&=-\sum_{i=1}^{d}\langle \nabla_i S_{t,w,\alpha}(\xx_t),
\frac{\tilde{\nabla}_iS_{t,w,\alpha}(\xx_t)}{\sqrt{\epsilon+\upsilon_{t+1,i}}} \rangle
+
\frac{\eta\gamma'}{2}\sum_{i=1}^{d}\frac{(\tilde{\nabla}_iS_{t,w,\alpha}(\xx_t))^2}{\epsilon+\upsilon_{t+1,i}}~,
\end{align*}
where the first inequality follows from the $\gamma'$-smoothness of the function $S_{t,w,\alpha}$ due to Lemma~\ref{lm:property} and the second step is by the
definition of $\xx_{t+1}$.
Then, summing over $t=1$ to $T$ gives
\begin{align}\label{eq:inner2}
\nonumber \frac{\sum_{t=1}^{T}\big({S_{t,w,\alpha}(\xx_{t+1})-S_{t,w,\alpha}(\xx_t)}\big)}{\eta}
&\le
-\sum_{i=1}^{d}\sum_{t=1}^{T}\langle \nabla_i S_{t,w,\alpha}(\xx_t),
\frac{\tilde{\nabla}_iS_{t,w,\alpha}(\xx_t)}{\sqrt{\epsilon+\upsilon_{t+1,i}}} \rangle
\\&\quad +
\frac{\eta\gamma'}{2}\sum_{i=1}^{d}\sum_{t=1}^{T}\frac{(\tilde{\nabla}_iS_{t,w,\alpha}(\xx_t))^2}{\epsilon+\upsilon_{t+1,i}}~.
\end{align}
From Lemma~\ref{app:lemma:descent_lemmaa2} and \eqref{eq:inner2}, we have
\begin{align}\label{plko2}
\nonumber &\frac{\sum_{t=1}^{T}\big({S_{t,w,\alpha}(\xx_{t+1})-S_{t,w,\alpha}(\xx_t)}\big)}{\eta}
\le
-\sum_{i=1}^{d}\sum_{t=1}^{T}\frac{(\nabla_i S_{t,w,\alpha}(\xx_t))^2}{4\sqrt{\epsilon+\tilde{\upsilon}_{t+1,i}}}
\\&+\left(\frac{\eta\gamma'}{2}+\frac{2\sqrt{\bar{ \mu}}}{\sqrt{W}}\right)
\sum_{i=1}^{d}\sum_{t=1}^{T}\frac{(\tilde{\nabla}_iS_{t,w,\alpha}(\xx_t))^2}{\epsilon+\upsilon_{t+1,i}}
+\frac{3(1-\alpha^w)^2\kappa^2}{W^2(1-\alpha)^2\sqrt{\epsilon}}\ln \frac{1}{\delta}~,
\end{align}
where $\bar{\mu}=\kappa^2\ln \frac{\exp\left(\frac{w\sum^{w-1}_{r=0}\alpha^{2r}}{W^2}\right)}{\delta}$,
 $\tilde{\upsilon}_{t+1,i}\triangleq \frac{1}{W}\big(\upsilon_{t,i} + (\nabla_i S_{t,w,\alpha}(\xx_t))^2 \big)+ \hat{\mu}_{T,i}$, and\\ $\hat{\mu}_{T,i}\triangleq \frac{1}{W}\left(\tilde{\nabla}_i S_{T,w,\alpha}(\xx_T)-\nabla_i S_{T,w,\alpha}(\xx_T)\right)^2$.
We rearrange terms of \eqref{plko2} to obtain
\begin{align}\label{eq:term22}
\nonumber \sum^{T}_{t=1}\sum_{i=1}^{d}&\frac{(\nabla_i S_{t,w,\alpha}(\xx_t))^2}{4\sqrt{\epsilon+\tilde{\upsilon}_{t+1,i}}}\le
\frac{\sum^{T}_{t=1}\big(S_{t,w,\alpha}(\xx_t)-S_{t,w,\alpha}(\xx_{t+1})\big)}{\eta}
\\&+\left(\frac{\eta\gamma'}{2}+\frac{2\sqrt{\bar{ \mu}}}{\sqrt{W}}\right)
\sum_{i=1}^{d}\sum^{T}_{t=1}\frac{(\tilde{\nabla}_iS_{t,w,\alpha}(\xx_t))^2}{\epsilon+\upsilon_{t+1,i}}
+\frac{3(1-\alpha^w)^2\kappa^2}{W^2(1-\alpha)^2\sqrt{\epsilon}}\ln \frac{1}{\delta}~.
\end{align}
Recall that $\upsilon_{t+1,i}=\sum_{j=1}^{t}(\tilde{\nabla}_iS_{j,w,\alpha}(\xx_j))^2$, $\upsilon_{1,i}=0$. Then, with probability at least $1-\delta$, the second term in \eqref{eq:term22} is bounded above by

\begin{align}\label{55}
\nonumber \sum_{i=1}^{d}\sum^{T}_{t=1}&\frac{(\tilde{\nabla}_iS_{t,w,\alpha}(\xx_t))^2}{\epsilon+\upsilon_{t+1,i}}
 =\sum_{i=1}^{d}\sum_{t=1}^{T}\frac{(\tilde{\nabla}_iS_{t,w,\alpha}(\xx_t))^2}{\epsilon+\sum_{j=1}^{t}(\tilde{\nabla}_iS_{j,w,\alpha}(\xx_j))^2}
\stackrel{(i)}\le \sum_{i=1}^{d}
\ln\left(1 +\frac{\sum^{T}_{t=1}(\tilde{\nabla}_iS_{t,w,\alpha}(\xx_t) )^2}{\epsilon} \right)
\\\nonumber&\stackrel{(ii)}\le
d \ln\left(\frac{1}{d}\sum_{i=1}^{d}\big(1 +\frac{\sum^{T}_{t=1}(\tilde{\nabla}_iS_{t,w,\alpha}(\xx_t) )^2}{\epsilon} \big)\right)
=  d\ln\left(1 +\frac{\sum^{T}_{t=1}\|\tilde{\nabla} S_{t,w,\alpha}(\xx_t) \|^2}{d\epsilon} \right)\\
\nonumber&\stackrel{(iii)}\leq
d\ln\left(1 +\frac{\sum^{T}_{t=1}(2\|\tilde{\nabla} S_{t,w,\alpha}(\xx_t)-\nabla S_{t,w,\alpha}(\xx_t) \|^2+2\|\nabla S_{t,w,\alpha}(\xx_t) \|^2)}{d \epsilon} \right)
\\& \stackrel{(iv)}\le d\ln\left(1 +\frac{2(\bar{\mu}+L'^2)T}{d\epsilon} \right)~,
\end{align}
where (i) is obtained by applying Lemma~\ref{lemma:sum_ratio} with $\beta_2=1$, (ii) is due to the convex inequality
$\frac{1}{d}\sum_{i=1}^{d}\ln (a_i)\leq \ln (\frac{1}{d}\sum_{i=1}^{d} a_i)$, (iii) is by $\| a+b\|^2\leq 2\|a \|^2+2\|b\|^2$, and (iv) follows from Lemma \ref{4543}(b) and Lemma \ref{lm:property}.

Therefore, plugging \eqref{55} into \eqref{eq:term22}, we get
\begin{align*}
\nonumber I_1&\triangleq\sum^{T}_{t=1}\sum_{i=1}^{d}\frac{(\nabla_i S_{t,w,\alpha}(\xx_t))^2}{4\sqrt{\epsilon+\tilde{\upsilon}_{t+1,i}}}
\le
\frac{\sum^{T}_{t=1}\big(S_{t,w,\alpha}(\xx_t)-S_{t,w,\alpha}(\xx_{t+1})\big)}{\eta}
\\\nonumber &\quad +
\left(\frac{\eta\gamma'}{2}+\frac{2\sqrt{\bar{ \mu}}}{\sqrt{W}}\right)d\ln\left(1 + \frac{2(\bar{\mu}+L'^2)T}{d\epsilon}\right)+\frac{3(1-\alpha^w)^2\kappa^2}{W^2(1-\alpha)^2\sqrt{\epsilon}}\ln \frac{1}{\delta}
\\\nonumber&=
\frac{\sum^{T}_{t=1}\big(S_{t,w,\alpha}(\xx_t)-S_{t+1,w,\alpha}(\xx_{t+1})\big)}{\eta}
+\frac{\sum^{T}_{t=1}\big(S_{t+1,w,\alpha}(\xx_{t+1})-S_{t,w,\alpha}(\xx_{t+1})\big)}{\eta}
\\&\quad+\left(\frac{\eta\gamma'}{2}+\frac{2\sqrt{\bar{ \mu}}}{\sqrt{W}}\right)d\ln\left(1 + \frac{2(\bar{\mu}+L'^2)T}{d\epsilon}\right)+\frac{3(1-\alpha^w)^2\kappa^2}{W^2(1-\alpha)^2\sqrt{\epsilon}}\ln \frac{1}{\delta}.
\end{align*}
Using Lemmas~\ref{lm:sumobj22} and \ref{lm:sumobj2}, we get
\begin{align}\label{C2}
\nonumber I_1&\leq \frac{2D(1-\alpha^w)T}{W(1-\alpha)\eta}+\frac{D(1+\alpha^{w-1})T}{W\eta}
+\frac{D(1-\alpha^{w-1})(1+\alpha)T}{W(1-\alpha)\eta}\\
&\quad+\left(\frac{\eta\gamma'}{2}+\frac{2\sqrt{\bar{ \mu}}}{\sqrt{W}}\right)
d\ln\left( 1+ \frac{2(\bar{\mu}+L'^2)T}{d\epsilon}\right)+\frac{3(1-\alpha^w)^2\kappa^2}{W^2(1-\alpha)^2\sqrt{\epsilon}}\ln \frac{1}{\delta}\triangleq C~.
\end{align}
What's more, recalling that $\up_{t+1}= \up_{t}+(\tilde{\nabla}S_{t,w,\alpha}(\xx_t))^2=\sum_{l=1}^{t}(\tilde{\nabla}S_{l,w,\alpha}(\xx_l))^2$, $\up_1=0$ and using the bound $\|a+b\|^2\leq 2\|a \|^2+2\| b\|^2$ yields that with probability at least $1-\delta$,

\begin{align}\label{3322}
\nonumber & \sum_{i=1}^{d}\upsilon_{T,i} +\sum_{i=1}^{d} (\nabla_i S_{T,w,\alpha}(\xx_{T}))^2 +\sum_{i=1}^{d}  \left(\tilde{\nabla}_i S_{T,w,\alpha}(\xx_T)-\nabla_i S_{T,w,\alpha}(\xx_T)\right)^2
\\\nonumber &=\sum_{t=1}^{T-1}\sum_{i=1}^{d}(\tilde{\nabla}_i S_{t,w,\alpha}(\xx_t))^2
+\sum_{i=1}^{d}(\nabla_i S_{T,w,\alpha}(\xx_{T}))^2+\sum_{i=1}^{d} \left(\tilde{\nabla}_i S_{T,w,\alpha}(\xx_T)-\nabla_i S_{T,w,\alpha}(\xx_T)\right)^2
\\\nonumber&\leq 2\sum_{t=1}^{T-1}\sum_{i=1}^{d}(\tilde{\nabla}_i S_{t,w,\alpha}(\xx_t)-{\nabla}_i S_{t,w,\alpha}(\xx_t))^2+2\sum_{t=1}^{T-1}\sum_{i=1}^{d}(\nabla_i S_{t,w,\alpha}(\xx_t))^2
\\\nonumber&\qquad+2\sum_{i=1}^{d}(\nabla_i S_{T,w,\alpha}(\xx_{T}))^2+2\sum_{i=1}^{d}\left(\tilde{\nabla}_i S_{T,w,\alpha}(\xx_T)-\nabla_i S_{T,w,\alpha}(\xx_T)\right)^2
\\ &\leq 2\sum_{t=1}^{T}\sum_{i=1}^{d}(\tilde{\nabla}_i S_{t,w,\alpha}(\xx_t)-{\nabla}_i S_{t,w,\alpha}(\xx_t))^2+2\sum_{t=1}^{T}\sum_{i=1}^{d}(\nabla_i S_{t,w,\alpha}(\xx_t))^2
\le\ 2T\bar{\mu}+ 2Z,
\end{align}
where $Z\triangleq\sum^{T}_{t=1}\|\nabla S_{t,w,\alpha}(\xx_t)\|^2$, $\bar{\mu}\triangleq \kappa^2\ln \frac{\exp\left(\frac{w\sum^{w-1}_{r=0}\alpha^{2r}}{W^2}\right)}{\delta}$, and the last inequality follows from Lemma \ref{4543}(b). Based on these results, with probability at least $1-\delta$, we have
\begin{align}\label{dd2}
\nonumber & \sum^{T}_{t=1}\sum_{i=1}^{d}\frac{(\nabla_i S_{t,w,\alpha}(\xx_t))^2}{4\sqrt{\epsilon+\tilde{\upsilon}_{t+1,i}}}
=\sum^{T}_{t=1}\sum_{i=1}^{d}\frac{(\nabla_i S_{t,w,\alpha}(\xx_t))^2}{4\sqrt{\epsilon+\frac{1}{W}\big(\upsilon_{t,i} + (\nabla_i S_{t,w,\alpha}(\xx_t))^2\big) + \hat{\mu}_{T,i}}}
\\\nonumber&\ge
\sum^{T}_{t=1}\sum_{i=1}^{d}\frac{(\nabla_i S_{t,w,\alpha}(\xx_t))^2}{4\sqrt{\epsilon+\frac{1}{W}\big(\upsilon_{T,i}
+ (\nabla_i S_{T,w,\alpha}(\xx_T))^2 + \sum_{t=1}^{T}(\nabla_i S_{t,w,\alpha}(\xx_t))^2 \big)+ \hat{\mu}_{T,i}}}
\\\nonumber&\ge
\sum^{T}_{t=1}\sum_{i=1}^{d}\frac{(\nabla_i S_{t,w,\alpha}(\xx_t))^2}{4\sqrt{\epsilon+\frac{1}{W}\big(\sum_{i=1}^{d}\upsilon_{T,i} + \sum_{i=1}^{d}(\nabla_i S_{T,w,\alpha}(\xx_T))^2 + \sum_{t=1}^{T}\sum_{i=1}^{d}(\nabla_i S_{t,w,\alpha}(\xx_t))^2\big) +\sum_{i=1}^{d}\hat{\mu}_{T,i}}}
\\&\ge
\frac{\sum^{T}_{t=1}\|\nabla S_{t,w,\alpha}(\xx_t)\|^2}{4\sqrt{\epsilon+\frac{1}{W}\big( 3Z + 2T\bar{\mu}\big)}}~,
\end{align}
where in the last inequality we used \eqref{3322}.
From \eqref{dd2} and \eqref{C2}, with probability at least $1-\delta$, we obtain
\begin{equation*}
\frac{Z}{4\sqrt{\epsilon+\frac{1}{W}\big( 3Z +2T\bar{\mu}\big)}}
\le C~.
\end{equation*}
Next, by applying Lemma \ref{lem:quadratic} to the above quadratic inequality, we get that
\begin{equation}\label{mmm2}
\sum^T_{t=1}\|\nabla S_{t,w,\alpha}(\xx_t)\|^2
\le 4C\sqrt{\epsilon}+4C\sqrt{\frac{2T\bar{\mu}}{W} }+\frac{48C^2}{W}~.
\end{equation}
Here, $C \triangleq \nu
+u d\ln\left(1 + \frac{2(\bar{\mu}+L'^2)T}{d\epsilon}\right)+y$, where
\begin{align*}
\nu &\triangleq  \frac{2D(1-\alpha^w)T}{W(1-\alpha)\eta}+\frac{D(1+\alpha^{w-1})T}{W\eta}+\frac{D(1-\alpha^{w-1})(1+\alpha)T}{W(1-\alpha)\eta},
\\ u &\triangleq \left(\frac{\eta\gamma'}{2}+\frac{2\sqrt{\bar{ \mu}}}{\sqrt{W}}\right),\qquad
\bar{\mu} \triangleq \kappa^2\ln \frac{\exp\left(\frac{w\sum^{w-1}_{r=0}\alpha^{2r}}{W^2}\right)}{\delta},\qquad y\triangleq\frac{3(1-\alpha^w)^2\kappa^2}{W^2(1-\alpha)^2\sqrt{\epsilon}}\ln \frac{1}{\delta}.
\end{align*}
Following the same argument as in \eqref{frd1}, we also obtain
$
 \nu \leq \frac{4D T}{W\eta}\Big(\frac{2-\alpha^{w}+\alpha^{w-1}}{1-\alpha}\Big).
$
Thus, as $\alpha \rightarrow 1^-$, we have
\begin{equation}\label{ffnf}
  \nu \leq \frac{4D T}{W\eta}\triangleq \varpi_1,\quad
  \bar{\mu} \stackrel{\alpha \rightarrow 1^-}=\kappa^2\ln \frac{e}{\delta}\triangleq \zeta,\quad
  u\stackrel{\alpha \rightarrow 1^-}=\frac{\eta\gamma'}{2}+\frac{2\sqrt{\zeta}}{\sqrt{W}}\triangleq\varpi_2,\quad
   y\stackrel{\alpha \rightarrow 1^-}=\frac{3\kappa^2}{\sqrt{\epsilon}}\ln \frac{1}{\delta}.
\end{equation}
Plugging \eqref{ffnf} into \eqref{mmm2}, we get the stated bound.
\end{proof}

\subsubsection*{ PROOF OF Corollary~\ref{corrh}}
\begin{cor*}
Under the same conditions stated in Theorem \ref{thm12}, using $w\in\Theta(T)$ and $\alpha\rightarrow 1^{-}$ yields a regret bound of order
\begin{equation*}
\sum^T_{t=1} \|\nabla S_{t,w,\alpha}(\xx_t)\|^2\leq \mathcal{O}(\ln T).
\end{equation*}
\end{cor*}
\begin{proof}
By Theorem \ref{thm12}, we have
\begin{equation*}
\sum^T_{t=1}\|\nabla S_{t,w,\alpha}(\xx_t)\|^2
\le 4C\sqrt{\epsilon}+4C\sqrt{\frac{2T\zeta}{W} }+\frac{48C^2}{W}=I_1+I_2+I_3~,
\end{equation*}
where $C \triangleq \varpi_1
+\varpi_2d\ln\left(1 + \frac{2(\zeta+L'^2)T}{d\epsilon}\right)+\frac{3\kappa^2}{\sqrt{\epsilon+\upsilon_{1,i}}}\ln \frac{1}{\delta}$, $\varpi_1 \triangleq  \frac{4DT}{W\eta}$, $\varpi_2 \triangleq \frac{\eta\gamma'}{2}+\frac{2\sqrt{\zeta}}{\sqrt{W}}$, and $\zeta \triangleq \kappa^2\ln \frac{e}{\delta}$.\\
Recall that $W\triangleq \sum_{r=0}^{w-1}\alpha^r$. As $\alpha\rightarrow 1^{-}$, we get the following equalities:
\begin{align*}
C&=\frac{4DT}{W\eta}
+\left(\frac{\eta\gamma'}{2}+\frac{2\sqrt{\kappa^2\ln \frac{e}{\delta}}}{\sqrt{W}}\right)d\ln\left(1 + \frac{2\big(\kappa^2\ln \frac{e}{\delta}+L'^2\big)T}{d\epsilon}\right)+\frac{3\kappa^2}{\sqrt{\epsilon+\upsilon_{1,i}}}\ln \frac{1}{\delta}=\mathcal{O}(\ln T),\\
  I_1&=4C\sqrt{\epsilon} =\mathcal{O}(\ln T),\quad I_2=4C\sqrt{\frac{2T \kappa^2\ln \frac{e}{\delta}}{W} }=\mathcal{O}(\ln T), \quad
  I_3=\frac{48C^2}{W}=\mathcal{O}(\ln T).
\end{align*}
Combine the above results we can easily have the desired result.

\end{proof}
%%%%%%%%ADAM
\subsubsection*{ PROOF OF LEMMA~\ref{app:lemma:descent_lemma2}}
\begin{lem*}
Suppose Assumptions~\ref{ass:sg}, \ref{ass} and \ref{212} hold.
Let $m_{t+1,i}$ and $\upsilon_{t+1,i}$ be the sequences defined in the Algorithm $\textsc{Dts-Ag}$.
Then, for any $\delta \in (0,1)$, $0<\alpha<1$, $w$, $\epsilon>0$, $0\leq \beta_1<\beta_2\leq 1$, $1\leq k\leq t$, and $S_{t,w,\alpha}(\xx_t)$ in \eqref{def}, with probability at least $1-\delta$, we have
\begin{align*}
\nonumber
 &   \sum_{i =1}^d \sum_{t=1}^{T} \langle \nabla_i S_{t,w,\alpha}(\xx_t) , \frac{m_{t+1,i}}{\sqrt{\epsilon+\upsilon_{t+1,i}}} \rangle \geq
        \sum_{i=1}^d \sum_{t=1}^{T} \sum_{k=0}^{t-1}\beta_1^k\frac{(\nabla_i S_{t+1-k,w,\alpha}(\xx_{t+1-k}) )^2}{4\sqrt{\epsilon+\tilde{\upsilon}_{t+1,i}}}
      \\\nonumber
 &\qquad  -\sum_{t=1}^{T}
     \sum_{k=0}^{t-1} \frac{1}{\sqrt{1-\beta_1}}
     \bbratio^k \left(\sqrt{k+1}+\frac{2\sqrt{\bar \mu} }{\sqrt{W}}\right) \|A_{t+1 -k}\|^2
     \\\nonumber&\qquad-\sum_{t=1}^{T}\frac{\eta_{t+1}^2 \gamma'^2}{4}\sqrt{1 - \beta_1} \sum_{l=1}^{t}\norm{B_{t+1-l}}^2\sum_{k=l}^{t }\beta_1^k \sqrt{k}-\sum_{t=1}^{T}\sum_{k=0}^{t-1}\beta_1^k\frac{\sqrt{1-\beta_1}\vartheta_t}{2\sqrt{k+1}}
     \\&\qquad-\frac{3(1-\alpha^w)^2\kappa^2}{W^2\beta_1^k(1-\alpha)^2\sqrt{\epsilon}}\ln \frac{1}{\delta}.
\end{align*}
Here, $\bar{\mu}=\kappa^2\ln \frac{\exp\left(\frac{w\sum^{w-1}_{r=0}\alpha^{2r}}{W^2}\right)}{\delta}$,
$\tilde{\upsilon}_{t+1,i}\triangleq \frac{1}{W}\big(\beta_2\upsilon_{t,i} + (\nabla_i S_{t,w,\alpha}(\xx_t))^2 \big)+ \hat{\mu}_{T,i}$,\\
 $\hat{\mu}_{T,i}\triangleq \frac{1}{W}\left(\tilde{\nabla}_i S_{T,w,\alpha}(\xx_T)-\nabla_i S_{T,w,\alpha}(\xx_T)\right)^2$ for all $i\in [d]$, and
\begin{align}\label{cvbb2}
\vartheta_t \triangleq \frac{8L'^2 }{W^2} +\frac{2(1-\alpha^{w-2})\gamma'^2}{W^2(1-\alpha)} \sum_{r=1}^{w-1}  \alpha^{r-1} \|\eta_{t-r+2-k}B_{t-r+2-k} \|^2,\quad
B_{t}\triangleq \frac{\mm_{t}}{\sqrt{\epsilon+\up_{t}}},\quad
A_{t}\triangleq \frac{\g_{t}}{\sqrt{\epsilon+\up_{t}}}.
\end{align}
\end{lem*}
\begin{proof}
First, we have
\begin{align} \label{app:eq:descent_big2}
\nonumber  & -\sum_{i=1}^d \sum_{t=1}^{T}\langle \nabla_i S_{t,w,\alpha}(\xx_t) , \frac{m_{t+1,i}}{\sqrt{\epsilon+\upsilon_{t+1,i}}} \rangle
\\&=
       - \sum_{i=1}^d \sum_{t=1}^{T}\sum_{k=0}^{t-1} \beta_1^k \langle \nabla_i S_{t,w,\alpha}(\xx_t) , \frac{\tilde{\nabla}_i S_{t+1-k,w,\alpha}(\xx_{t+1-k})}{\sqrt{\epsilon+\upsilon_{t+1,i}}} \rangle
      =I_1+ I_2,
     \end{align}
     where
     \begin{align*}
     I_1=&-\sum_{i=1}^d \sum_{t=1}^{T}\sum_{k=0}^{t-1}\beta_1^k \langle \nabla_i S_{t+1-k,w,\alpha}(\xx_{t+1-k}) , \frac{\tilde{\nabla}_i S_{t+1-k,w,\alpha}(\xx_{t+1-k})}{\sqrt{\epsilon+\upsilon_{t+1,i}}} \rangle,\\
       I_2=& \sum_{i=1}^d \sum_{t=1}^{T}\sum_{k=0}^{t-1}  \beta_1^k \langle (\nabla_i S_{t+1-k,w,\alpha}(\xx_{t+1-k})-\nabla_i S_{t,w,\alpha}(\xx_t)) , \frac{\tilde{\nabla}_i S_{t+1-k,w,\alpha}(\xx_{t+1-k})}{\sqrt{\epsilon+\upsilon_{t+1,i}}} \rangle.
     \end{align*}
Next, we proceed to upper bound $I_1$ and $I_2$ terms.
For convenience, we denote
\begin{align*}
g_{t+1-k,i}&\triangleq \tilde{\nabla}_i S_{t+1-k,w,\alpha}(\xx_{t+1-k}),\,\, \text{for all}\,\, i\in [d].
\end{align*}
\begin{itemize}

\item Bound for $I_2$.

By the same argument as in \eqref{app:eq:B_bound_second}, we get the following inequality:

\begin{align}
\nonumber
    \abs{I_2} &\leq
                \frac{\eta_{t+1}^2 \gamma'^2}{4 }\sqrt{1 - \beta_1} \sum_{l=1}^{t}\norm{B_{t+1-l}}^2\sum_{k=l}^{t}\beta_1^k \sqrt{k}
                + \sum_{k=0}^{t-1} \beta_1^k
        \frac{\sqrt{1 - \beta_1}\vartheta_t}{2  \sqrt{k+1}}
       \\& \quad+\frac{1}{\sqrt{1 - \beta_1}} \sum_{k=0}^{t-1} \bbratio^k \sqrt{k+1}\norm{A_{t+1-k}}^2
    \label{app:eq:B_bound_second2},
\end{align}
where $A_{t+1-k}$ has been defined in \eqref{cvbb2}.
 \item Bound for $I_1$.\\
Recalling $\tilde{\upsilon}_{t+1,i}$ for all $i\in [d]$ we have that
\begin{align}
     I_1 = I_{11} + I_{12},
    \label{app:eq:cool_plus_bad2}
\end{align}
where
\begin{align*}
I_{11}&=-\sum_{t=1}^{T}\sum_{i=1}^{d}\sum_{k=0}^{t-1}\frac{1}{\sqrt{\epsilon+\tilde{\upsilon}_{t+1,i}}}\langle \nabla_i S_{t+1-k,w,\alpha}(\xx_{t+1-k}) , g_{t+1-k,i}\rangle,\\
I_{12}&=\sum_{t=1}^{T}\sum_{i=1}^{d}\sum_{k=0}^{t-1}(\frac{1}{\sqrt{\epsilon+\tilde{\upsilon}_{t+1,i}}}-\frac{1}{\sqrt{\epsilon+\upsilon_{t+1,i}}})\langle \nabla_i S_{t+1-k,w,\alpha}(\xx_{t+1-k}) , g_{t+1-k,i} \rangle.
\end{align*}
First, we give an upper bound on $I_{11}$ term in \eqref{app:eq:cool_plus_bad2}.
Observe that
\begin{align}\label{ggf2}
\nonumber I_{11}&= -\sum_{t=1}^{T}\sum_{i=1}^{d}\sum_{k=0}^{t-1}\langle \nabla_i S_{t+1-k,w,\alpha}(\xx_{t+1-k}), \frac{\tilde{\nabla}_i S_{t+1-k,w,\alpha}(\xx_{t+1-k})\rangle - \nabla_i S_{t+1-k,w,\alpha}(\xx_{t+1-k})}{\sqrt{\epsilon+\tilde{\upsilon}_{t+1,i}}}\rangle\\\nonumber
&\qquad-\sum_{t=1}^{T}\sum_{i=1}^{d}\sum_{k=0}^{t-1}\frac{1}{\sqrt{\epsilon+\tilde{\upsilon}_{t+1,i}}}\langle \nabla_i S_{t+1-k,w,\alpha}(\xx_{t+1-k}) , \nabla_i S_{t+1-k,w,\alpha}(\xx_{t+1-k})\rangle
\\\nonumber &\leq
\frac{(1-\alpha^w)}{W(1-\alpha)}(\frac{3}{4}\lambda \sum_{t=1}^{T}\sum_{i=1}^{d}\sum_{k=0}^{t-1}
\frac{(\nabla_i S_{t+1-k,w,\alpha}(\xx_{t+1-k}))^2}{\sqrt{\epsilon+\tilde{\upsilon}_{t+1,i}}\sqrt{\epsilon}} \kappa^2+\frac{1}{\lambda}\ln \frac{1}{\delta})
\\&\qquad -\sum_{t=1}^{T}\frac{(\nabla_i S_{t+1-k,w,\alpha}(\xx_{t+1-k}) )^2}{\sqrt{\epsilon+\tilde{\upsilon}_{t+1,i}}},
%\leq \frac{\|\nabla_i S_{t+1-k,w,\alpha}(x_{t-k}) \|^2}{\sqrt{\epsilon+\tilde{\upsilon}_{t+1-k,i}}}~,
\end{align}
where the last inequality follows from the same argument for \eqref{er}.

We next bound the $I_{12}$ in \eqref{app:eq:cool_plus_bad2}. Similar to the arguments for \eqref{dss}, we have
\begin{align}\label{2m}
\nonumber | I_{12}|
& \leq \sum_{t=1}^{T}\sum_{i=1}^{d}\sum_{k=0}^{t-1} \frac{(\nabla_i S_{t+1-k,w,\alpha}(\xx_{t+1-k}))^2}{2\sqrt{\epsilon+\tilde{\upsilon}_{t+1,i}}}
+\sum_{t=1}^{T}\sum_{i=1}^{d}\sum_{k=0}^{t-1}
\frac{2\sqrt{\hat{\mu}_{T,i}}}{\sqrt{1-\beta_1}\beta_2^k}\|A_{t+1-k}\|^2\\
& \leq \sum_{t=1}^{T}\sum_{i=1}^{d}\sum_{k=0}^{t-1} \frac{(\nabla_i S_{t+1-k,w,\alpha}(\xx_{t+1-k}))^2}{2\sqrt{\epsilon+\tilde{\upsilon}_{t+1,i}}}
+\frac{2\sqrt{\bar \mu} }{\sqrt{W}\sqrt{1-\beta_1}}\sum_{t=1}^{T}\sum_{i=1}^{d}\sum_{k=0}^{t-1}
\frac{1}{\beta_2^k}\|A_{t+1-k}\|^2.
\end{align}
where the last inequality follows from \eqref{ww}.

Therefore, substituting \eqref{ggf2} and \eqref{2m} into \eqref{app:eq:cool_plus_bad2} and setting $\lambda=\frac{W(1-\alpha)\sqrt{\epsilon}\beta_1^k}{3(1-\alpha^w)\kappa^2}$, we then obtain
\begin{align}\label{app:eq:descent_proof:almost_there2}
\nonumber  \sum_{t=1}^{T} I_1 &\leq - \sum_{t=1}^{T}\sum_{k=0}^{t-1}\sum_{i=1}^d \beta_1^k\frac{(\nabla_i S_{t+1-k,w,\alpha}(\xx_{t+1-k}) )^2}{2\sqrt{\epsilon+\tilde{\upsilon}_{t+1,i}}}
\\\nonumber &\qquad +\frac{2\sqrt{\bar \mu} }{\sqrt{W}\sqrt{1-\beta_1}}\sum_{t=1}^{T}\sum_{k=0}^{t-1}
     \bbratio^k  \|A_{t+1 -k}\|^2\\&\qquad
     +\frac{1}{4} \sum_{t=1}^{T}\sum_{i=1}^{d} \sum_{k=0}^{t-1} \beta_1^k \frac{(\nabla_i S_{t+1-k,w,\alpha}(\xx_{t+1-k}))^2}{\sqrt{\epsilon+\tilde{\upsilon}_{t+1,i}}}
     +\frac{3(1-\alpha^w)^2\kappa^2}{W^2\beta_1^k(1-\alpha)^2\sqrt{\epsilon}}\ln \frac{1}{\delta}.
\end{align}

\end{itemize}
Finally, injecting \eqref{app:eq:descent_proof:almost_there2} and \eqref{app:eq:B_bound_second2} into \eqref{app:eq:descent_big2} gives the desired result.
\end{proof}
\subsubsection*{ PROOF OF THEOREM~\ref{adam2}}
\begin{thm*}(\textsc{Adam})
Suppose Assumptions~\ref{ass:sg}, \ref{ass} and \ref{212} hold. Let $\textsc{Dts-Ag}$ be the algorithm $\mathcal{A}$ in Algorithm~\ref{algo:mlol} with parameters $\eta_{t+1} = \eta (1 - \beta_1) \sqrt{\sum_{j=0}^{t} \beta_2^j}$ with $0<\beta_2<1$, $\eta>0$, $0<\beta_1<\beta_2$, and $\alpha \rightarrow 1^-$.
Furthermore, let $\sqrt{\sum_{r=0}^{t} \beta_2^r}\geq \frac{\varsigma}{\sqrt{1-\beta_2}}$ for some $\varsigma>0$ and $t\in [T]$.
Then, for any $\delta \in (0,1)$ and $S_{t,w,\alpha}(\xx_t)$ in \eqref{def}, with probability at least $1-\delta$,  the iterates $\xx_t$ satisfy the following bound
\begin{align*}
\sum^T_{t=1}\|\nabla S_{t,w,\alpha}(\xx_t)\|^2
&\le
\frac{4\sqrt{1-\beta_2}C}{\varsigma\eta(1-\beta_1)}\Big(\sqrt{\epsilon}+\sqrt{\frac{2\zeta}{W}}\Big)+\frac{48(1-\beta_2)C^2}{W\varsigma^2\eta^2(1-\beta_1)^2}.
\end{align*}
Here, $$C \triangleq \varpi_1 + \varpi_2 \left(d\ln\big(1 +\frac{2(\zeta+L'^2)}{d\epsilon(1-\beta_2)} \big) - T \ln(\beta_2)\right)+\varpi_3,$$
where
\begin{align*}
\varpi_1&\triangleq\frac{4D T}{W}+\frac{8T\eta (1-\beta_1)L'^2}{\beta_1\sqrt{(1-\beta_2)}W^2},\qquad \qquad \zeta \triangleq \kappa^2\ln \frac{e}{\delta},\\
    \varpi_2 &\triangleq \frac{ d \eta^2(1- \beta_1)\gamma'}{2(1 - \beta_2)(1 - \beta_1 / \beta_2)} +
     \frac{d \eta^3  \gamma'^2 \beta_1}{(1 - \beta_1 / \beta_2) (1 - \beta_2)^{3/2}}
     \\&\qquad +
     \frac{ 2d \eta (1+\frac{\sqrt{\zeta}}{\sqrt{W}}) \sqrt{1-\beta_1}}{(1 - \beta_1 / \beta_2)^{3/2}\sqrt{1 - \beta_2}}
     +\frac{2\eta^3(1-\beta_1)^2\gamma'^2}{\beta_1 (1-\beta_2)^{3/2}(1 - \beta_1 / \beta_2)},\\
     \varpi_3 &\triangleq \frac{3\eta(1-\beta_1)\kappa^2}{W^2\beta_1^T\sqrt{1-\beta_2}\sqrt{\epsilon}}\ln \frac{1}{\delta}.
\end{align*}
\end{thm*}

\begin{proof}
The proof follows along similar lines as Theorem \ref{adam} with some important differences.
By the $\gamma'$-smoothness of $\ell_t$ functions, $S_t$ is $\gamma'$-smooth as well. Hence, we have
\begin{align*}
    S_{t,w,\alpha}(\xx_{t+1})-S_{t,w,\alpha}(\xx_{t})  &\leq  - \eta_{t+1} \langle \nabla S_{t,w,\alpha}(\xx_t) , \frac{\mm_{t+1}}{\sqrt{\epsilon+\up_{t+1}}} \rangle
 + \frac{\eta_{t+1}^2 \gamma'}{2}\norm{\frac{\mm_{t+1}}{\sqrt{\epsilon+\up_{t+1}}}}^2.
\end{align*}
Then, Summing over $t = 1$ to $T$ and using Lemma~\ref{app:lemma:descent_lemma2} gives,
\begin{align}
\nonumber
  &  \sum_{t=1}^{T}\big(S_{t,w,\alpha}(\xx_{t+1})-S_{t,w,\alpha}(\xx_{t})\big)  \leq
                  -
                  \sum_{i=1}^d \sum_{t=1}^{T}\sum_{k=0}^{t-1}\eta_{t+1}\beta_1^k\frac{(\nabla_i S_{t+1-k,w,\alpha}(\xx_{t+1-k}) )^2}{4\sqrt{\epsilon+\tilde{\upsilon}_{t+1,i}}}
 \\\nonumber
   & \quad  +
   \sum_{t=1}^{T}\sum_{k=0}^{t-1}  \frac{ \eta_{t+1}}{\sqrt{1-\beta_1}}
     \bbratio^k \left(\sqrt{k+1}+\frac{2\sqrt{\bar \mu} }{\sqrt{W}}\right) \norm{A_{t+1 -k}}^2
     \\\nonumber&\quad+\sum_{t=1}^{T}\frac{\eta_{t+1}^3 \gamma'^2}{4}\sqrt{1 - \beta_1} \sum_{l=1}^{t}\norm{B_{t+1-l}}^2\sum_{k=l}^{t }\beta_1^k \sqrt{k}
     + \sum_{t=1}^{T}\sum_{k=0}^{t-1}\eta_{t+1}\beta_1^k\frac{\sqrt{1-\beta_1}\vartheta_t}{2\sqrt{k+1}}
 \\&\quad  +\frac{3(1-\alpha^w)^2\kappa^2\eta_{t+1}}{\beta_1^kW^2(1-\alpha)^2\sqrt{\epsilon}}\ln \frac{1}{\delta}+\sum_{t=1}^{T} \frac{\eta_{t+1}^2 \gamma'}{2}\norm{B_{t+1}}^2,
 \label{app:eq:common_first2}
\end{align}
where $A_t$, $B_t$, and $\vartheta_t$ are defined as in \eqref{cvbb2}, $\bar{\mu}=\kappa^2\ln \frac{\exp\left(\frac{w\sum^{w-1}_{r=0}\alpha^{2r}}{W^2}\right)}{\delta}$, and $\tilde{\upsilon}_{t+1,i}\triangleq \frac{1}{W}\big(\beta_2\upsilon_{t,i} + (\nabla_i S_{t,w,\alpha}(\xx_t))^2 \big)+ \hat{\mu}_{T,i}$, $\hat{\mu}_{T,i}\triangleq \frac{1}{W}\left(\tilde{\nabla}_i S_{T,w,\alpha}(\xx_T)-\nabla_i S_{T,w,\alpha}(\xx_T)\right)^2$.
Hence, rearranging the above inequality, and using the fact that $\eta_{t+1}$ is non-decreasing, we obtain:
\begin{align}\label{kk2}
\nonumber &\sum_{t=1}^{T}
                  \sum_{i=1}^d \sum_{k=0}^{t-1}\eta_{t+1}\beta_1^k\frac{(\nabla_i S_{t+1-k,w,\alpha}(\xx_{t+1-k}) )^2}{4\sqrt{\epsilon+\tilde{\upsilon}_{t+1,i}}}
                  \leq  \underbrace{\sum_{t=1}^{T} \big(S_{t,w,\alpha}(\xx_{t})-S_{t,w,\alpha}(\xx_{t+1})\big)}_{I_1}\\\nonumber
                  &+ \underbrace{\frac{ \eta_{T+1}}{\sqrt{1-\beta_1}}\sum_{t=1}^T \sum_{k=0}^{t-1}
     \bbratio^k \left(\sqrt{k+1}+\frac{2\sqrt{\bar \mu} }{\sqrt{W}}\right) \norm{A_{t+1 -k}}^2}_{I_2}
     + \underbrace{\frac{\eta_{T+1}^2 \gamma'}{2}\sum_{t=1}^{T}\norm{B_{t+1}}^2}_{I_3}\\\nonumber &+ \underbrace{\eta_{T+1}\sum_{t=1}^{T}\sum_{k=0}^{t-1}\beta_1^k\frac{\sqrt{1-\beta_1}\vartheta_t}{2\sqrt{k+1}}}_{I_4}
                 +\underbrace{\frac{\eta_{T+1}^3 \gamma'^2}{4}\sqrt{1 - \beta_1} \sum_{t=1}^{T} \sum_{l=1}^{t}\norm{B_{t+1-l}}^2\sum_{k=l}^{t }\beta_1^k \sqrt{k} }_{I_5}\\
                 &+\eta_{T+1}\frac{3(1-\alpha^w)^2\kappa^2}{\beta_1^kW^2(1-\alpha)^2\sqrt{\epsilon}}\ln \frac{1}{\delta}.
\end{align}
Along the same lines of proof of Theorem \ref{adam},
i.e. from \eqref{eq:cons}-\eqref{app:eq:common_c_bound} we get that
\begin{subequations}\label{bbbv2}
\begin{align}
 I_1&\leq \frac{2D(1-\alpha^w)T}{W(1-\alpha)}+\frac{D(1+\alpha^{w-1})T}{W}
+\frac{D(1-\alpha^{w-1})(1+\alpha)T}{W(1-\alpha)}
\label{eq:cons2}~,
\end{align}

\begin{align}\label{app:eq:common_d_bound2}
    I_2 & \leq  \frac{2 \eta_{T+1}   (1+\frac{\sqrt{\bar \mu} }{\sqrt{W}}) }{\sqrt{1-\beta_1}(1 - \beta_1 / \beta_2)^{3/2}}\sum_{i=1}^d \Big( \ln\big(1 + \frac{\sum^{T}_{t=1}\beta_2^{T-t}(\tilde{\nabla}_i S_{t,w,\alpha}(\xx_t) )^2}{\epsilon}\big) - T \ln(\beta_2)\Big),
\end{align}

\begin{align}
   I_3 &\leq \frac{\eta_{T+1}^2 \gamma'}{2 (1 - \beta_1)(1 - \beta_1 / \beta_2)}\sum_{i=1}^d \Big( \ln \big(1 + \frac{\sum^{T}_{t=1}\beta_2^{T-t}(\tilde{\nabla}_iS_{t,w,\alpha}(\xx_t) )^2}{\epsilon}\big)- T \ln(\beta_2)\Big),
    \label{app:eq:common_b_bound2}
\end{align}

\begin{align}\label{I52}
\nonumber I_4&\leq\frac{T\eta_{T+1} 8L'^2 }{\beta_1W^2}
+\frac{2\eta_{T+1}^3(1-\alpha^{w-2})^2\gamma'^2}{\beta_1 W^2(1-\alpha)^2(1-\beta_1)(1 - \beta_1 / \beta_2)}
\\&\qquad \qquad \qquad \quad \sum_{i=1}^d \Big( \ln \big(1 + \frac{\sum^{T}_{t=1}\beta_2^{T-t}(\tilde{\nabla}_iS_{t,w,\alpha}(\xx_t) )^2}{\epsilon}\big)- T \ln(\beta_2)\Big),
\end{align}

\begin{align}\label{app:eq:common_c_bound2}
   I_5 & \leq  \frac{\eta_{T+1}^3  \gamma'^2 \beta_1}{ (1 - \beta_1)^{3}(1 - \beta_1/\beta_2)}
     \sum_{i=1}^d \Big( \ln \big(1
    + \frac{\sum^{T}_{t=1}\beta_2^{T-t}(\tilde{\nabla}_iS_{t,w,\alpha}(\xx_t) )^2}{\epsilon}\big)
     - T \ln(\beta_2)\Big).
\end{align}

\end{subequations}
Moreover, we have
 \begin{align}\label{hhj2}
\nonumber &\sum_{i=1}^{d}\ln\left(1 +\frac{\sum^{T}_{t=1}\beta_2^{T-t}(\tilde{\nabla}_iS_{t,w,\alpha}(\xx_t) )^2}{\epsilon} \right)
\stackrel{(i)} \leq d\ln\left(\frac{1}{d}\sum_{i=1}^{d}\big(1 +\frac{\sum^{T}_{t=1}\beta_2^{T-t}(\tilde{\nabla}_iS_{t,w,\alpha}(\xx_t) )^2}{\epsilon} \big)\right)\\\nonumber
 &= d\ln \left(1 +\frac{\sum^{T}_{t=1}\beta_2^{T-t}\sum_{i=1}^{d}(\tilde{\nabla}_iS_{t,w,\alpha}(\xx_t) )^2}{ d\epsilon} \right)
\\\nonumber &\stackrel{(ii)} \le
d\ln\left(1 +\frac{2\sum^{T}_{t=1}\beta_2^{T-t}\|\tilde{\nabla}S_{t,w,\alpha}(\xx_t)-\nabla S_{t,w,\alpha}(\xx_t) \|^2
+2\sum^{T}_{t=1}\beta_2^{T-t}\|\nabla S_{t,w,\alpha}(\xx_t) \|^2}{ d\epsilon} \right)
\\&\stackrel{(iii)}\le d\ln\left(1 +\frac{2(\bar{\mu}+L'^2)}{d\epsilon(1-\beta_2)} \right)~,
\end{align}
where (i) is due to the convex inequality
$\frac{1}{d}\sum_{i=1}^{d}\ln (a_i)\leq \ln (\frac{1}{d}\sum_{i=1}^{d} a_i)$, (ii) follows from $\| a+b\|^2\leq 2\|a \|^2+2\|b\|^2$, (iii) follows from Lemmas~\ref{lm:property} and
\ref{4543}(b).

Substituting \eqref{eq:cons2}-\eqref{app:eq:common_c_bound2} into \eqref{kk2} and using \eqref{hhj2} as well as the fact that
$\eta_{T+1}\leq \eta \frac{1-\beta_1}{\sqrt{1-\beta_2}}, $
we obtain

\begin{align}\label{kkb2}
\nonumber \sum_{t=1}^{T}\eta_{t+1}\sum_{i=1}^d \sum_{k=0}^{t-1}\beta_1^k&\frac{(\nabla_i S_{t+1-k,w,\alpha}(\xx_{t+1-k}))^2}{4\sqrt{\epsilon+\tilde{\upsilon}_{t+1,i}}}
                  \leq
    \nu + u \left(d\ln\Big(1 +\frac{2(\bar{\mu}+L'^2)}{d\epsilon(1-\beta_2)}  \Big) - T \ln(\beta_2)\right)\\
    &+\frac{3\eta(1-\beta_1)(1-\alpha^w)^2\kappa^2}{\beta_1^TW^2\sqrt{1-\beta_2}(1-\alpha)^2\sqrt{\epsilon}}\ln \frac{1}{\delta}\triangleq C,
\end{align}
where
\begin{align}\label{242}
\nonumber \nu&\triangleq\frac{2D(1-\alpha^w)T}{W(1-\alpha)}+\frac{D(1+\alpha^{w-1})T}{W}
    +\frac{D(1-\alpha^{w-1})(1+\alpha)T}{W(1-\alpha)}+\frac{T\eta (1-\beta_1) 8L'^2 }{\sqrt{1-\beta_2}\beta_1W^2}
    \\&\qquad \qquad+\frac{3\eta(1-\beta_1)(1-\alpha^w)^2\kappa^2}{\beta_1^TW^2\sqrt{1-\beta_2}(1-\alpha)^2\sqrt{\epsilon}}\ln \frac{1}{\delta},\\\nonumber
   % \vartheta_t_t &\triangleq \frac{4L'^2 \left(1 + \alpha^{2w-2} \right)}{W^2} + \frac{4L'^2(1 - \alpha^{2w-2})(1 + \alpha^2)}{W^2 (1- \alpha^2)},\\\nonumber
    u &\triangleq \frac{ d \eta^2(1- \beta_1)\gamma'}{2(1 - \beta_2)(1 - \beta_1 / \beta_2)} +
     \frac{d \eta^3  \gamma'^2 \beta_1}{(1 - \beta_1 / \beta_2) (1 - \beta_2)^{3/2}}\\&\qquad+
     \frac{2 d \eta \left(1+\frac{\sqrt{\bar \mu} }{\sqrt{W}}\right) \sqrt{1-\beta_1}}{(1 - \beta_1 / \beta_2)^{3/2}\sqrt{1 - \beta_2}}
     +\frac{2\eta^3(1-\beta_1)^2(1-\alpha^{w-2})^2\gamma'^2}{\beta_1 W^2(1-\alpha)^2(1-\beta_2)^{3/2}(1 - \beta_1 / \beta_2)}\label{252}.
\end{align}

Now, recalling that $\up_{t+1}=\beta_2 \up_{t}+(\tilde{\nabla}S_{t,w,\alpha}(\xx_t))^2=\sum_{l=1}^{t}\beta_2^{t-l}(\tilde{\nabla}S_{l,w,\alpha}(\xx_l))^2$, $\up_{1}=0$
and using the bound $\|a+b\|^2\leq 2\|a \|^2+2\| b\|^2$
yields that with probability at least $1-\delta$,
\begin{align}\label{mk2}
\nonumber &\beta_2 \sum_{i=1}^{d}\upsilon_{T,i} +\sum_{i=1}^{d} (\nabla_i S_{T,w,\alpha}(\xx_{T}))^2
+ \sum_{i=1}^{d}\big(\tilde\nabla_i S_{T,w,\alpha}(\xx_{T})-\nabla_i S_{T,w,\alpha}(\xx_{T})\big)^2 \\
\nonumber &=\sum_{i=1}^{d}\sum_{t=1}^{T-1}\beta_2^{T-t}(\tilde{\nabla}_i S_{t,w,\alpha}(\xx_t))^2
+\sum_{i=1}^{d}(\nabla_i S_{T,w,\alpha}(\xx_{T}))^2+ \sum_{i=1}^{d}\big(\tilde\nabla_i S_{T,w,\alpha}(\xx_{T})-\nabla_i S_{T,w,\alpha}(\xx_{T})\big)^2
\\ \nonumber&\leq
2\sum_{i=1}^{d}\sum_{t=1}^{T-1}\beta_2^{T-t}(\tilde{\nabla}_i S_{t,w,\alpha}(\xx_t)-\nabla_i S_{t,w,\alpha}(\xx_t))^2+2\sum_{i=1}^{d}\sum_{t=1}^{T-1}\beta_2^{T-t}(\nabla_i S_{t,w,\alpha}(\xx_t))^2
\\\nonumber &\qquad \qquad+2\sum_{i=1}^{d}(\nabla_i S_{T,w,\alpha}(\xx_{T}))^2
 + 2\sum_{i=1}^{d}\big(\tilde\nabla_i S_{T,w,\alpha}(\xx_{T})-\nabla_i S_{T,w,\alpha}(\xx_{T})\big)^2
\\ &\leq 2 \sum_{t=1}^{T}\sum_{i=1}^{d}(\tilde{\nabla}_i S_{t,w,\alpha}(\xx_t)-\nabla_i S_{t,w,\alpha}(\xx_t))^2+2\sum_{i=1}^{d}\sum_{t=1}^{T}(\nabla_i S_{t,w,\alpha}(\xx_t))^2
\le\
  2T\bar{\mu}+2Z ,
\end{align}
where $Z\triangleq\sum^{T}_{t=1}\|\nabla S_{t,w,\alpha}(\xx_t)\|^2$, $\bar{\mu}= \kappa^2\ln \frac{\exp\left(\frac{w\sum^{w-1}_{r=0}\alpha^{2r}}{W^2}\right)}{\delta}$, and the last inequality follows from Lemma \ref{4543}(b).
Using \eqref{mk2}, definition of $\tilde{\upsilon}_{t+1,i}$, and $\eta_{t+1}= \eta (1 - \beta_1) \sqrt{\sum_{r=0}^{t} \beta_2^r}$, with probability at least $1-\delta$, we obtain
\begin{align}\label{432}
\nonumber &\qquad \sum_{t=1}^{T}\eta_{t+1}\sum_{i=1}^d \frac{\sum_{k=0}^{t-1}\beta_1^k(\nabla_i S_{t+1-k,w,\alpha}(\xx_{t+1-k}))^2}{4\sqrt{\epsilon+\tilde{\upsilon}_{t+1,i}}}
\\&\nonumber=\eta  \sum_{t=1}^{T} \sqrt{\sum_{r=0}^{t} \beta_2^r}\sum_{i=1}^d \frac{\sum_{k=0}^{t-1}(1 - \beta_1) \beta_1^k(\nabla_i S_{t+1-k,w,\alpha}(\xx_{t+1-k}))^2}{4\sqrt{\epsilon+\frac{1}{W}\big(\beta_2\upsilon_{t,i} + (\nabla_i S_{t,w,\alpha}(\xx_t))^2 \big)+ \hat{\mu}_{T,i}}}\\\nonumber
&\ge
\eta\sum_{t=1}^{T} \sqrt{\sum_{r=0}^{t} \beta_2^r}\sum_{i=1}^d\frac{\sum_{k=0}^{t-1}(1-\beta_1)\beta_1^k(\nabla_i S_{t+1-k,w,\alpha}(\xx_{t+1-k}))^2}{4\sqrt{\epsilon+\frac{1}{W}\big(\beta_2\upsilon_{T,i}
+ (\nabla_i S_{T,w,\alpha}(\xx_T))^2 + \sum_{t=1}^{T}  (\nabla_i S_{t,w,\alpha}(\xx_t))^2\big)+\hat{\mu}_{T,i}}}
\\\nonumber
&\stackrel{(i)}\ge
\frac{\varsigma\eta}{\sqrt{1-\beta_2}} \sum_{t=1}^{T}\sum_{i=1}^d\frac{\sum_{k=0}^{t-1}(1-\beta_1)\beta_1^k(\nabla_i S_{t+1-k,w,\alpha}(\xx_{t+1-k}))^2}{4\sqrt{\epsilon+\frac{1}{W}\big(\beta_2\sum_{i=1}^{d}\upsilon_{T,i} +\sum_{i=1}^{d} (\nabla_i S_{T,w,\alpha}(\xx_T))^2 + \sum_{t=1}^{T} \sum_{i=1}^{d} (\nabla_i S_{t,w,\alpha}(\xx_t))^2 \big)+\hat{\mu}_{T,i}}}
\\\nonumber
&\stackrel{(ii)}\ge
\frac{\varsigma\eta(1-\beta_1)\sum_{j=1}^{T}\sum_{i=1}^d(\nabla_i S_{j,w,\alpha}(\xx_{j}))^2}{4\sqrt{1-\beta_2}\sqrt{\epsilon+\frac{1}{W}\big(\beta_2\sum_{i=1}^{d}\upsilon_{T,i} +\sum_{i=1}^{d} (\nabla_i S_{T,w,\alpha}(\xx_T))^2 + Z\big) +\hat{\mu}_{T,i}}}\\
&\stackrel{(iii)}\ge
\frac{\varsigma\eta(1-\beta_1)Z}{4\sqrt{1-\beta_2}\sqrt{\epsilon+\frac{1}{W}\big( 3Z + 2 T \bar{\mu}\big)}}~,
\end{align}
where (i) is obtained by using our assumption that $\sqrt{\sum_{r=0}^{t} \beta_2^r}\geq \frac{\varsigma}{\sqrt{1-\beta_2}}$, (ii) is derived by following the arguments
in \eqref{cv} that by changing of index $j = t +1-k$, for all $i\in [d]$ we have
\begin{align*}\label{cv2}
\nonumber &\sum_{t=1}^{T}
                  \sum_{k=0}^{t-1}(1-\beta_1)\beta_1^k (\nabla_i S_{t+1-k,w,\alpha}(\xx_{t+1-k}))^2
\geq  (1-\beta_1)\sum_{j=1}^{T} (\nabla_i S_{j,w,\alpha}(\xx_{j}))^2.
\end{align*}
The (iii) follows from \eqref{mk2}.\\
Considering equations \eqref{kkb2} and \eqref{432}, we then observe that with probability at least $1-\delta$,
\begin{equation*}
\frac{\varsigma\eta(1-\beta_1)Z}{4\sqrt{1-\beta_2}\sqrt{\epsilon+\frac{1}{W}\big( 3Z + 2T\bar{\mu}\big)}}
\le C~.
\end{equation*}
In view of Lemma~\ref{lem:quadratic}, we get
\begin{align}\label{992}
 & \sum^T_{t=1}\|\nabla S_{t,w,\alpha}(\xx_t)\|^2\le
\frac{4\sqrt{1-\beta_2}C}{\varsigma\eta(1-\beta_1)}\Big(\sqrt{\epsilon}+\sqrt{\frac{2T\bar{\mu}}{W}}\Big)
+\frac{48(1-\beta_2)C^2}{W\varsigma^2\eta^2(1-\beta_1)^2}.
\end{align}
Here,
$C \triangleq \nu + u \left(d\ln\big(1 +\frac{2(\bar{\mu}+L'^2)}{d\epsilon(1-\beta_2)} \big) - T \ln(\beta_2)\right)+y,$ and
$$\bar{\mu}\triangleq \kappa^2\ln \frac{\exp\left(\frac{w\sum^{w-1}_{r=0}\alpha^{2r}}{W^2}\right)}{\delta}, \qquad y\triangleq\frac{3\eta(1-\beta_1)(1-\alpha^w)^2\kappa^2}{\beta_1^T\sqrt{1-\beta_2}W^2(1-\alpha)^2\sqrt{\epsilon}}\ln \frac{1}{\delta},$$
where $\nu$ and $u$ are defined in \eqref{242} and \eqref{252}, respectively.
Further, we can decompose
\begin{align}\label{vdd2}
 \nu&=\nu_1+\frac{T\eta (1-\beta_1) 8L'^2 }{\sqrt{1-\beta_2}\beta_1W^2},
\end{align}
where
\begin{align*}
\nu_1&=\frac{2D(1-\alpha^w)T}{W(1-\alpha)}+\frac{D(1+\alpha^{w-1})T}{W}+\frac{D(1-\alpha^{w-1})(1+\alpha)T}{W(1-\alpha)}.
\end{align*}
Now use a similar argument as in equation \eqref{frd1} in the term $\nu_1$ to have
\begin{align}\label{frd2}
 \nu_1&\leq
\frac{4D T}{W}\Big(\frac{2-\alpha^{w}+\alpha^{w-1}}{1-\alpha}\Big).
\end{align}
Then plugging \eqref{frd2} into \eqref{vdd2} yields
\begin{align*}
\nu &\leq \frac{4D T}{W}\Big(\frac{2-\alpha^{w}+\alpha^{w-1}}{1-\alpha}\Big)+\frac{T\eta (1-\beta_1) 8L'^2 }{\sqrt{1-\beta_2}\beta_1W^2}.
\end{align*}
As $\alpha \rightarrow 1^{-}$, we get
\begin{equation}
\begin{aligned}\label{nnb2}
  &\nu \leq \frac{4D T}{W}+\frac{T\eta (1-\beta_1) 8L'^2 }{\sqrt{1-\beta_2}\beta_1W^2}\triangleq\varpi_1,\\
  &\bar{\mu}\stackrel{\alpha \rightarrow 1^{-}}=\kappa^2\ln \frac{e}{\delta}\triangleq\zeta,\\
&  u \stackrel{\alpha \rightarrow 1^{-}}= \frac{ d \eta^2(1- \beta_1)\gamma'}{2(1 - \beta_2)(1 - \beta_1 / \beta_2)} +
     \frac{d \eta^3  \gamma'^2 \beta_1}{(1 - \beta_1 / \beta_2) (1 - \beta_2)^{3/2}}
     \\&\qquad+
     \frac{2 d \eta \left(1+\frac{\sqrt{\zeta} }{\sqrt{W}}\right) \sqrt{1-\beta_1}}{(1 - \beta_1 / \beta_2)^{3/2}\sqrt{1 - \beta_2}}
     +\frac{2\eta^3(1-\beta_1)^2\gamma'^2}{\beta_1 (1-\beta_2)^{3/2}(1 - \beta_1 / \beta_2)}\triangleq \varpi_2,\\
 &    y\stackrel{\alpha \rightarrow 1^{-}}=\frac{3\eta(1-\beta_1)\kappa^2}{\beta_1^T\sqrt{1-\beta_2}\sqrt{\epsilon}}\ln \frac{1}{\delta}\triangleq \varpi_3.
\end{aligned}
\end{equation}
The desired result then follows by inserting \eqref{nnb2} to \eqref{992}.
\end{proof}

\subsubsection*{ PROOF OF COROLLARY~\ref{corr112}}
\begin{cor*}
Under the same conditions stated in Theorem \ref{adam2}, using $\beta_2=1-1/T$, $\eta=\eta_1/\sqrt{T}$, $\beta_1/\beta_2\approx \beta_1$, $w\in\Theta(T)$, and $\alpha\rightarrow 1^{-}$ yields a regret bound of order
\begin{equation*}
\sum^T_{t=1} \|\nabla S_{t,w,\alpha}(\xx_t)\|^2 \leq \mathcal{O}(\ln T).
\end{equation*}
\end{cor*}
\begin{proof}
By Theorem \ref{adam2}, we have
\begin{equation*}
\sum^T_{t=1}\|\nabla S_{t,w,\alpha}(\xx_t)\|^2
\le \frac{4\sqrt{1-\beta_2}C}{\varsigma\eta(1-\beta_1)}\Big(\sqrt{\epsilon}+\sqrt{\frac{2\zeta T}{W}}\Big)+\frac{48(1-\beta_2)C^2}{W\varsigma^2\eta^2(1-\beta_1)^2}\triangleq I_1+I_2+I_3~,
\end{equation*}
where $C \triangleq \varpi_1 + \varpi_2 \left(d\ln\big(1 + \frac{2(\zeta+L'^2)}{d\epsilon ( 1- \beta_2)}\big) - T \ln(\beta_2)\right)+\varpi_3$,
$\zeta \triangleq \kappa^2\ln \frac{e}{\delta}$,  $\varpi_1$, $\varpi_2$ and $\varpi_3$ are defined as in \eqref{w1}, \eqref{w2} and \eqref{w3}, respectively.
Recall that $W\triangleq \sum_{r=0}^{w-1}\alpha^r$. As $\alpha\rightarrow 1^{-}$ and $w\in\Theta(T)$, we get the following equalities:
\begin{align*}
\varpi_1&=\frac{4D T}{W}+\frac{8T\eta_1 (1-\beta_1)L'^2}{\beta_1W^2}=\mathcal{O}(1+\frac{1}{T}),\\
    \varpi_2 &\approx \frac{ d \eta_1^2\gamma'}{2} +
     \frac{d \eta_1^3  \gamma'^2 \beta_1 }{(1 - \beta_1 )} +
     \frac{ 2d \eta_1 (1+\sqrt{\frac{\kappa^2\ln \frac{e}{\delta}}{W}}) }{(1 - \beta_1 )}
     +\frac{2\eta_1^3(1-\beta_1)\gamma'^2}{\beta_1 }=\mathcal{O}(1+\frac{1}{\sqrt{T}}),\\
     \varpi_3 &= \frac{3\eta_1(1-\beta_1)\kappa^2}{W^2\beta_1^T\sqrt{\epsilon}}\ln \frac{1}{\delta}=\mathcal{O}(\frac{1}{T^2}),\\
C&=\varpi_1 + \varpi_2 \left(d\ln\big(1 + \frac{2(\frac{\sigma^2}{W}+L'^2)T}{d\epsilon }\big) - T \ln(1-1/T)\right)+\varpi_3\\
&=\varpi_1 + \varpi_2 \left(d\ln\big(1 + \frac{2(\frac{\sigma^2}{W}+L'^2)T}{d\epsilon }\big) +1\right)+\varpi_3=\mathcal{O}(\ln T).
\end{align*}
As a result,
\begin{align*}
  I_1&=\frac{4\sqrt{1-\beta_2}C\sqrt{\epsilon}}{\varsigma\eta(1-\beta_1)}  =\frac{4C\sqrt{\epsilon}}{\varsigma\eta_1(1-\beta_1)} =\mathcal{O}(\ln T),\\
  I_2&=\frac{4\sqrt{1-\beta_2}C\sqrt{2\zeta T}}{\varsigma\eta(1-\beta_1)\sqrt{W}}=\frac{4C}{\varsigma\eta_1(1-\beta_1)}\sqrt{\frac{2\kappa^2\ln \frac{e}{\delta}}{W} T}=\mathcal{O}(\ln T), \\
  I_3&=\frac{48(1-\beta_2)C^2}{W\varsigma^2\eta^2(1-\beta_1)^2}=\frac{48C^2}{W\varsigma^2\eta_1^2(1-\beta_1)^2}=\mathcal{O}(\ln T).
\end{align*}
Combine the above results we can easily have the desired result.
\end{proof}

\begin{lem}\label{bn}
Given $0 <  a < 1$ and $Q \in \nat$, we have,
\begin{equation*}
    \label{app:eq:sum_geom_sqrt}
    \sum_{q = 0}^{Q - 1} \frac{a^q}{\sqrt{q+1}}
    \leq \frac{2}{a\sqrt{(1 - a)}}.
\end{equation*}
\end{lem}

\begin{proof}
Observe that
\begin{align*}
\nonumber
& \sum_{q = 0}^{Q - 1} \frac{a^q}{2\sqrt{q+1}}  \leq   \int_0^{\infty} \frac{a^{x}}{2 \sqrt{x+1}}\dd{}x =
        \int_0^{\infty} \frac{\ee^{\ln(a) x}}{2 \sqrt{x+1}}\dd{}x
        \\&\quad\quad \stackrel {y = \sqrt{x+1}}= \int_0^{\infty}\ee^{\ln(a) (y^2-1)}\dd{}y
       \\&\quad \quad\stackrel{u = \sqrt{-2 \ln(a)} y}= \frac{1}{a\sqrt{-2\ln(a)}}\int_0^{\infty} \ee^{-u^2 / 2}\dd{}u
        \\&\quad\quad= \frac{\sqrt{\pi}}{2 a\sqrt{-\ln(a)}}\leq \frac{\sqrt{\pi}}{2 a\sqrt{1-a}}\leq \frac{2}{2a\sqrt{1-a}},
\end{align*}
where the last inequality is by the fact that $\sqrt{1-a}\leq \sqrt{-\ln (a)}$.
\end{proof}

\begin{lem}\label{lem:quadratic}
Let $a,b,c \geq 0$. If for $Z \geq 0$, $\displaystyle \frac{Z}{\sqrt{cZ+a}} \leq b$, then $\displaystyle Z \leq c b^2 + b\sqrt{a}$.
\end{lem}
\begin{proof}
Consider,
\begin{align*}
    Z^2 -c b^2Z - b^2a \leq 0.
\end{align*}
Note that for the equation of second order, we have $\Delta = c^2 b^4 + 4b^2a \geq 0$ which yields
\begin{align*}
    Z &\leq \frac{c b^2 + \sqrt{c^2 b^4 + 4b^2a}}{2}
      \leq \frac{c b^2 + \sqrt{c^2 b^4} + \sqrt{4b^2a}}{2}
      =c b^2 + b\sqrt{a}.
\end{align*}
\end{proof}

\begin{lem}\cite[Lemma 3.2]{aydore2019dynamic}\label{lm:sumobj22}
 Given Assumption~\ref{ass:loss}(iv), for any $0<\alpha<1$ and $w$, we have:
$$S_{t+1,w,\alpha}(\xx_{t+1})-S_{t,w,\alpha}(\xx_{t+1})\le \frac{D(1+\alpha^{w-1})}{W}+\frac{D(1-\alpha^{w-1})(1+\alpha)}{W(1-\alpha)}.$$
\end{lem}

\begin{lem}\cite[Lemma 3.3]{aydore2019dynamic}
 Given Assumption~\ref{ass:loss}(iv), for any $0<\alpha<1$ and $w$, we have:
$$S_{t,w,\alpha}(\xx_{t})-S_{t+1,w,\alpha}(\xx_{t+1})\le\frac{2D(1-\alpha^w)}{W(1-\alpha)}.$$
\label{lm:sumobj2}
\end{lem}

\begin{lem}\cite[Lemma 6.2]{defossez2020simple}
\label{lemma:sum_ratio}
We assume we have $0 < \beta_2 \leq 1$ and a non-negative sequence $\{a_n\}_{n\in \nat}$. We define
$b_n = \sum_{j=1}^n \beta_2^{n - j} a_j$ with the convention $b_0 = 0$. Then, we have:
\begin{equation*}
    \sum_{j=1}^N \frac{a_j}{\epsilon + b_j} \leq \ln\left(1 + \frac{b_N}{\epsilon}\right) - N
    \ln(\beta_2).
\end{equation*}
\end{lem}

\begin{lem}\cite[Lemma A.2]{defossez2020simple}\label{app:lemma:sum_ratio}
We assume we have $0<\beta_2 \leq 1$ and $0 < \beta_1 < \beta_2$, and a sequence of real numbers $\{a_n\}_{n\in \nat}$. We define
$b_n \deq \sum_{j=1}^n \beta_2^{n - j} a_j^2$ and $c_n \deq \sum_{j=1}^n \beta_1^{n-j} a_j$.
Then, we have:
\begin{equation*}
    \sum_{j=1}^n \frac{c_j^2}{\epsilon + b_j} \leq
    \frac{1}{(1 - \beta_1)(1 - \beta_1 / \beta_2)} \left(\ln\left(1 + \frac{b_n}{\epsilon}\right) - n\ln(\beta_2)\right).
\end{equation*}
\end{lem}

\begin{lem}\cite[Lemma A.3]{defossez2020simple}
\label{app:lemma:sum_geom_sqrt}
Given $0 <  a < 1$ and $Q \in \nat$, we have,
\begin{equation*}
    \label{app:eq:sum_geom_sqrt}
    \sum_{q = 0}^{Q - 1} a^q \sqrt{q + 1} \leq \frac{1}{1 - a} \left( 1 + \frac{\sqrt{\pi}}{2 \sqrt{-\ln(a)}}\right)
    \leq \frac{2}{(1 - a)^{3/2}}.
\end{equation*}
\end{lem}

\begin{lem}\cite[Lemma A.4]{defossez2020simple}\label{app:lemma:sum_geom_32}
Given $0 <  a < 1$ and $Q \in \nat$, we have,
\begin{equation*}
    \label{app:eq:sum_geom_32}
    \sum_{q = 0}^{Q - 1} a^q \sqrt{q} (q + 1)
    \leq \frac{4 a}{(1 - a)^{5/2}}.
\end{equation*}
\end{lem}

\end{document}